%% file: main.tex
\documentclass[10pt,twocolumn,letterpaper]{article}

\usepackage[pagenumbers]{cvpr} 

\usepackage{times}
\usepackage{epsfig}
\usepackage{graphicx}
\usepackage{amsmath}
\usepackage{amssymb}
\usepackage{amsthm}
\usepackage{bm}
\usepackage{multirow}
\usepackage{booktabs}
\usepackage[table,xcdraw]{xcolor}  
\usepackage{overpic}
\usepackage{nicefrac}

\newtheorem{proposition}{Proposition}

\input{preamble}

\definecolor{cvprblue}{rgb}{0.21,0.49,0.74}
\usepackage[pagebackref,breaklinks,colorlinks,citecolor=cvprblue]{hyperref}

\def\paperID{5021} 
\def\confName{CVPR}
\def\confYear{2025}

\title{HotSpot: Signed Distance Function Optimization with an Asymptotically Sufficient Condition}

\author{%
Zimo Wang\footnotemark[1] \quad Cheng Wang\footnotemark[1] \quad 
Taiki Yoshino \quad Sirui Tao \quad Ziyang Fu \quad Tzu-Mao Li\\
UC San Diego\\
}

\begin{document}

\Crefname{equation}{Eq.}{Eqs.}
\Crefname{figure}{Fig.}{Figs.}

\twocolumn[{%
 \renewcommand\twocolumn[1][]{#1}%
 \maketitle
 
 \vspace{-5mm}
 \centering
 \includegraphics[width=1.0\linewidth]{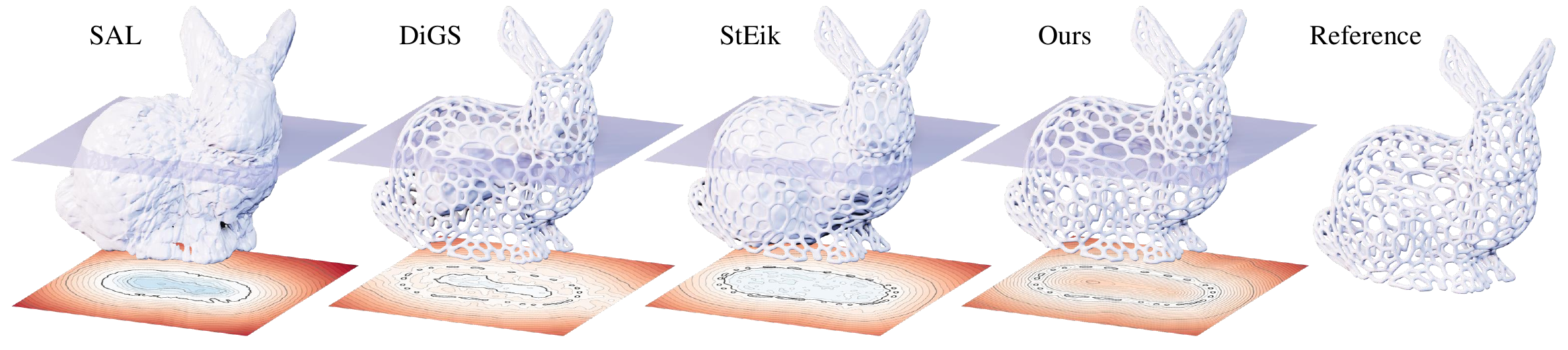}
 \vspace{-15pt}
 \captionof{figure}{ 
   \label{fig:teaser} We propose \MethodName, a neural signed distance function optimization method that establishes an asymptotic sufficient condition to guarantee convergence to a true distance function, enabling precise surface reconstruction and level set representation for complex shapes.
   Here we show a reconstruction from a point cloud sampled from the reference bunny (taken from Mehta et al.~\cite{Mehta:2022:LST}) on the right.
   In the inset, we visualize the recovered signed distance function on a horizontal slice, using warm colors for positive values and cool for negative (zoom in for details).
   Our reconstruction is significantly more accurate than prior works (SAL~\cite{atzmon2020sal}, DiGS~\cite{ben2022digs}, and StEik~\cite{yang2023steik}).
  }
   \vspace{6mm}
}]

\input{sec/0_abstract}
\renewcommand{\thefootnote}{\fnsymbol{footnote}}
\footnotetext[1]{Equal contribution.}
\renewcommand{\thefootnote}{\arabic{footnote}}
\input{sec/1_intro}
\input{sec/2_related}
\input{sec/3_background}
\input{sec/4_method}

\input{sec/5_experiments}
\input{sec/6_discussion}

\input{sec/7_conclusion}

\paragraph{Acknowledgements.}

This work was supported in part by NSF grants 2127544, 2238839, 2100237, 2120019, and gifts from Adobe and Google. Additionally, we would like to thank Bing Xu, Zilu Li, Yash Belhe, Ishit Mehta, and Jianan Xiao for their suggestions, discussions, proofreading, and providing reference materials. 

{
    \small
    \bibliographystyle{unsrt}
    \bibliography{main}
}

\input{sec/X_suppl}

\end{document}

%% file: preamble.tex
\newcommand{\MethodName}[0]{\textsc{HotSpot}}

\usepackage{graphicx}  
\usepackage{adjustbox} 
\usepackage{booktabs}  
\usepackage{subcaption} 
\usepackage{multirow}
\usepackage{wrapfig}
\usepackage[resetlabels,labeled]{multibib}
\newcites{supp}{Supplementary References}

%% file: sec/0_abstract.tex
\begin{abstract}
We propose a method, \MethodName, for optimizing neural signed distance functions.
Existing losses, such as the eikonal loss, act as necessary but insufficient constraints and cannot guarantee that the recovered implicit function represents a true distance function, even if the output minimizes these losses almost everywhere.
Furthermore, the eikonal loss suffers from stability issues in optimization.
Finally, in conventional methods, regularization losses that penalize surface area distort the reconstructed signed distance function.
We address these challenges by designing a loss function using the solution of a screened Poisson equation.
Our loss, when minimized, provides an asymptotically sufficient condition to ensure the output converges to a true distance function. 
Our loss also leads to stable optimization and naturally penalizes large surface areas.
We present theoretical analysis and experiments on both challenging 2D and 3D datasets and show that our method provides better surface reconstruction and a more accurate distance approximation.
\vspace{-20pt}
\end{abstract}

%% file: sec/1_intro.tex
\section{Introduction}
\label{sec:intro}

Learning neural signed distance function requires designing loss functions so that the zero crossings of the implicit function are at the desired locations (e.g., adhere to a sparse input point cloud), and that the implicit function returns the correct signed distance.
Designing the loss function proves to be challenging for shapes with complex topology and details (\Cref{fig:teaser}).
We propose \MethodName, a method based on a relation between the screened Poisson equation and distance that enjoys both theoretical and practical benefits.

\begin{figure}[t]
    \centering
    \includegraphics[width=1.0\linewidth]{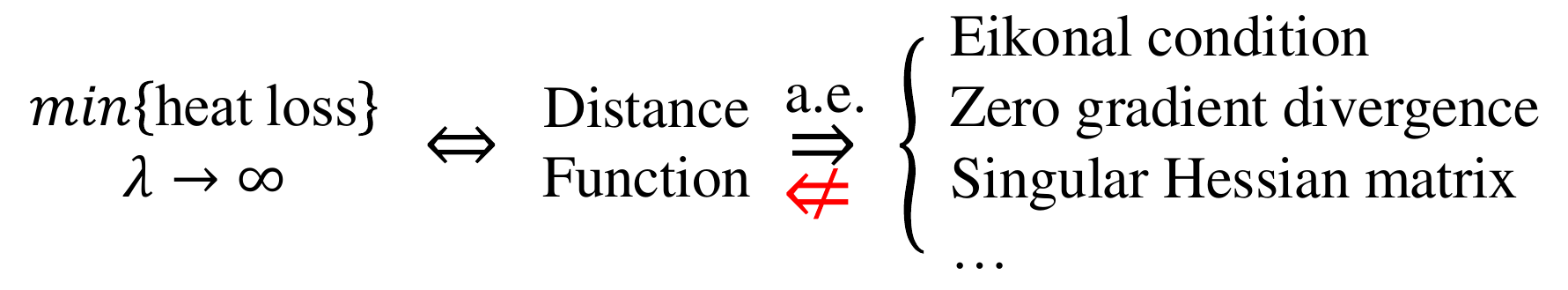}
    \vspace{-20pt}
    \caption{Existing constraints are only \textbf{necessary} but \textbf{insufficient} conditions. That is, there exist solutions, like \Cref{fig:1d}, that satisfy them almost everywhere yet fail to be distance functions. In contrast, we use a heat loss, modeled by a screened Poisson equation with parameter $\lambda$. As the parameter $\lambda$ goes to infinity, the minimizer converges to the distance function and excludes other solutions. Thus, our method is asymptotically both necessary and sufficient.}
    \label{fig:nands}
    \vspace{-10pt}
\end{figure}

\begin{figure}[t]
    \centering
    \includegraphics[width=0.75\linewidth]{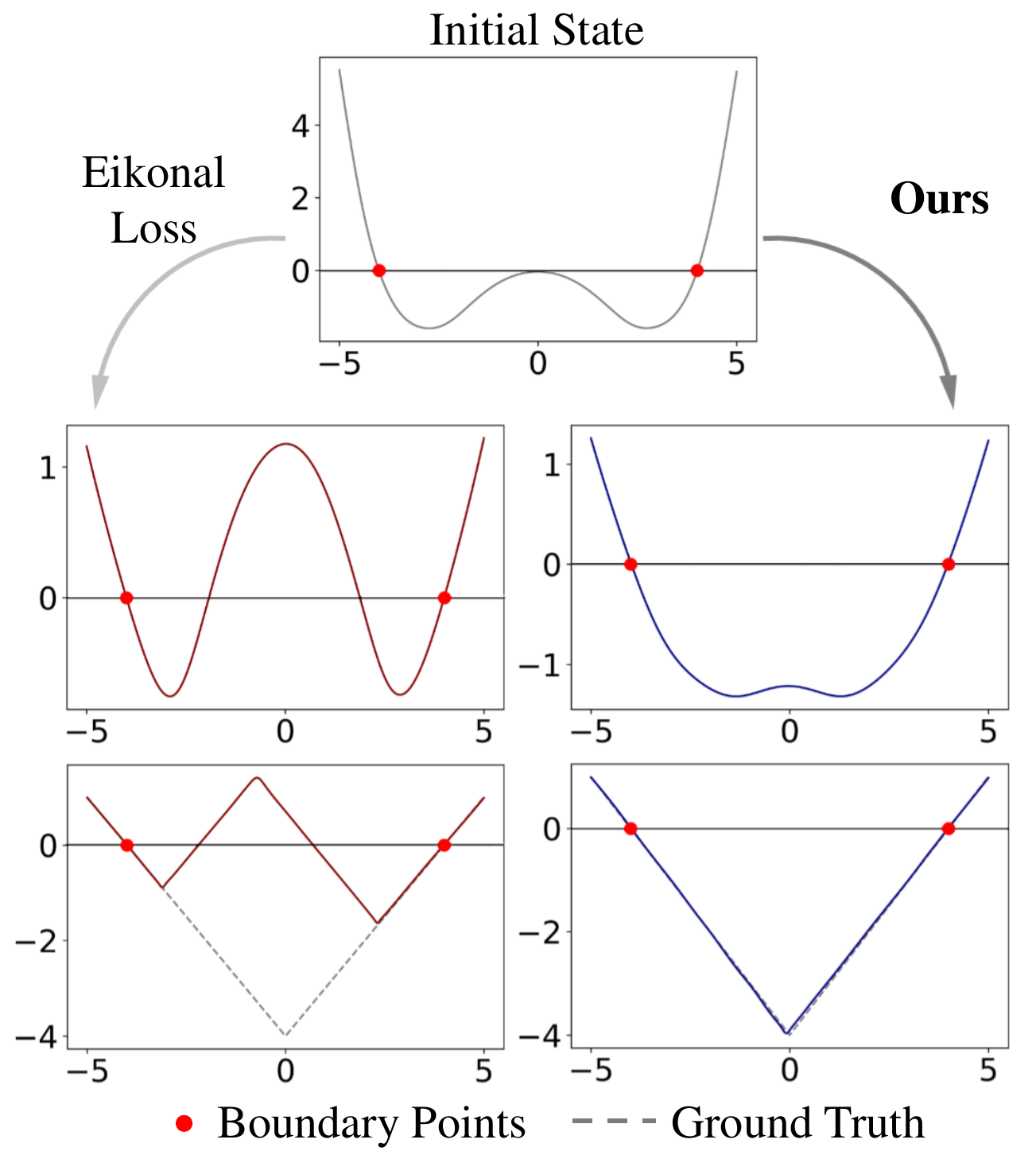}
    \vspace{-5pt}
    \caption{
    We illustrate a 1D neural signed distance function optimization using the classical eikonal loss~\cite{gropp2020implicit} and our model. 
    The $x$ axis is the domain and the $y$ axis shows the output of the implicit function. 
    The middle and bottom rows show the intermediate and final states of the optimization respectively. The eikonal loss, as a necessary but insufficient condition, is incapable of converging to the actual signed distance function (dashed line), even when the function satisfies the eikonal equation almost everywhere. 
    }
    \vspace{-15pt}
    \label{fig:1d}
\end{figure}

A major challenge in signed distance function optimization is ensuring the implicit function follows the true distance.
A common regularization loss used is the eikonal equation, which constrains the gradient norm of an implicit function to be $1$ almost everywhere. 
If the implicit function is a signed distance function, then it satisfies the eikonal equation. However, the converse is not true. 
That is, satisfying the eikonal equation is only a \emph{necessary condition} for the signed distance function, not a \emph{sufficient condition}~\cite{marschner2023constructive}.
\Cref{fig:nands} presents the logic and \Cref{fig:1d} shows an example.

Another key challenge lies in the optimization. 
It is known that the eikonal equation regularization is \emph{unstable}~\cite{yang2023steik}, and the instability can make optimization converge to suboptimal results or even diverge.

Furthermore, to address optimization ill-posedness and instability, existing methods penalize the surface area of the shape.
We prove this degrades the reconstruction quality and distorts the output level sets, thereby leading to a delicate balance between over- and under-smoothing.

We propose a simple model that can address all the challenges above. 
Our model, \MethodName, is based on a screened Poisson equation and a classical relation between heat transfer and distance~\cite{Varadhan:1967:BFS} that was popular for approximating both Euclidean and geodesic distance~\cite{Crane:2013:GHN,belyaev2015variational,Belyaev:2020:ASD}.
Using this relation, we design a loss function for optimizing neural signed distance functions.
As our contributions, we show, both theoretically and empirically, that our model
\begin{itemize}
    \item serves as a sufficient condition ensuring the difference between the minimizer and the actual distance is bounded and decreases linearly as the parameter $\lambda$ increases (\Cref{fig:nands} and \Cref{fig:1d}, right).
    \item is \emph{stable} both spatially and temporally, in the sense that a slight perturbation of the implicit function only introduces changes in a local region, and that the dynamical system formed by the gradient flow will converge in the long run.
    \item penalizes the surface area naturally, while remaining an undistorted signed distance function.
    \item works well in both 2D and 3D shapes, and it excels at approaching signed distance functions while optimizing for complex, high-genus shapes.
\end{itemize}

%% file: sec/2_related.tex
\section{Related Work}
\label{sec:related_work}

\noindent \textbf{Implicit surfaces.} Modeling and reconstructing surfaces using implicit functions has a long history in computer vision and graphics~\cite{Blinn:1982:GAS,Bloomenthal:1990:ITI,Hoppe:1992:SRU,Curless:1996:VMB,Izadi:2011:KR3}. 
Recently, there is a surge of interest to model surfaces using \emph{neural} implicit functions~\cite{mescheder2019occupancy,chen2019learning,Davies:2020:EWN,Williams:2021:NSF,Xie:2022:NFV}.
Neural implicit representations are favored for their compactness, ability to represent detailed surfaces, and compatibility to gradient-based optimization~\cite{Mehta:2022:LST}.

\noindent \textbf{Neural signed distance functions.} Among the classes of implicit surfaces, \emph{signed distance functions}, where the implicit function outputs the distance to the closest point on the surface, and the sign represents whether the point is inside or outside the solid object, is one of the most popular variants. 
Signed distance functions enable an efficient rendering algorithm called sphere tracing~\cite{Hart:1996:STG,Seyb:2019:NST,Galin:2020:STL,Sharp:2022:SDG} and helps geometry compression~\cite{Takikawa:2021:NGL}.
They can also be used for collision detection~\cite{Fuhrmann:2003:DFR,Guendelman:2003:NRB,Macklin:2020:LOR} and numerical simulation~\cite{Sawhney:2020:MCG}.
Traditionally, signed distance functions are often represented as grids~\cite{pumarola2022visco,peng2021shape,chen2023gridpull}.
More recently, neural signed distance functions have become an appealing geometric representation~\cite{Park:2019:DLC,sitzmann2020implicit,Coiffier:2024:1ND}.
(Neural) Signed distance functions are also often used in inverse rendering~\cite{Jiang:2020:SDR,Kellnhofer:2021:NLR,Niemeyer:2020:DVR,Yariv:2020:MNS,Zhang:2021:PIR,Bangaru:2022:DRN,Vicini:2022:DSD}, for their flexibility to obtain surfaces with complex topology.

\noindent \textbf{Regularizing and initializing neural signed distance functions.}
A commonly used loss for optimizing neural signed distance functions is the eikonal regularization, which encourages the norm of the gradient of the implicit function to be one~\cite{gropp2020implicit,Yariv:2020:MNS}. 
However, the eikonal regularization alone is far from sufficient, as 1) minimizing eikonal loss does not necessarily lead to a distance field~\cite{marschner2023constructive,pumarola2022visco}, and 2) its optimization is unstable~\cite{yang2023steik}.
Previous works have proposed initialization to bootstrap the optimization with a simple shape~\cite{gropp2020implicit,ben2022digs}.
Recent works show that minimizing the divergence and the surface area of the implicit function can lead to better reconstruction~\cite{ben2022digs,Zhang:2022:CRN} and stabilizes the optimization~\cite{yang2023steik}.
Second-order information is also often used~\cite{Zhang:2022:CRN,Wang:2024:AGH, wang2023neural, dong2024neurcadrecon} for regularization. All losses above are based on \emph{necessary} properties of real distance but neither establish convergence nor exclude non-SDF solutions.

Specialized architectures have been proposed to limit the Lipschitz constant of the implicit function~\cite{Coiffier:2024:1ND}, satisfy $p$-Poisson equation~\cite{park2024p}, or represent high-frequency details~\cite{sitzmann2020implicit,Takikawa:2021:NGL,Li:2023:NHN}.
Some research~\cite{Lipman:2021:PTD,dai2022neural} has shown a link between occupancy function optimization and an approximate signed distance field regularization.
Our experiments further reveal that some implementations~\cite{atzmon2020sal,ma2020neural,Atzmon:2021:SSA} optimizing SDFs with the closest point may struggle when querying near the interpolated surface.

We propose a new loss and method for optimizing signed distance functions. 
While derived from very different mathematics, our loss turns out to have a close connection to Lipman's PHASE loss~\cite{Lipman:2021:PTD}.
We show that our theory leads to significantly more numerically stable optimization, and a different conclusion in the parameter setting that leads to drastically different results.

\noindent \textbf{Distance approximation using heat transfer.}
Our derivation is inspired by a relation between heat transfer and approximated distance~\cite{Varadhan:1967:BFS}.
This relation has been used in computer graphics for efficiently computing the geodesic distance on surfaces~\cite{Crane:2013:GHN,belyaev2015variational,Belyaev:2020:ASD,Sharp:2019:VHM,Feng:2024:HMG}. 
We apply the relation for optimizing neural signed distance functions, and analyze the properties of the resulting optimization in this context.







%% file: sec/3_background.tex
\section{Background and Motivation}
\label{sec:bg}

Given an input point cloud $\bm{x}_i \in \mathbb{R}^d$ (we focus on 2D and 3D in this work) with $N$ points without normal information that lies on an unknown surface $\Gamma$ of a solid object $S$, our goal is to find a signed distance function $f(\bm{x})$, such that
\begin{equation}
    f(\bm{x}) = 
    \begin{cases}
        -d_{\Gamma}(\bm{x}) & \text{ if } \bm{x} \in S \\
        d_{\Gamma}(\bm{x}) & \text{ otherwise, }
    \end{cases}
\end{equation}
where $d_{\Gamma}(\bm{x})$ denotes the closest distance from point $\bm{x}$ to the surface $\Gamma$. 
In practice, we approximate the signed distance function $f(\bm{x})$ using a neural network $u(\bm{x}): \mathbb{R}^d \rightarrow \mathbb{R}$ parameterized by its weights and use gradient-based optimization to figure out the weights.
Previous work (e.g.,~\cite{gropp2020implicit}) often uses a linear combination of a boundary loss $L_{\text{boundary}}$ and a eikonal loss $L_{\text{eikonal}}$:
\begin{align}
    \label{eq:boundary_loss}
    L_{\text{boundary}} &= \frac{1}{N} \sum_{i=1}^{N} \left| u(\bm{x}_i) \right|^{p}, \\
    \label{eq:eikonal_loss}
    L_{\text{eikonal}} &= \int_{\Omega} \left| \left\| \nabla u(\bm{x}) \right\| - 1 \right|^p \, \mathrm{d}\bm{x}, \quad p = 1 \, \text{or} \, 2,
\end{align}
where $\Omega \subset \mathbb{R}^d$ is our domain of interest.
The boundary loss encourages the implicit function $u$ to output $0$ at the boundary, and the eikonal loss is based on that observation that if a function $f$ is a distance function, the $L^2$ norm of its gradient $\left\| \nabla f(\bm{x}) \right\| = 1$ holds almost everywhere. Thus the approximation $u$ is encouraged to behave the same.

However, as noted, simply optimizing for the two terms above does not guarantee that $u$ would be a distance function even when satisfied almost everywhere (\Cref{fig:nands,fig:1d}), and the extra regularization previous work added to minimize the area and divergence~\cite{ben2022digs,yang2023steik} could further tamper with the distance approximation quality. 
We derive an additional loss that can be used together with both of the losses above, and analyze our loss theoretically and empirically. 

\begin{figure}[t]
    \centering
    \begin{subfigure}[b]{0.49\linewidth}
        \includegraphics[width=\linewidth]{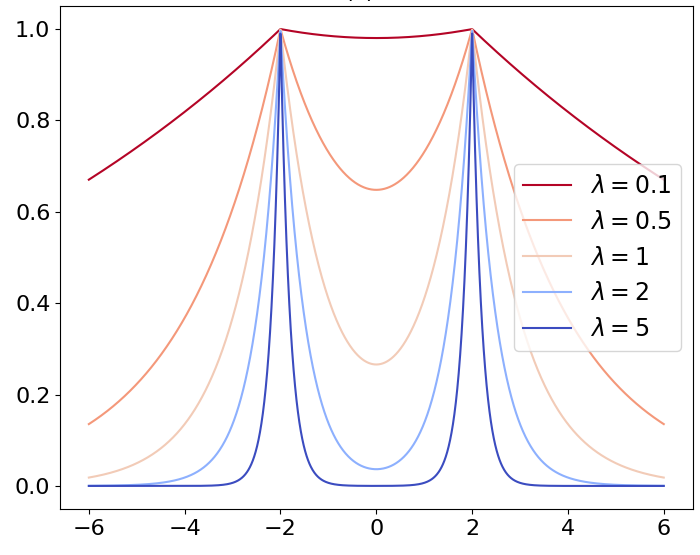}
        \caption{1D Heat Simulation via \Cref{eq:heat}}
    \end{subfigure}
    \begin{subfigure}[b]{0.49\linewidth}
        \includegraphics[width=\linewidth]{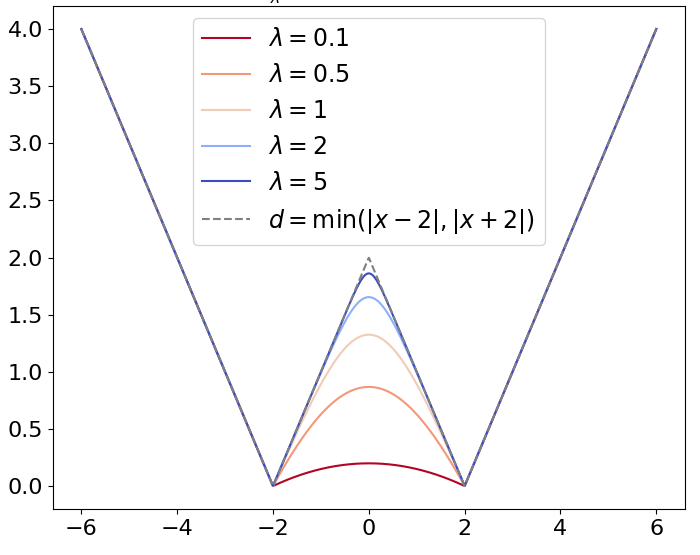}
        \caption{Distance Reconstruction}
    \end{subfigure}
    \begin{subfigure}[b]{0.24\linewidth}
        \includegraphics[width=\linewidth]{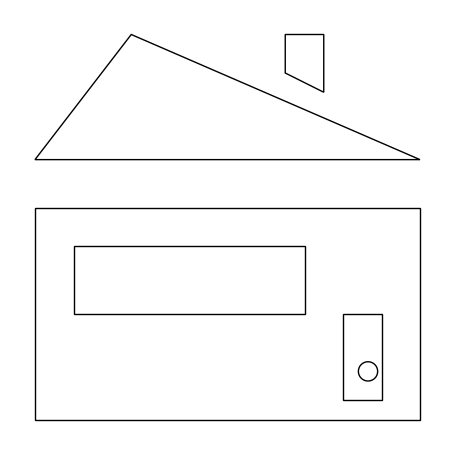}
        \caption{\textsc{House}}
    \end{subfigure}
    \begin{subfigure}[b]{0.24\linewidth}
        \includegraphics[width=\linewidth]{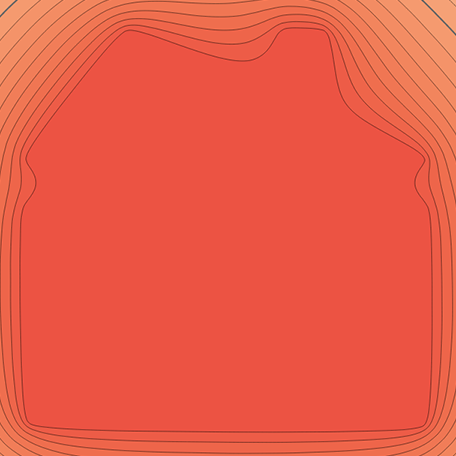}
        \caption{$\lambda = 1$}
    \end{subfigure}
    \begin{subfigure}[b]{0.24\linewidth}
        \includegraphics[width=\linewidth]{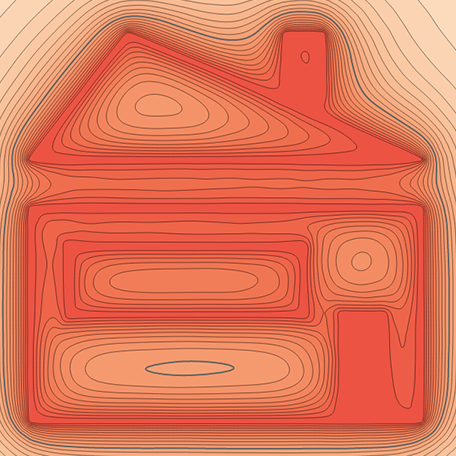}
        \caption{$\lambda = 5$}
    \end{subfigure}
    \begin{subfigure}[b]{0.24\linewidth}
        \includegraphics[width=\linewidth]{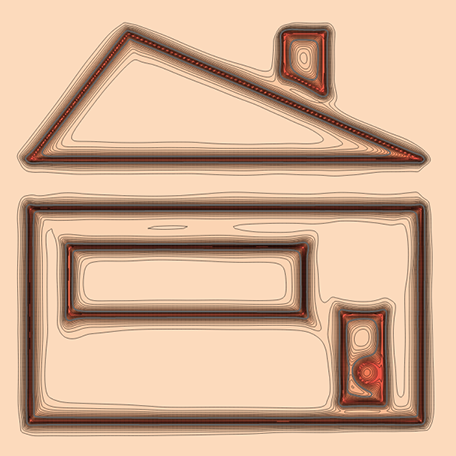}
        \caption{$\lambda = 50$}
    \end{subfigure}
    \vspace{-5pt}
    \caption{
    We show an illustration of the relation between screened Poisson equation and a distance field. The top shows a 1D example, where on the left we show a solution to \Cref{eq:heat} with different absorption $\lambda$, and on the right we show the reconstructed distance $-\frac{1}{\lambda} \ln(h)$ (\Cref{eq:heat_to_distance}). Boundary loss for real boundary points acts as an isothermal heat source, while heat loss diffuses the heat. When $\lambda$ is increased, the heat decays faster, and the error of the reconstructed function approaches 0.
    The bottom shows a 2D example.
    }
    \vspace{-10pt}
    \label{fig:differentlambdas}
\end{figure}

We will be using a key relation between the screen Poisson equation and the distance. 
We first define a screened Poisson equation over a \emph{heat field} $h(\bm{x}) : \mathbb{R}^d \rightarrow \mathbb{R}$, with an absorption coefficient $\lambda$, and set a Dirichlet boundary condition to be $1$ on our target surface $\Gamma$:
\begin{equation}
    \left\{
    \begin{aligned}
        &\nabla^2 h(\bm{x}) - \lambda^2 h(\bm{x}) = 0 \quad \forall \bm{x} \in \mathbb{R}^d \setminus \Gamma\\
        & \lim_{||\bm{x}|| \to\infty} h(\bm{x}) = 0, \quad h(\bm{x}) = 1 \quad \forall \bm{x} \in \Gamma.
    \end{aligned}
    \right.
    \label{eq:heat}
\end{equation}
If a heat field $h_{\lambda}$ is a solution to the partial differential equation above, then the following equation holds~\cite{Varadhan:1967:BFS}:
\begin{equation}
     \lim_{\lambda \to\infty} \frac{1}{\lambda}\ln(h_{\lambda}(\bm{x})) = -d_{\Gamma}(\bm{x}).
     \label{eq:heat_to_distance}
\end{equation}
\Cref{fig:differentlambdas} shows the intuition. The screened Poisson equation simulates heat transfer until an equilibrium is reached, under absorption controlled by the $\lambda$ coefficient.
As we increase the absorption, the reconstructed distance $\frac{1}{\lambda} \ln(h_\lambda)$ would approach the real distance.
In practice, we could only obtain a discrete point set $\Gamma_0$. 
Provided that the neighbor points are close enough, if there are some boundaries connecting them, the boundary condition of \Cref{eq:heat} still holds. 
This can be interpreted as a superset $\Gamma$ replacing $\Gamma_0$, effectively transforming the boundary from a discrete point cloud into a continuous surface, which is required and precisely the goal of the signed distance function reconstruction. 
Obtaining a distance function to the point cloud rather than the interpolated surface is not our objective.
The reconstructed distance can still faithfully represent the distance to $\Gamma$ because the approximation in \Cref{eq:heat_to_distance} still holds. 
Later, we will prove that the measure of $\Gamma$ is bounded by our loss.

%% file: sec/4_method.tex
\section{Method}

In this section, we introduce our method, \MethodName, based on the screened Poisson equation (\Cref{eq:heat}) and derive a theoretical analysis. 
We use the screened Poisson equation to derive a loss we can directly apply to an implicit function $u$, and prove that there is a bound to the solution of the screened Poisson equation to the true distance (\Cref{sec:method-loss}).
We further derive a stability analysis of our loss on both the spatial stability over small perturbations and the temporal convergence (\Cref{sec:method-stability}).
Next, we show that our loss can penalize surfaces with large area (\Cref{sec:method-area}).
Finally, we discuss the relation to prior work, PHASE (\Cref{sec:method-phase}).

\subsection{Screened Poisson Equation Informed Signed Distance Functions Optimization}
\label{sec:method-loss}

Given an implicit function $u$, we want to derive a loss based on the screened Poisson equation (\Cref{eq:heat}) and its relation to the distance (\Cref{eq:heat_to_distance}). We achieve this by applying the following substitution based on \Cref{eq:heat_to_distance}:
\begin{equation}
    h(\bm{x}) = e^{-\lambda |u(\bm{x})|}.
    \label{eq:trans}
\end{equation}
We want to design a loss such that its minimizer satisfies the screened Poisson equation.
This can be achieved by minimizing the energy functional \(\frac{1}{2} \int_{\Omega} \|\nabla h(\bm{x})\|^2  + \lambda^2 h(\bm{x})^2 \, \mathrm{d}\bm{x}\).
After substitution and removing of the constant factor $\lambda$, we obtain our heat loss:
\begin{equation}
    L_{\text{heat}} = \frac{1}{2} \int_{\Omega} e^{-2\lambda |u(\bm{x})|} \left(  \| \nabla u(\bm{x}) \|^{2} + 1 \right) \, \mathrm{d}\bm{x}.
    \label{eq:heat_loss}
\end{equation}
If we take the derivative of $L_{\text{heat}}$ with respect to the heat field $h$, we recover the screened Poisson equation $\nabla^2 h - \lambda^2 h = 0$. 
The boundary condition can be enforced by the standard boundary loss (\Cref{eq:boundary_loss}), by observing that $u(\bm{x})=0$ implies $h(\bm{x})=1$. Together with the eikonal loss, we optimize:
\begin{equation}
    L = w_b L_{\text{boundary}} + w_e L_{\text{eikonal}} + w_h L_{\text{heat}}.
\end{equation}






While prior work~\cite{Varadhan:1967:BFS} has studied the limiting behavior of the absorption coefficient $\lambda$, to our knowledge, the bounds of the approximation and the convergence speed remain unknown. In the supplementary material A.3, we prove that, for a boundary $\Gamma$ formed by a discrete set of points and a small volume around them, the distance approximation obtained from the solution of the screened Poisson equation converges linearly with respect to $\lambda$:
\begin{equation}
      \frac{1}{\lambda} C_{\text{lower}}(\bm{x}) \leqslant \frac{1}{\lambda}\ln(h_{\lambda}(\bm{x})) + d_{\Gamma}(\bm{x}) \leqslant \frac{1}{\lambda} C_{\text{upper}}(\bm{x}).
      \label{eq:distance_bounds}
\end{equation}
Here, both $C_{\text{upper}}(\bm{x})$ and $C_{\text{lower}}(\bm{x})$ are scalar fields independent of $\lambda$. 
The inequalities are tight, which means that one can find a case where the equality holds. 

\begin{figure}[t]
    \centering
    \includegraphics[width=1.0\linewidth]{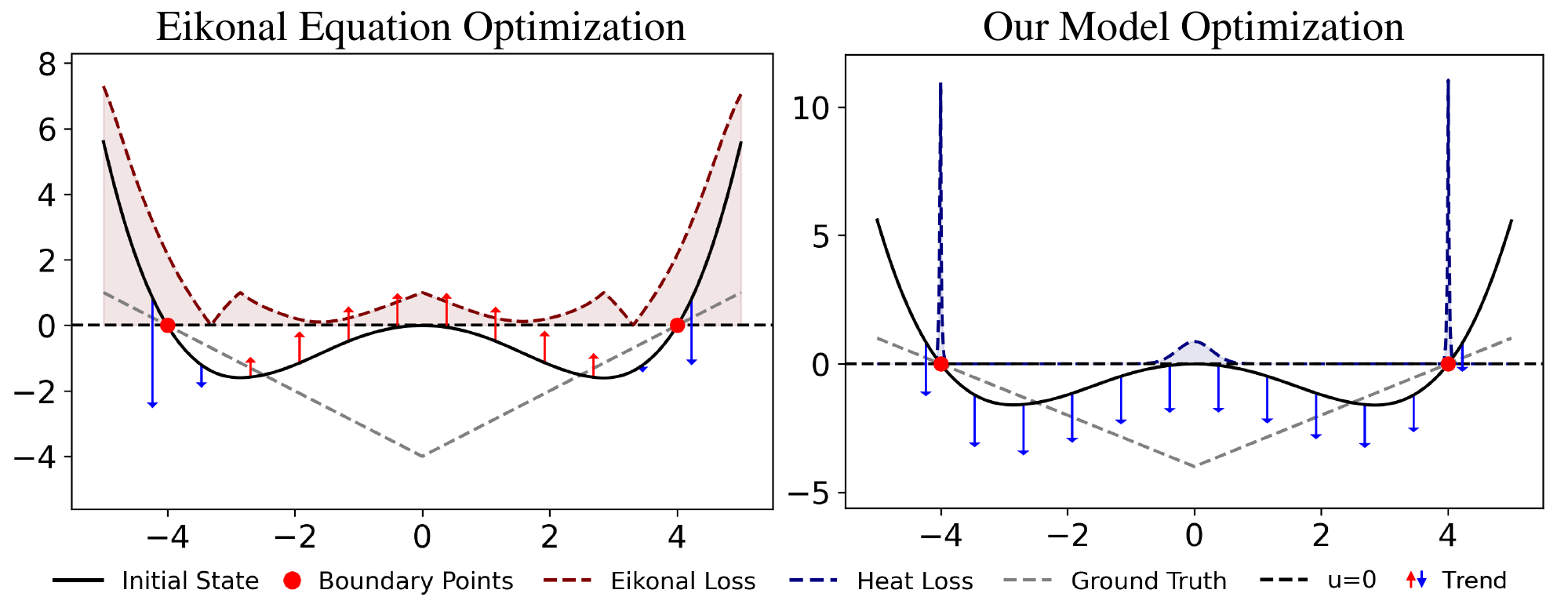}
    \vspace{-15pt}
    \caption{
    We visualize the direction of the changes of the implicit function during one iteration of optimization. 
    Left shows the result with only the eikonal loss and right shows the result after adding our loss.
    Our loss helps to escape the local minimum in \Cref{fig:1d}. The heat diffusion encourages the reconstructed absolute distance to increase monotonically away from the boundary.
    }
    \vspace{-10pt}
    \label{fig:gradcomparison}
\end{figure}
Compared to the eikonal loss (\Cref{eq:eikonal_loss}), where the distance error can be unbounded even when the equation holds almost everywhere, \Cref{eq:distance_bounds} shows that even with a finite absorption $\lambda$, our loss can lead to much more accurate distance. 
In fact, any loss that only depends on the norm of the gradients $\left\| \nabla u \right\|$ is not a sufficient condition of the distance function (\Cref{fig:nands}), due to the discontinuous jumps of the gradient.
Adding our loss (\Cref{eq:heat_loss}) ensures that the absolute value of the implicit function converges to the actual distance.
\Cref{fig:gradcomparison} visualizes the direction of how an implicit function would evolve when using only the eikonal loss and after adding our loss. Our loss ensures that the implicit function evolves towards the actual distance.



\subsection{Stability Analysis}
\label{sec:method-stability}


Our heat loss enjoys other theoretical benefits apart from the convergence to an actual distance. Here, we show that our loss leads to easier optimization. 
We separately analyze \emph{spatial} stability and \emph{temporal} stability, where the spatial stability studies the effect of small perturbation error added to the corresponding partial differential equation, and the \emph{temporal} stability studies the convergence of the gradient flow.

\subsubsection{Spatial Stability}

We show that for the screened Poisson equation, an error introduced to a single point in the solution decays exponentially and does not propagate to infinitely distant positions.
On the contrary, for the eikonal equation, the error can propagate to infinitely far.
The property of the screened Poisson equation ensures that the perturbation introduced during optimization does not lead to a drastically different target and stabilizes the optimization.


First, we show that for the eikonal equation, an additive error introduced to a point can presist along a line segment or a ray.
\begin{proposition}
\label{prop:1}
Given a field $u(\bm{x})$ that satisfies the eikonal equation $\left\|\nabla u \right\| = 1$, we introduce an additive error $u_{0e}$ at the point $\bm{x}_0$ where $u(\bm{x}_0) = u_0$ to obtain a perturbed field $u'$ where $u'(\bm{x}_0) = u_0 + u_{0e}$ and $\left\|\nabla u' \right\| = 1$.
In the perturbed field $u'$, there exists a line segment parameterized by $\bm{x}(s)$ starting from $\bm{x}_0$ and $s \in [0, M]$, where $M > 0$ and can be infinite, such that every point along the line has the same error:
\begin{equation}
    u'(\bm{x}(s)) - u(\bm{x}(s)) = u_{0e} 
\end{equation}
\end{proposition}
The boundary points can contain errors when approximating the true boundaries, and these errors remain constant along the line segment. 
As a result, the propagated error can become significant and potentially detrimental.

Next, we demonstrate that the error field $h_e$ in \Cref{eq:heat} decays exponentially from the perturbed point $\bm{x}_0$.
\begin{proposition}
\label{prop:2}
In 3D space, consider an additive error $h_{0e}$ within a small ball $B(\bm{x_0}, \epsilon)$ around the point $\bm{x_0}$, where the original field $h(\bm{x})$ and the perturbed field $h'(\bm{x}), \, \bm{x} \in \mathbb{R}^n \setminus B$ both satisfy the screened Poisson equation in \Cref{eq:heat}. The resulting error field $h_e = h' - h$ is radially symmetric with respect to $\bm{x}_0$ within some maximum radius $R$, and is given by
\begin{equation}
    h_e(r) = \frac{\epsilon}{r} h_{0e} e^{\lambda (\epsilon - r)}, \, \forall r > \epsilon,
\end{equation}
where $r = \|\bm{x} - \bm{x_0}\|$ is the radial distance from $\bm{x_0}$.
\end{proposition}

The proofs of these two propositions are in the supplementary material A.1 and A.2. 
By comparing the error fields, we observe that the partial differential equation governed by the screened Poisson equation exhibits greater spatial robustness than the one based on the eikonal equation.

\subsubsection{Temporal Stability}

We next demonstrate the temporal stability of the heat loss and show that the dynamical system of the gradient update is stable, that is, the solution of the system converges as the time $t$ approaches infinity.
This is supported by von Neumann stability analysis, similar to Yang et al.'s work~\cite{yang2023steik}.


Yang et al. showed that for the eikonal loss (\Cref{eq:eikonal_loss}), the temporal gradient flow of the optimization is:
\begin{equation}
\begin{aligned}
    \frac{\partial u}{\partial t} &= - \frac{\delta L_{\text{eikonal}}}{\delta u} = \nabla \cdot (\kappa_e \nabla u) \approx \kappa_e \nabla^2 u, \, \text{where} \\
    \kappa_e(\bm{x}) &= \left\{\begin{array}{ll}\frac{1}{|\nabla u(\bm{x})|} \operatorname{sgn}(|\nabla u(\bm{x})| - 1) & p=1 \\ 1-|\nabla u(\bm{x})|^{-1} & p=2\end{array}\right.
\end{aligned}
\end{equation}

They analyze the stability by applying Fourier transform:
\begin{equation}
    \frac{\partial \hat{u}}{\partial t}(t, \omega)=-\kappa_e|\omega|^2 \hat{u}(t, \omega) \Rightarrow \hat{u}(t, \omega) \propto e^{-\kappa_e|\omega|^2 t},
\end{equation}
where $\omega = (\omega_1, \dots, \omega_n)$ denotes the frequency variable and $\hat{u}$ represents the Fourier transform of $u$. In the frequency domain, if the original function converges, the term $\frac{\partial \hat{u}}{\partial t}$ should also converge. However, when $\kappa_e < 0$, this process corresponds to backward diffusion, which diverges and is inherently unstable regardless of the numerical implementation.

We extend the analysis to our heat loss (\Cref{eq:heat_loss}). The gradient flow is
\begin{equation}
    \frac{\partial h}{\partial t} = - \frac{\delta L_{\text{heat}}}{\delta h} = \nabla^2 h - \lambda^2 h
\end{equation}

This is exactly the heat equation, and is known to be stable.
We can further verify the stability by applying Fourier transform to both sides:
\begin{equation}
    \frac{\partial \hat{h}}{\partial t}(t, \omega)=-(|\omega|^2+\lambda^2) \hat{h}(t, \omega) \Rightarrow \hat{h}(t, \omega) \propto e^{- (|\omega|^2+\lambda^2) t}
\end{equation}

This analysis reveals that the original process $h(\bm{x}, t)$ converges over time and remains stable. Due to the constraint $|u| = -\ln(h) / \lambda$, the term $|u(\bm{x}, t)|$ also converges during the optimization process. Since $u$ is continuous, it converges as well. Thus, our method exhibits greater spatial and temporal stability compared to the eikonal equation.


\subsection{Surface Area Regularization}
\label{sec:method-area}

We next show that our heat loss (\Cref{eq:heat_loss}) penalizes large area of the reconstructed surfaces, by showing its relation to the coarea loss used by Pumarola et al.~\cite{pumarola2022visco}. 
Pumarola et al. showed that the loss $\int_{\Omega} e^{-\lambda |u|} \| \nabla u \| \, \mathrm{d}\bm{x}$ approximates the surface area of the zero level set of the implicit function $u$. 
We then form the following inequality:
\begin{equation}
\begin{split}
    \int_{\Omega} e^{-\lambda |u|} \| \nabla u \| \, d\bm{x} &< \int_{\Omega} \sqrt{e^{-2\lambda |u|} ({\| \nabla u \|}^2 + 1)} \, d\bm{x} \\ 
    \leqslant \sqrt{| \Omega|} &\cdot \sqrt{\int_{\Omega} e^{-2\lambda |u|} ({\| \nabla u \|}^2 + 1) \, d\bm{x}}
\end{split}
\label{eq:areaine}
\end{equation}
Therefore, minimizing the right hand side (our heat loss) leads to a reduction in the upper bound of the left hand side.

Furthermore, we can show that the coarea loss is too \emph{aggressive} for our purpose of signed distance field reconstruction.
If we directly minimize the coarea loss (the left hand side in \Cref{eq:areaine}), according to Euler-Lagrange equation, the minimizer would satisfy:
\begin{equation}
    1 - \nabla^2 u = 0,
\end{equation}
which does not relate to the regularization parameter $\lambda$.
Unfortunately, this partial differential equation system formed by the coarea loss above has a unique solution under Dirichlet boundary condition, and the gradient of its solution does not have a unit norm.
This makes the coarea loss unable to accurately shape a signed distance function.
By contrast, our heat loss has bounds for its distance approximation and converges to the true distance as we increase the parameter $\lambda$ (\Cref{eq:distance_bounds}). Another simple way to verify our convergence and the compatibility with the eikonal equation is to rewrite \Cref{eq:heat} in terms of $u$ following \Cref{eq:trans}:
\vspace{-5pt}
\begin{equation}
     {||\nabla u||}^2 - 1 - \frac{\text{sgn(u)}}{\lambda} \nabla^2u = 0.
\end{equation}

\begin{figure}[t]
    \centering
    \includegraphics[width=1.0\linewidth]{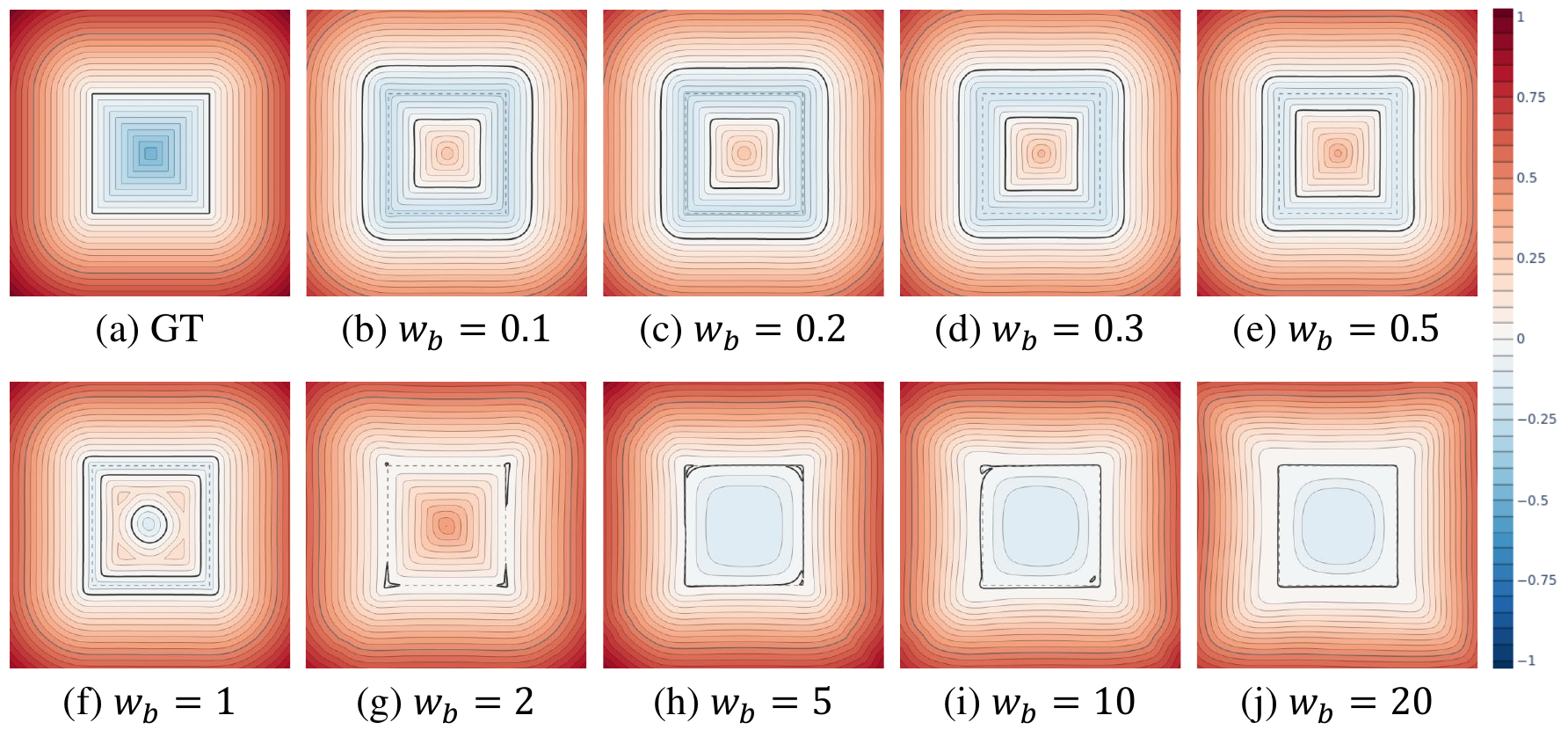}
    \vspace{-20pt}
    \caption{
        We reproduce the PHASE model~\cite{Lipman:2021:PTD} and show the log-transformed occupancy field, i.e., the optimized signed distance function. PHASE recommends boundary weight range $w_{b} \in \left(0.10, 0.32\right)$ based on their $\Gamma$-convergence theory.
        But in practice, only a stronger boundary weight constrains the boundary to follow the input points.
        We found that this phenomenon often occurs and show more examples in the supplementary material.
    }
    \vspace{-10pt}
    \label{fig:phase-boundary-weight}
\end{figure}

\vspace{-10pt}
\subsection{Relation to PHASE~\cite{Lipman:2021:PTD}}
\label{sec:method-phase}

While derived from different mathematical principles and motivations, our method is closely related to the PHASE model proposed by Lipman. It is motivated by smoothly approximating an \emph{occupancy function} $o(\bm{x})$, which is defined as $1$ outside, $-1$ inside, and $0$ on the surface.


In the supplement, we show that there is a close relation between our heat field $h$ and the smoothed occupancy function $o$. 
We also show that our theory and derivation provide additional insights and significant differences in implementation and results.
In particular, PHASE's theory introduces a condition stating that the boundary loss weight $w_b$ should be small, whereas our theory emphasizes the need for a firm boundary condition (\Cref{eq:heat} and \Cref{fig:phase-boundary-weight}).

We further found that optimizing occupancy field in the PHASE model, when used together with the eikonal loss, can lead to unbounded network weight derivatives, impeding the optimization.
Our method avoids this issue.

%% file: sec/5_experiments.tex
\section{Experiments}

We experiment on both 2D and 3D datasets.
We construct a 2D dataset containing 14 shapes, ranging from simple geometric shapes to complex nested structures. 
We also evaluate on two 3D benchmarks: a subset of ShapeNet~\cite{chang2015shapenet} and the Surface Reconstruction Benchmark (SRB)~\cite{berger2013benchmark}. 
Additionally, we generate a dataset with high-genus geometries from Mehta et al.~\cite{Mehta:2022:LST}. 
In all of our experiments, we use only the position information of the point cloud.

We report the Intersection over Union (IoU), Chamfer distance, and Hausdorff distance between reconstructed shapes and ground truth shapes as metrics for surface reconstruction. 
In addition, we report Root Mean Squared Error (RMSE), Mean Absolute Error (MAE), and Symmetric Mean Absolute Percentage Error (SMAPE) as metrics for distance field evaluation. 
We further evaluate non-zero level set quality near the surface, and render the zero level set of the signed distance functions via sphere tracing.

We use a network with 5 hidden layers and 128 hidden channels for our method in all experiments.
The hyperparameters are kept consistent within each dataset. 
We additionally apply importance sampling to our heat loss integral.
Full training details can be found in the supplement.

\subsection{2D Dataset}

An overview of our 2D dataset is attached in the supplementary material.
For 2D experiments, we simply use a linear network and geometric initialization, as in DiGS~\cite{ben2022digs}.
We compare our method with DiGS~\cite{ben2022digs} and StEik~\cite{yang2023steik} with the hyperparameters and training configuration provided in their released code, and report the surface reconstruction metrics and distance field metrics in \Cref{tab:2d_metrics}.
Our method outperforms theirs across all metrics with only 8,192 sample points, whereas both DiGS~\cite{ben2022digs} and StEik~\cite{yang2023steik} use 15,000 sample points based on their default settings.



\subsection{Ablation Study on 2D Dataset}

To more comprehensively compare over different losses, we conduct an ablation study on our 2D dataset using the same network architecture, parameter initialization, and sampling strategy.
The only difference lies in the loss functions used.
Quantitative results are shown in \Cref{tab:ablation}.

The last row corresponds to our \MethodName~model, combining boundary, eikonal, and heat losses. 
Our method reconstructs the correct topology for all 14 shapes, while others struggle with the harder ones. 
It outperforms across nearly all metrics, demonstrating superior reconstruction quality and distance query accuracy.
Although \MethodName~slightly trails SAL~\cite{atzmon2020sal} in RMSE and MAE, it captures the complex topology where SAL fails. 
A visualization is shown in \Cref{fig:fragments}. 
Full qualitative results and detailed experiment parameters are provided in the supplementary material. 
Models relying heavily on area losses~\cite{ben2022digs,yang2023steik} often face a trade-off between preserving original details and eliminating redundant boundaries, which distorts the distance field (\Cref{sec:method-area}). 
Moreover, methods without our heat loss fall in local optimum easily.

\begin{table*}[t]
    \centering
    \begin{minipage}[t]{0.32\textwidth}
            \centering
            \scriptsize
            \begin{tabular}{ lccc }
                \toprule
                    & IoU $\uparrow$ & $d_C$ $\downarrow$ & $d_H$ $\downarrow$ \\
                \midrule
                    DiGS \cite{ben2022digs} & 0.7882 & 0.0055 & 0.1267 \\
                    StEik \cite{yang2023steik} & 0.6620 & 0.0073 & 0.1425 \\
                    Ours & \textbf{0.9870} & \textbf{0.0014} & \textbf{0.0153} \\
                \bottomrule
                \toprule
                    & RMSE $\downarrow$ & MAE $\downarrow$ & SMAPE $\downarrow$ \\
                \midrule
                    DiGS \cite{ben2022digs} & 0.0597 & 0.0315 & 0.3363 \\
                    StEik \cite{yang2023steik} & 0.0725 & 0.0419 & 0.4222 \\
                    Ours & \textbf{0.0199} & \textbf{0.0101} & \textbf{0.0699} \\
                \bottomrule
            \end{tabular}
            \vspace{-5pt}
            \caption{Comparison of 2D reconstruction and distance query.
            $d_C$ and $d_H$ stand for the Chamfer and Hausdorff distances respectively.
            Our method outperforms DiGS~\cite{ben2022digs} and StEik~\cite{yang2023steik} on all metrics.}
            \label{tab:2d_metrics}
    \end{minipage}%
    \hspace{0.2cm}
    \begin{minipage}[t]{0.61\textwidth}
            \vspace{-45.5pt}
            \centering
            \scriptsize
            \begin{tabular}{m{0.15cm} m{0.15cm} m{0.15cm} m{0.15cm} m{0.3cm} m{0.15cm} m{0.3cm} || c c c c c c}
                \toprule
                $\mathcal{L}_B$ & $\mathcal{L}_E$ & $\mathcal{L}_A$ & $\mathcal{L}_D$ & $\mathcal{L}_{DD}$ & $\mathcal{L}_S$ & $\mathcal{L}_H$ & IoU $\uparrow$ & $d_C$ $\downarrow$ & $d_H$ $\downarrow$ & RMSE $\downarrow$ & MAE $\downarrow$ & SMAPE $\downarrow$\\
                \midrule
                \checkmark & \checkmark & & & & &  & 0.8192 & 0.0068 &  0.0712 & 0.0377 & 0.0158 & 0.1315 \\
                \checkmark & & & & & \checkmark & & \textbf{0.8936} & \textbf{0.0029} & \textbf{0.0579} & \textbf{\underline{0.0165}} & \textbf{\underline{0.0081}} & \textbf{0.1205} \\
                \checkmark & \checkmark & & \checkmark & & & & 0.3907 & 0.0843 & 0.5598 & 0.2966 & 0.2388 & 1.7283 \\
                \checkmark & \checkmark & \checkmark & \checkmark & & & & 0.6338 & 0.0566 & 0.4221 & 0.2058 & 0.1616 & 1.0919 \\
                \checkmark & \checkmark & & & \checkmark & & & 0.2938 & 0.0377 & 0.3444 & 0.2990 & 0.2355 & 1.6958 \\
                \checkmark & \checkmark & \checkmark & & \checkmark & & & 0.6037 & 0.0414 & 0.3656 & 0.1304 & 0.0937 & 0.7792 \\
                \checkmark & \checkmark & & & & & \checkmark & \textbf{\underline{0.9851}} & \textbf{\underline{0.0016}} & \textbf{\underline{0.0160}} & \textbf{0.0199} & \textbf{0.0105} & \textbf{\underline{0.0754}} \\
                \bottomrule
            \end{tabular}
            \vspace{-5pt}
            \caption{Ablation study of different loss combinations and metrics. From top to bottom, the loss combinations are from: IGR~\cite{gropp2020implicit}, SAL~\cite{atzmon2020sal}, DiGS w/o area loss, DiGS~\cite{ben2022digs}, StEik w/o loss, StEik~\cite{yang2023steik}, and our \MethodName. From left to right, the losses are boundary loss, eikonal loss, area loss, divergence loss, directional divergence loss, SAL loss, and heat loss. Bold and underlined data: optimal; bold only: suboptimal. Same below.}
            \label{tab:ablation}
    \end{minipage}
    \vspace{-15pt}
\end{table*}

\begin{figure}[t]
    \centering
    \includegraphics[width=0.80\linewidth]{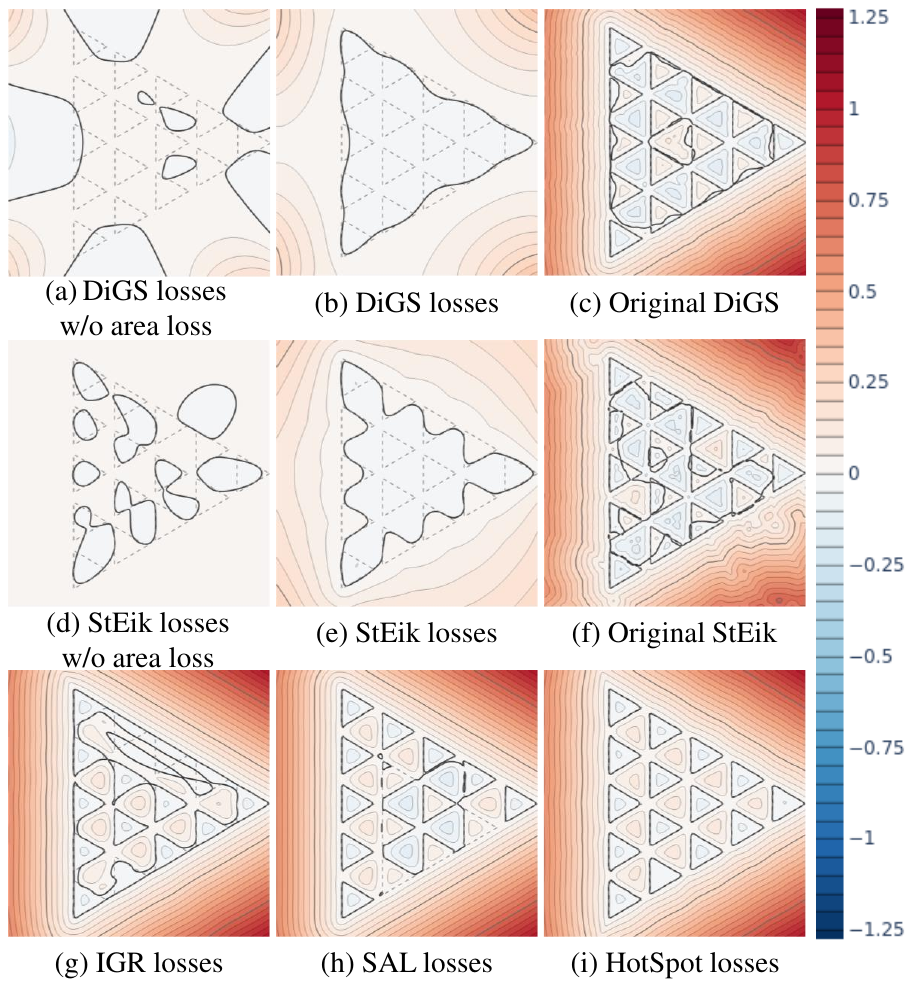}
    \vspace{-5pt}
    \caption{Our heat loss achieves the best reconstruction of triangle fragments. 
    Dashed lines show real boundaries, while the boldest line represents reconstructed ones. 
    (c) and (f) are from the original models by DiGS~\cite{ben2022digs} and StEik~\cite{yang2023steik}; the rest are from our ablation study with different loss combinations. Other combinations fail to remove extra boundaries or accurately capture the topology. More fundamentally, (c), (f), (g), and (h) fall into local optima, leading to incorrect depressions and elevations. 
    }
    \vspace{-17pt}
    \label{fig:fragments}
\end{figure}

\subsection{3D Datasets}

We evaluate our method on a processed subset~\cite{Williams:2021:NSF, mescheder2019occupancy, stutz2020learning} of ShapeNet~\cite{chang2015shapenet}, which provides surface points sampled on preprocessed watertight meshes for 260 shapes across 13 categories. 
To ensure a reasonable spatial density for passing the heat, we scale the point cloud so that 70\% of the points lie within a sphere centered at the origin with a radius of 0.45. 
For a fair comparison, we transform our outcomes back into the coordinate system used in StEik~\cite{yang2023steik} and DiGS~\cite{ben2022digs}. Additionally, we implement a scheduler to gradually increase the absorption parameter $\lambda$, which enhances the representation of level set details.

We present the quantitative results in \Cref{tab:shapenet_metrics} and visual results in \Cref{fig:shapenet_visual}.
More baselines~\cite{ma2020neural,dong2024neurcadrecon} are compared in the supplement.
Our \MethodName~model outperforms current state-of-the-art models across all surface reconstruction metrics, and achieves near-top accuracy in distance queries. 
While SAL~\cite{atzmon2020sal} uses the closest point distance in their loss function, leading to slightly lower total absolute error, their performance drops significantly in relative error and near-surface metrics. 
This suggests that the closest point distance alone is unreliable near the surface.
In contrast, as mentioned at the end of \Cref{sec:bg}, after the interpolation driven by spectral bias \cite{Rahaman:2019:SBN}, our loss remains effective in faithfully capturing the distance to the interpolated surface.

\begin{figure}
    \centering
    \vspace{-15pt}
    \includegraphics[width=1.0\linewidth]{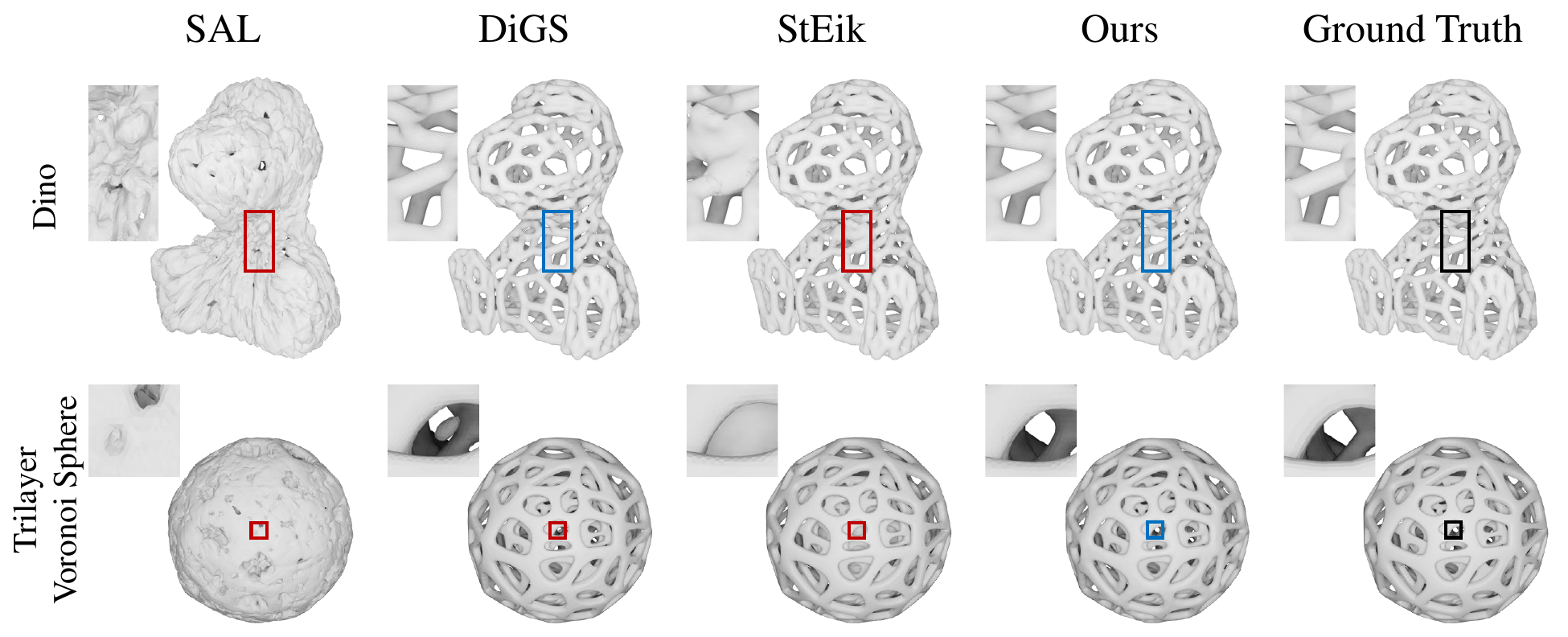}
    \caption{Visual examples on high genus dataset. Our method uses significantly fewer iterations but reconstructs the topology correctly for all shapes, whereas others generate extra boundaries.}
    \label{fig:nie}
    \vspace{-17pt}
\end{figure}

\Cref{fig:teaser} and \Cref{fig:nie} show high-genus surface reconstruction with shapes from Mehta et al.~\cite{Mehta:2022:LST}. Existing methods struggle with the complex topology, while our method accurately reconstructs both the surface and the level set. In the supplement, we present evaluations on other high-genus shapes from Mehta et al. and the SRB dataset.

\begin{figure*}
    \centering
    \includegraphics[width=0.8\linewidth]{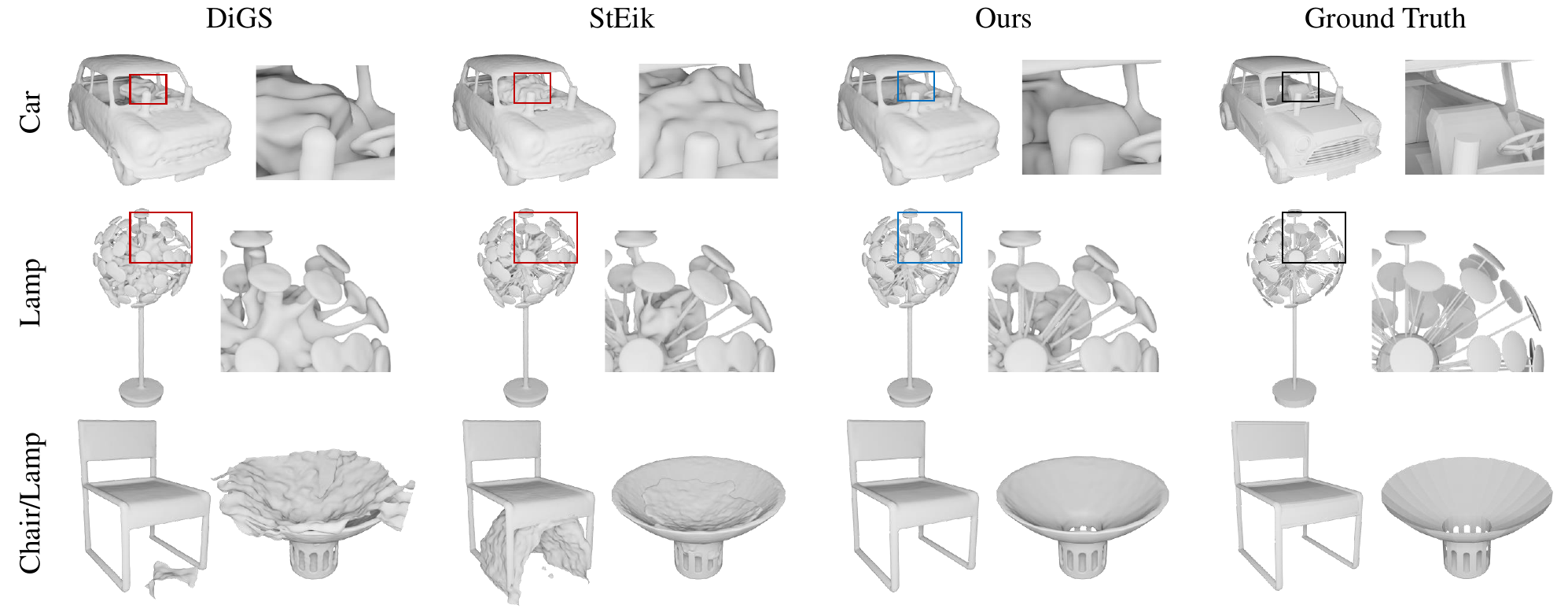}
    \vspace{-10pt}
    \caption{Visual results of DiGS \cite{ben2022digs}, StEik \cite{yang2023steik}, and \MethodName~on ShapeNet \cite{chang2015shapenet}. Previous methods generate extra boundaries and fail to escape local optima, while our method successfully reconstructs the correct topology.}
    \vspace{-5pt}
    \label{fig:shapenet_visual}
\end{figure*}

\begin{table*}
    [ht!]
    \centering
    \scriptsize
    \begin{tabular}{ lccccccccc }
        \toprule
                                                & IoU $\uparrow$  & $d_C$ $\downarrow$ & $d_H$ $\downarrow$ & RMSE $\downarrow$ & MAE $\downarrow$ & SMAPE $\downarrow$ & RMSE$_{0.1}$ & MAE$_{0.1}$ & SMAPE$_{0.1}$ \\
        \midrule
        SAL \cite{atzmon2020sal}                & 0.7400          & 0.0074               & 0.0851                 & \textbf{\underline{0.0251}} & \textbf{\underline{0.0142}} & 0.1344          & 0.0245          & 0.0182          & 0.6848          \\
        SIREN wo/ n \cite{sitzmann2020implicit} & 0.4874          & 0.0051               & 0.0558                 & 0.5009          & 0.4261          & 1.2694          & 0.0513          & 0.0382          & 0.8858          \\
        DiGS \cite{ben2022digs}                 & 0.9636          & \textbf{0.0031}               & 0.0435                 & 0.1194          & 0.0725          & 0.2140          & 0.0152          & \textbf{0.0081}          & \textbf{0.1760}          \\
        StEik \cite{yang2023steik}              & \textbf{0.9641}          & 0.0032               & \textbf{0.0368}                 & 0.0387          & 0.0248          & \textbf{0.0931}          & \textbf{0.0147}          & \textbf{0.0081}          & 0.1770          \\
        Ours                                    & \textbf{\underline{0.9796}} & \textbf{\underline{0.0029}}      & \textbf{\underline{0.0250}}        & \textbf{0.0281}          & \textbf{0.0176}         & \textbf{\underline{0.0540}} & \textbf{\underline{0.0094}} & \textbf{\underline{0.0047}} & \textbf{\underline{0.1206}} \\
        \bottomrule
    \end{tabular}
    \vspace{-5pt}
    \caption{Surface reconstruction metrics on ShapeNet \cite{chang2015shapenet}, where $d_C$ and $d_H$ stand for the Chamfer and Hausdorff distances, RMSE, MAE, and SMAPE are distance field metrics, and RMSE$_{0.1}$, MAE$_{0.1}$, and SMAPE$_{0.1}$ are the same distance field metrics but only for points within distance $0.1$ of the surface. See the supplementary material for detailed results with more baselines.}
    \vspace{-10pt}
    \label{tab:shapenet_metrics}
\end{table*}

\subsection{Sphere Tracing}



\begin{figure}[t]
    \centering
    \includegraphics[width=1.0\linewidth]{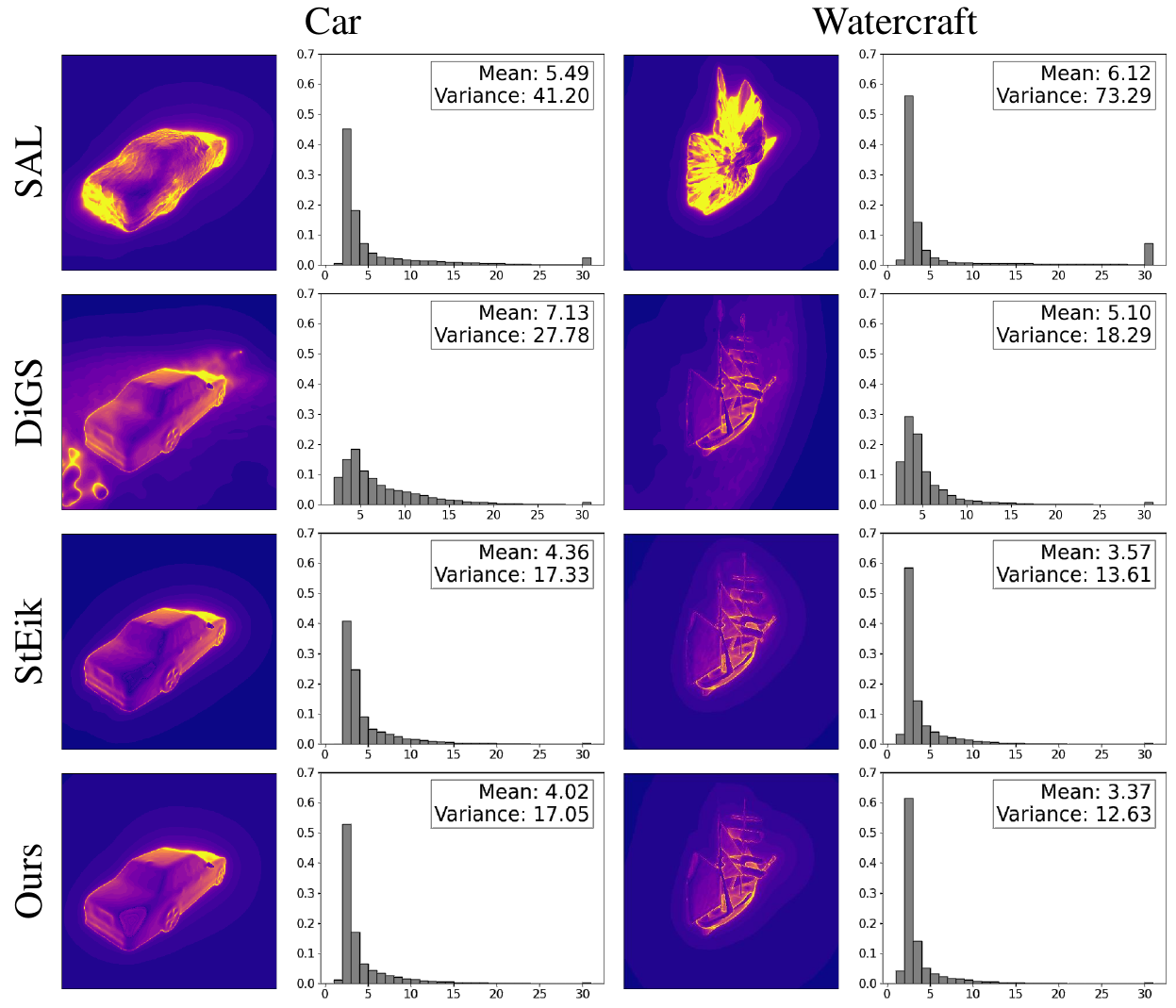}
    \vspace{-20pt}
    \caption{Visualizations and statistics of sphere tracing~\cite{Hart:1996:STG} iteration counts required to locate the surface. In the visual results brighter pixels indicate higher iteration counts. The histograms show the distribution of iterations for all pixels across 10 different camera poses. 
    \MethodName~finds the surface in fewer steps. 
    }
    \vspace{-17pt}
    \label{fig:sphere_tracing}
\end{figure}

\Cref{fig:sphere_tracing} visualizes the number of steps a sphere tracer~\cite{Hart:1996:STG} takes to render the shapes obtained by SAL, DiGS, StEik, and our method.
Our method has the fewest iterations on average, since it provides the most accurate level sets.
More results can be found in the supplement.

%% file: sec/6_discussion.tex
\vspace{-5pt}
\section{Discussion}
\label{sec:disc}

\noindent \textbf{Efficiency.} Our method uses a loss function of only the neural network output and its first-order derivatives, and thus can be more efficient than previous methods which involve second-order derivatives.
We report the runtime of our method and StEik~\cite{yang2023steik} for training a model from ShapeNet~\cite{chang2015shapenet} over 100 iterations using a single NVIDIA A10 GPU with varying network sizes, while keeping all other settings identical, as shown in \cref{tab:efficiency}.

\noindent \textbf{Determining absorption coefficient $\lambda$.} 
The absorption coefficient $\lambda$ has the units of $[\text{L}]^{-1}$, where $\text{L}$ denotes a length unit. Rescaling the space is equivalent to rescaling $\lambda$. 
While a larger $\lambda$ can improve accuracy, it introduces two challenges: 
(1) our heat loss includes the factor $e^{-2\lambda|u|}$, which may become smaller than the precision limit of floating-point numbers, leading to round-off errors;
(2) even if round-off errors are ignored, a large $\lambda |u|$ may reduce the optimizer's ability to effectively shape the field.
Nevertheless, our loss still pushes $u$ away from $0$, preventing unwanted surfaces where $\lambda |u|$ should be large.
We compensate by combining eikonal loss with heat losses. When $\lambda |u|$ is large, the eikonal loss becomes dominant. Our $\lambda$ scheduler also effectively shapes the distant regions.
Spatially-adaptive parameters may also improve the results.



\begin{table}[t]
    \scriptsize
    \vspace{-0.5em}
    \centering
    \begin{tabular}{ c c c c c }
    \toprule
    Structure & \(5 \times 128\) & \(5 \times 256\) & \(8 \times 128\) & \(8 \times 256\) \\
    \midrule
    StEik~\cite{yang2023steik} & 5.65s & 10.62s & 8.78s & 17.00s \\
    Ours & 4.49s & 8.43s & 6.79s & 12.86s \\
    \bottomrule
    \end{tabular}
    \vspace{-0.5em}
    \caption{Runtime for 100 iterations with different network sizes.}
    \label{tab:efficiency}
    \vspace{-16pt}
\end{table}

\noindent \textbf{Necessity of firm boundary condition.}
The heat diffusion of the screened Poisson equation is based on a well-established boundary condition. We enforce the boundary condition through the boundary loss over discrete points, with neural networks interpolating through spectral bias~\cite{Rahaman:2019:SBN}. When the input points are sparse and the absorption coefficient $\lambda$ is high, an overly strong heat loss ($w_b$ being too small) can \emph{tear} the boundary, causing the signed distance to collapse into an unsigned one like \Cref{fig:phase-boundary-weight}. 
Thus, our theory highlights the importance of a high boundary weight $w_b$, while absorption $\lambda$ is adjusted to match the input point density or by scaling the point cloud itself.
Our experiments show that proper parameter setting and rescaling prevent the collapse.
Future research is required for very sparse boundaries, and applications to inverse rendering.

%% file: sec/7_conclusion.tex
\vspace{-5pt}
\section{Conclusion}
\vspace{-3pt}
We propose a new model for neural signed distance function optimization based on the screened Poisson equation. 
We analyze our loss theoretically and show that it is an asymptotically sufficient condition to the true distance, is stable both to a small perturbation and in the temporal dynamics, and penalizes large surface area. 
Our experiments show that we reconstruct both better surfaces and better distance approximations compared to many existing methods, especially on complex and high-genus shapes.

%% file: sec/X_suppl.tex
\clearpage
\setcounter{page}{1}
\maketitlesupplementary
\appendix

\section{Proofs and Derivations}

\subsection{Proof of Proposition \ref{prop:1}.}

Here we analyze the 2D case, but the following derivation naturally generalizes to 3D. 
We denote $\frac{\partial u}{\partial x}$ as $p_1$, $\frac{\partial u}{\partial y}$ as $p_2$, and let $n^2(x, y) = p_1^2+p_2^2$. Following the derivation from Kulyabov et al.~\cite{kulyabov2019numerical}, we can obtain the characteristic equations as follows. For any $a \neq 0$ if any field $u$ satisfies the eikonal equation,
\begin{equation}
    \begin{split}
        \frac{p_1}{a} \frac{\partial p_1}{\partial x} + \frac{p_2}{a} \frac{\partial p_1}{\partial y} = \frac{n}{a} \frac{\partial n}{\partial x} \\
        \frac{p_1}{a} \frac{\partial p_2}{\partial x} + \frac{p_2}{a} \frac{\partial p_2}{\partial y} = \frac{n}{a} \frac{\partial n}{\partial y}
    \end{split}
\end{equation}

This implies that any curve $(x(s), y(s))$ satisfying $\frac{dx}{ds}=\frac{p_1}{a}, \frac{dy}{ds}=\frac{p_2}{a}$ is the characteristic curve of $p_1$ and $p_2$. Let $a = n^2$,
\begin{equation}
    \begin{split}
        \frac{du}{ds} &= u_x \frac{dx}{ds} + u_y \frac{dy}{ds} = p_1 \frac{dx}{ds} + p_2 \frac{dy}{ds} \\
        &= \frac{p_1^2+p_2^2}{a} \equiv 1
    \end{split}
\end{equation}

This equation holds for both $u$ and $u'$ when we take the characteristic curve as

\begin{equation}
    \label{eq:eikon_dir}
    \left\{
    \begin{aligned}
        \frac{dx}{ds} &= \frac{u_x}{u_x^2+u_y^2}, \\
        \frac{dy}{ds} &= \frac{u_y}{u_x^2+u_y^2}
    \end{aligned}
    \right.
\end{equation}

We assume $s \in (0, M)$ ($M > 0$ can be infinity) is a differentiable domain such that every derivative exists and all of these equations hold. Then we prove such a parametric curve is a ray or a segment.

\begin{equation}
    \begin{split}
        \frac{d}{ds} u_x &= u_{xx} \frac{dx}{ds} + u_{xy} \frac{dy}{ds} = u_{xx} \frac{u_x}{n} + u_{xy} \frac{u_y}{n} \\
         &= \frac{1}{2n}(2u_x u_{xx} + 2u_y u_{xy}) \\ 
         &= \frac{1}{2n} \frac{\partial (u_x^2+u_y^2)}{\partial x} \equiv 0
    \end{split}
\end{equation}

Similarly, $u_y$ is also a constant. Then the direction vector in \Cref{eq:eikon_dir} is also a constant vector. Hence, $(x(s), y(s))$ is a ray or a segment. Given $u(x_0, y_0) = u_0$, we can solve this ordinary differential equation and obtain $u(x(s), y(s)) = u_0 + s$. Respectively, $u'(x_0, y_0) = u_0 + u_{0e}$ and $u'(x(s), y(s)) = u_0 + s + u_{0e}$. On this parametic line, we have the following.
\begin{equation}
    u'(\bm{x}(s)) - u(\bm{x}(s)) = u_{0e}.
\end{equation}

\subsection{Proposition \ref{prop:2} proof.}

We first consider the 3D case. First, for the original solution of \Cref{eq:heat}, we denote it as $h$. For the following disturbance equation, we denote its solution as $h_e$:
\begin{equation}
    \left\{
    \begin{aligned}
        &(\nabla^2 - \lambda^2) h_e = 0 \quad \forall \bm{x} \in \mathbb{R}^3 \setminus B(\bm{x_0}, \epsilon) \\
        & \lim_{||\bm{x}|| \to\infty} h_e < \infty, \; h_e(\bm{x}) = h_{0e} \; \forall \bm{x} \in \partial B(\bm{x_0}, \epsilon)
    \end{aligned}
    \right.
    \label{eq:heat_e}
\end{equation}

Then the perturbed $h' = h_e + h$ still satisfies the screened Poisson equation in $\mathbb{R}^d \setminus (B \cup \Gamma)$. Given the spherical condition, we can analytically solve this equation by transforming it to an ordinary differential equation. Knowing that $\nabla ^2 h(r) = r^{-2} \cdot d(r^2 \frac{dh}{dr})/dr$, we have:
\begin{equation}
    \begin{split}
        &\Leftrightarrow \frac{1}{r^2} \frac{d}{dh} \left( r^2 \frac{dh}{dr} \right) - \lambda^2 h = 0 \\
        &\Leftrightarrow \frac{1}{r^2} \left( 2r \frac{dh}{dr} + r^2 \frac{d^2 h}{dr^2} \right) - \lambda^2 h = 0 \\
        &\Leftrightarrow \frac{d^2 h}{dr^2} + \frac{2}{r} \frac{dh}{dr} - \lambda^2 h = 0 \\
        &\Leftrightarrow h(r) = \frac{A e^{-\lambda r}}{r} + \frac{B e^{\lambda r}}{r}, \quad r \geqslant \epsilon
    \end{split}
\end{equation}

Given the boundary conditions, we obtain $A = \epsilon h_{0e} e^{\lambda \epsilon}$ and $ B = 0$. Consequently, the solution is as follows.

\begin{equation}
    \label{eq:3dsolution}
    h_e(r) = \frac{\epsilon}{r} h_{0e} e^{\lambda (\epsilon - r)}, \quad \forall r \geqslant \epsilon
\end{equation}

We can further extend our analysis to 2D. Given $\nabla ^2 h(r) = r^{-1} \cdot d(r \frac{dh}{dr})/dr$, we have another ODE in 2D:

\begin{equation}
    \begin{split}
        &\Leftrightarrow \frac{d^2 h}{dr^2} + \frac{1}{r} \frac{dh}{dr} - \lambda^2 h = 0 \\
        &\Leftrightarrow r^2 \frac{d^2 h}{d r^2} + r \frac{d h}{d r} - \lambda^2 r^2 h =0
        \\
        &\Leftrightarrow z^2 \frac{d^2 h}{d z^2} + z \frac{d h}{d z} - z^2 h =0, \quad z = \lambda r
    \end{split}
\end{equation}

Then we find it becomes a modified Bessel function with the following general solution:

\begin{equation}
    h(z) = A K_0(z) + B I_0(z),
\end{equation}

where $K_0$ represents the modified Bessel function of the second kind and $I_0$ represents the modified Bessel function of the first kind. Given the boundary conditions, we obtain $A = \frac{h_{0e}}{K_0(\lambda \epsilon)}$ and $B = 0$. Then we place $r$ back and have the final solution as follows:

\begin{equation}
    h(r) = \frac{h_{0e}}{K_0(\lambda \epsilon)}K_0(\lambda r)
\end{equation}

It should be noted that $K_0$ converges to $0$ when $\lambda r$ goes to infinity. More rigorously, $K_0(\lambda r)$ and $e^{-\lambda r}/\sqrt{\lambda r}$ are infinitesimals of the same order~\cite{yang2017approximating}. Related property then becomes similar to the 3D case.

\subsection{Convergence speed proof.}

Here we consider the 3D case. For single point $\bm{x}$, given the following screened Poisson equation

\begin{equation}
    \left\{
    \begin{aligned}
        &(\nabla^2 - \lambda^2) h = 0 \quad \forall \bm{x} \in \mathbb{R}^3 \setminus B(\bm{x_0}, \epsilon) \\
        & \lim_{||\bm{x}|| \to\infty} h = 0, \; h(\bm{x}) = e^{-\lambda \epsilon} \; \forall \bm{x} \in \partial B(\bm{x_0}, \epsilon)
    \end{aligned}
    \right.
\end{equation}

Its solution is similar to \Cref{eq:3dsolution}:

\begin{equation}
    \label{eq:onepointsolu}
    h(r) = \frac{\epsilon}{r} e^{- \lambda r}, \quad \forall r \geqslant \epsilon
\end{equation}

Then we denote $S_i = \partial B(\bm{x}_i, \epsilon)$, $h_i$ is the corresponding single point solution, $r_0$ is the minimum distance for any two points from $\Gamma$. When $\epsilon < r_0$ and $\lambda$ is large enough, we can assume that when $i \neq j$, $h_j(S_i) = h_j(\bm{x}_i)$. Given a finite set of boundary points $\Gamma = \{ \bm{x}_1, \bm{x}_2, \dots, \bm{x}_N \}$ that is the input of the signed distance function reconstruction task, we can use a linear combination of single point solutions to obtain a new function satisfying the screened Poisson equation and a new boundary condition that is:

\begin{equation}
    \label{eq:totalspe}
    \left\{
    \begin{aligned}
        &(\nabla^2 - \lambda^2) h = 0 \quad \forall \bm{x} \in \mathbb{R}^3 \setminus \cup_{\forall i} B(\bm{x_i}, \epsilon) \\
        & \lim_{||\bm{x}|| \to\infty} h = 0, \quad h(\bm{x}) = e^{-\lambda \epsilon} \quad \forall \bm{x} \in \cup_{\forall i} S_i
    \end{aligned}
    \right.
\end{equation}

We then set the boundary condition equation to solve the coefficients:

\begin{equation}
    \label{eq:boundaryconditions}
    \begin{pmatrix}
        h_1(S_1) & h_2(S_1) & \cdots & h_N(S_1) \\
        h_1(S_2) & h_2(S_2) & \cdots & h_N(S_2) \\
        \vdots    & \vdots    & \ddots & \vdots    \\
        h_1(S_N) & h_2(S_N) & \cdots & h_N(S_N)
    \end{pmatrix}
    \begin{pmatrix}
        c_1 \\
        c_2 \\
        \vdots \\
        c_N
    \end{pmatrix}
    =
    \begin{pmatrix}
        e^{-\lambda \epsilon} \\
        e^{-\lambda \epsilon} \\
        \vdots \\
        e^{-\lambda \epsilon}
    \end{pmatrix}
\end{equation}

We denote the matrix here as $H \in \mathbb{R}^{N \times N}$. Its every diagonal element is $e^{-\lambda \epsilon}$, which is the largest in that row. This equation has at least one solution $C \in \mathbb{R}^N$ and every $c_i \in (0, 1)$. Finally, $h = (h_1, \dots, h_N) C$ is the solution of \Cref{eq:totalspe}. If we set $r_i = {||\bm{x} - \bm{x}_i||}_2$, then

\begin{equation}
    h(\bm{x}) = \epsilon \sum_{i=1}^n c_i \frac{e^{-\lambda r_i}}{r_i}
\end{equation}

Now we analyze the bound of $\frac{1}{\lambda}\ln(h_{\lambda}(\bm{x})) + d_{\Gamma}(\bm{x})$. First, we prove its upper bound is $\frac{1}{\lambda}[\ln \frac{\epsilon}{d_{\Gamma}(\bm{x})} + \ln(N)]$.

\begin{equation}
\begin{aligned}
\frac{1}{\lambda} \ln h & +d_\Gamma=\frac{1}{\lambda}\left(\ln \epsilon+\ln \sum_{i=1}^n c_i \frac{e^{-\lambda r_i}}{r_i}\right)+d_\Gamma \\
& \leqslant \frac{1}{\lambda}\left(\ln \epsilon+\ln \frac{e^{-\lambda d_\Gamma}}{d_{\Gamma}}+\ln \sum_{i=1}^N c_i\right)+d_\Gamma \\
& \leqslant \frac{1}{\lambda}\left(\ln \frac{\epsilon}{d_\Gamma}+\ln N\right)
\end{aligned}
\end{equation}

Here, the first inequality holds because $d_\Gamma \leqslant r_i$ for all $i$ and every $h_i(r)$ is monotonically decreasing w.r.t. $r$. Then every $c_i < 1$, so we can obtain the second inequality.

Second, we prove its lower bound is $\frac{1}{\lambda}\ln \frac{\epsilon}{d_{\Gamma}(\bm{x})}$.

For any row in \Cref{eq:boundaryconditions}, since we know that the largest element is on the diagonal equal to $e^{-\lambda \epsilon}$, we have

\begin{equation}
    e^{-\lambda \epsilon} = \sum_{i=1}^N c_i h_i(S_j) \leqslant \sum_{i=1}^N c_i h_j(S_j) = e^{-\lambda \epsilon} \sum_{i=1}^N c_i
\end{equation}

Hence, $\sum_{i=1}^N c_i \geqslant 1$. Next, we can set $h_0(r_i) = h_i(\bm{x})$, where $h_0$ is the decay pattern in \Cref{eq:onepointsolu}. $h_0$ is a convex and monotonically decreasing function so that we can apply these inequalities.

\begin{equation}
    \begin{split}
        \frac{1}{\sum_{i=1}^N c_i} h(\bm{x}) = \frac{\sum_{i=1}^N c_i h_0(r_i)}{\sum_{i=1}^N c_i} \\
        \geqslant h_0(\frac{\sum_{i=1}^N c_i r_i}{\sum_{i=1}^N c_i}) \geqslant h_0(d_{\Gamma}(\bm{x}))
    \end{split}
\end{equation}

Given $\sum_{i=1}^N c_i \geqslant 1$, we have $h(\bm{x}) \geqslant \epsilon \frac{e^{-\lambda d_\Gamma}}{d_\Gamma}$. By basic transformations of this, we obtain the whole inequality as follows:

\begin{equation}
    \frac{1}{\lambda}\ln \frac{\epsilon}{d_\Gamma} \leqslant d_\Gamma - |u_\lambda| \leqslant \frac{1}{\lambda}\left(\ln \frac{\epsilon}{d_\Gamma}+\ln N\right)
\end{equation}

where $|u_\lambda| = -\frac{1}{\lambda} \ln h_\lambda $. It should also be noted that $\ln \frac{\epsilon}{d_\Gamma(\bm{x})}$ is a constant scalar field so that when $\bm{x}$ is fixed and $\lambda$ increases to infinity, $d_\Gamma - |u_\lambda|$ is first-order infinitesimal.

\section{Experiments}

In all of our experiments, we use $p = 1$ for \Cref{eq:boundary_loss} and \Cref{eq:eikonal_loss}.

\subsection{1D Verification}

We can easily identify which loss function is not a sufficient constraint by observing its input. As discussed in the main text, if a loss function depends solely on the first-order derivative of the SDF or any higher-order derivatives, it is inherently insufficient as a constraint and cannot exclude other non-SDF solutions.

Our experiments also demonstrate that implementations optimizing SDFs with the closest point~\cite{atzmon2020sal, ma2020neural, Atzmon:2021:SSA} may struggle when making queries near the interpolated surface. As mentioned at the end of \Cref{sec:bg}, after the neural network interpolates these points due to spectral bias~\cite{Rahaman:2019:SBN}, our loss remains effective in faithfully capturing the distance to the interpolated surface.

For other cases, we can easily test whether a constraint is insufficient with simple 1D verifications.

We found one possible candidate to be a sufficient constraint from Marschner et al.~\cite{marschner2023constructive} with following loss:

\begin{equation}
    L_{\text{CP}} = \int_{\Omega} |u[\bm{x}- u(\bm{x})\nabla u(\bm{x})]|^{2}d\bm{x}
\end{equation}

The loss is designed to optimize constructive solid geometries and formulated using both neural network outputs and their gradients, ensuring that when taking a step in the direction indicated by the SDF, the new position should have a distance value close to zero.

\begin{figure}
    \centering
    \includegraphics[width=1\linewidth]{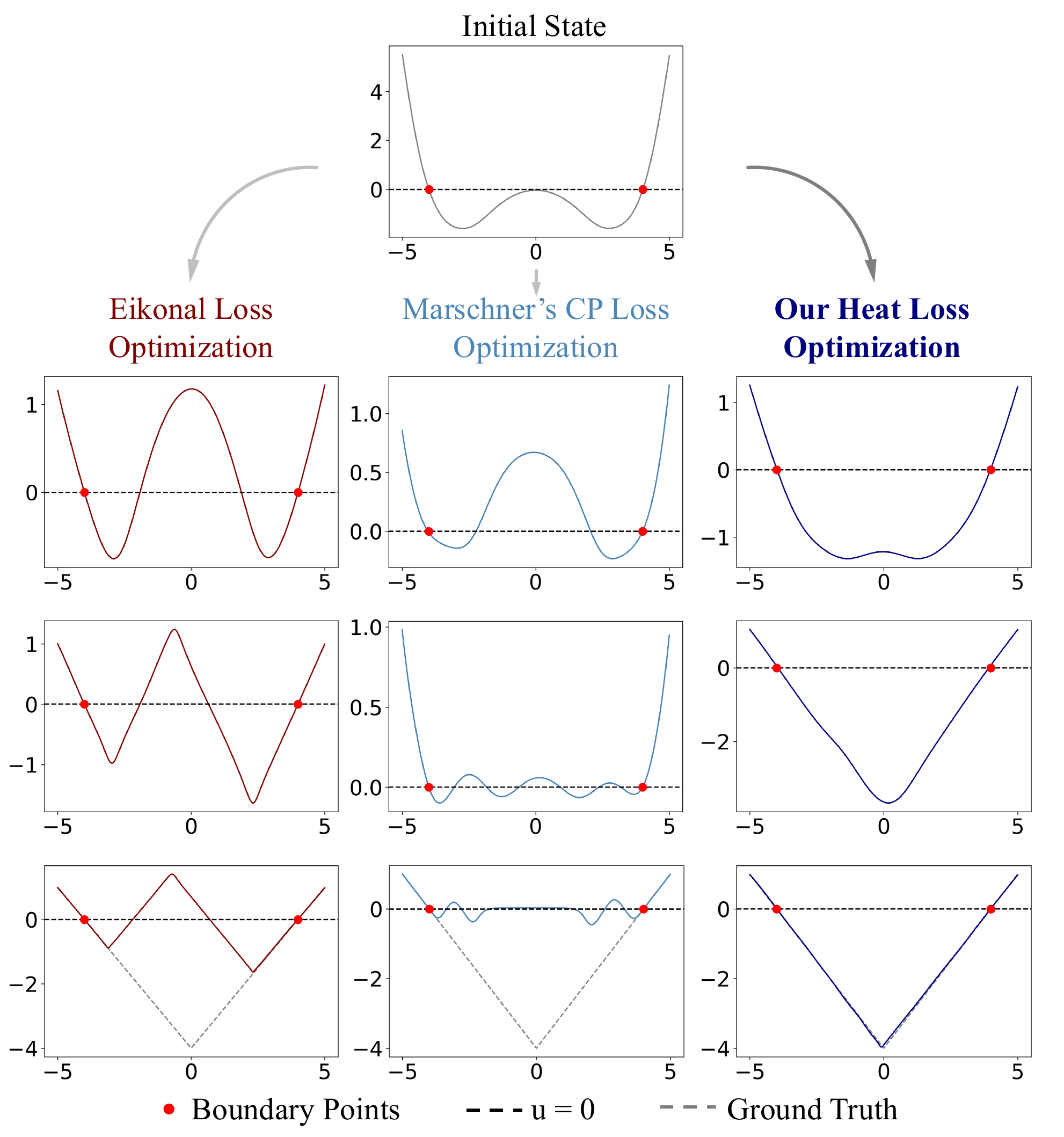}
    \caption{Marschner et al.~\cite{marschner2023constructive} proposed a CP loss which is a possible candidate to be a sufficient constraint. However, our 1D experiment illustrates that it is incapable of converging to the actual signed distance function (dashed line), even when the output minimizes this loss almost everywhere.
    The $x$ axis is the domain and the $y$ axis shows the output of the implicit function. 
    The middle two rows display the intermediate states of the optimization, while the bottom row presents the final results.}
    \label{fig:1dverification}
\end{figure}

However, our experiment in \Cref{fig:1dverification} reveals that it cannot exclude non-SDF solutions like $u(\bm{x})\equiv 0$, confirming that it is not a sufficient constraint.

\subsection{2D Dataset}

\begin{figure*}
    \centering
    \includegraphics[width=1.0\linewidth]{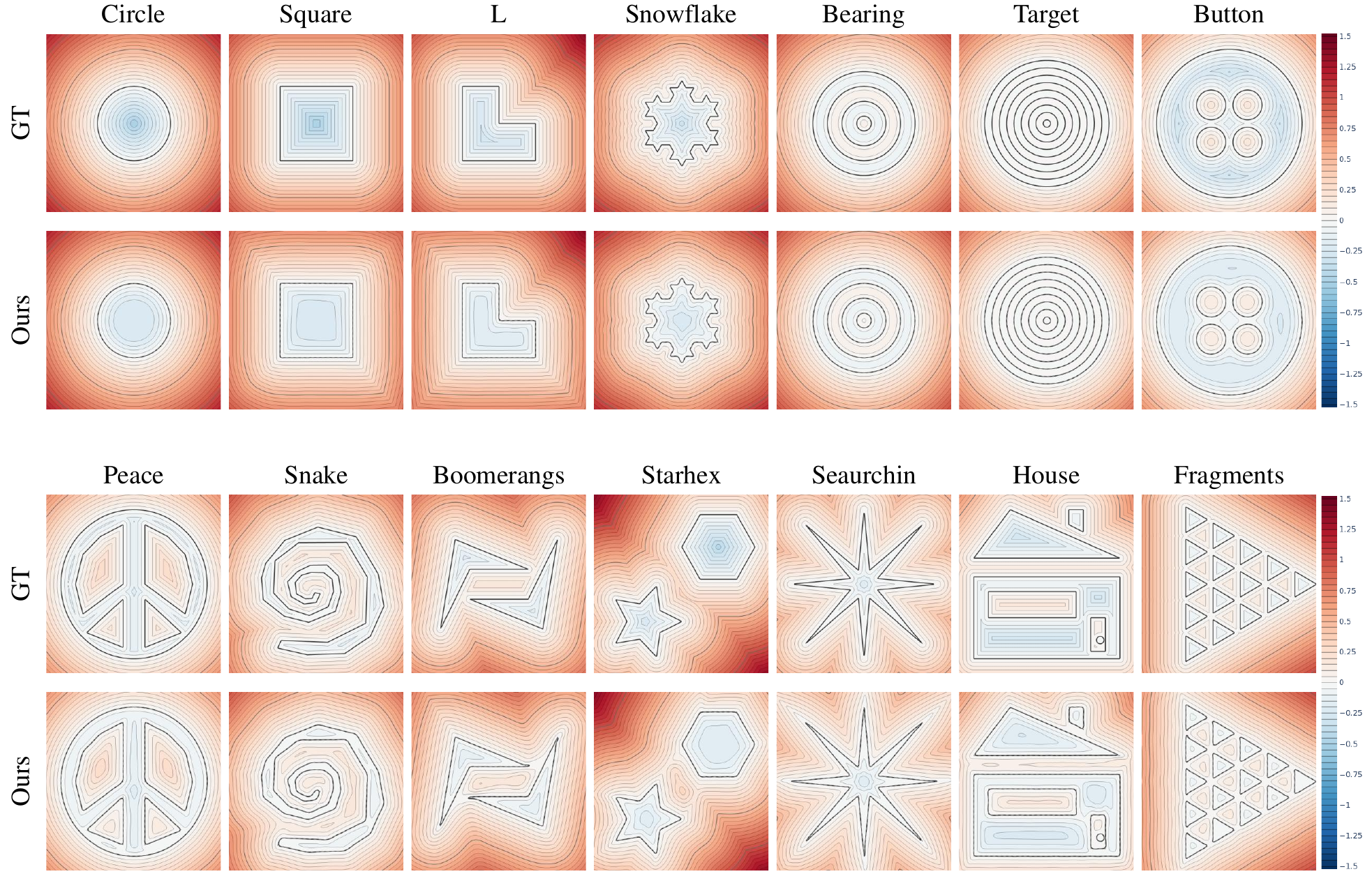}
    \caption{This figure illustrates all 14 vector shapes in our 2D dataset. Each visualization lies within the range ${[-1.2, 1.2]}^2$. Spanning a spectrum from simple to complex, the dataset encompasses various topologies, smooth and irregular boundaries, as well as configurations with single and multiple objects. Our \MethodName~model accurately reconstructs all curves while preserving their correct topologies.}
    \label{fig:2dgt}
\end{figure*}

\Cref{fig:2dgt} provides an overview of the ground truths in the 2D dataset. Starting with simple shapes from DiGS~\cite{ben2022digs} and StEik~\cite{yang2023steik}, we extended the dataset to include 14 shapes. We generate a total of 150,000 points along the boundaries in a single vector image. For each iteration, we randomly select 10\% of the generated points to compute the boundary loss.

In our first 2D experiments (\Cref{tab:2d_distance_metrics}), we adopt the original hyperparameters and settings from DiGS~\cite{ben2022digs} and StEik~\cite{yang2023steik} as our baselines, making only one modification: extending the training iterations from 10k to 20k to better learn complex shapes.

In our setup, we compute the heat loss using the importance sampling method, which combines a mixture distribution of 1:1 uniform samples in $[-1.5, 1.5]^2$ and Gaussian samples with an isotropic $\sigma=0.5$. For each iteration, we use 4,096 points for both the uniform and Gaussian samples, whereas DiGS and StEik generate 15,000 points to compute their derivative-based loss. We fix $\lambda$ in the heat loss while employing two schedulers: one for the heat loss and another for the eikonal loss. Towards the end of training, the heat loss is gradually weakened, and the eikonal loss is strengthened. This strategy aligns with the approach used in DiGS and StEik. Our results are presented in \Cref{fig:2dgt} as well. However, as shown in the ablation study (\Cref{tab:ablation}), removing the scheduler does not result in significant differences.

\begin{figure*}
    \centering
    \includegraphics[width=1.0\linewidth]{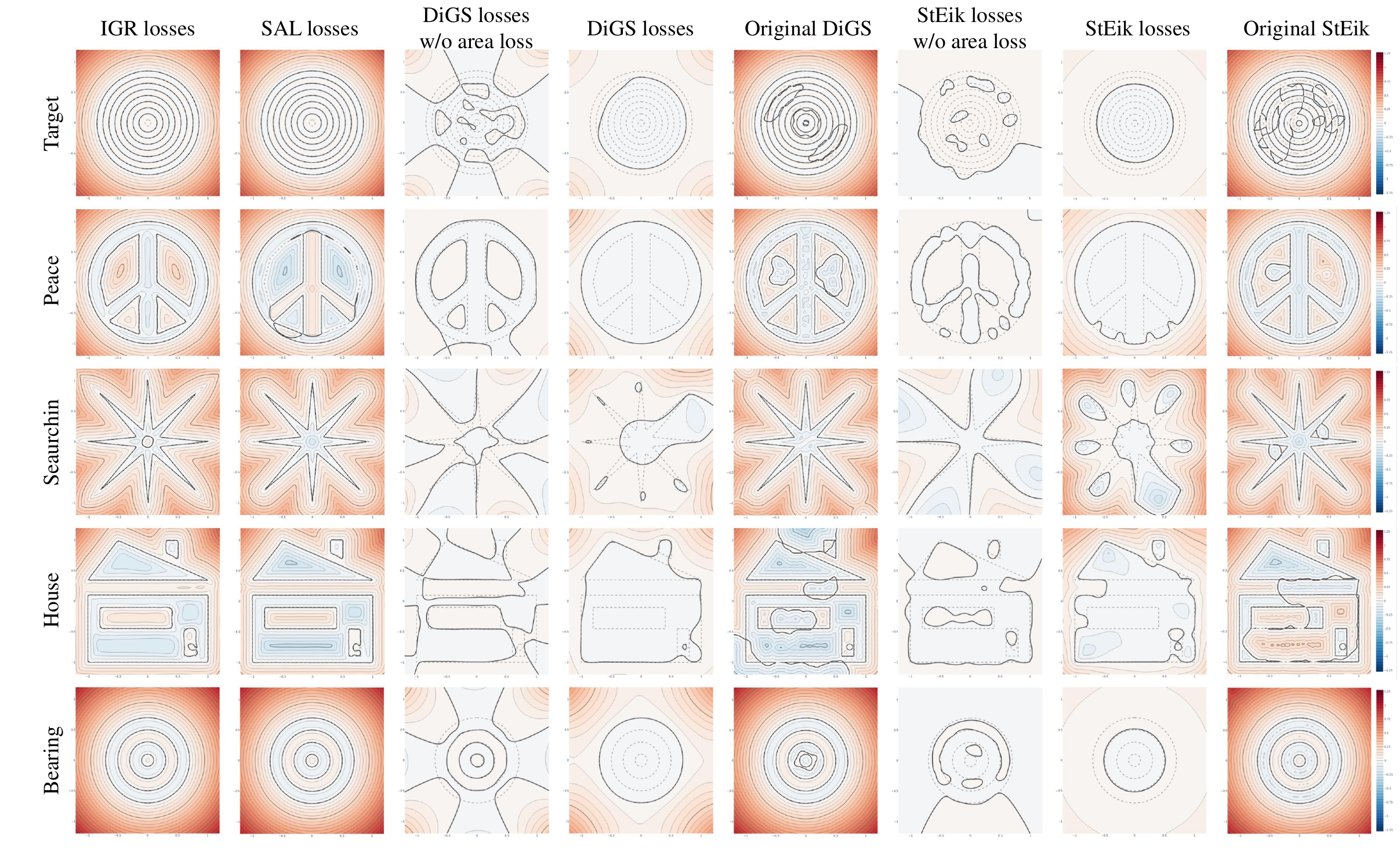}
    \caption{This figure demonstrates how the baselines fail with complex shapes. The difference between the fourth and fifth columns (from left to right), as well as between the seventh and eighth columns, is that the former one is generated in the ablation study, while the other is generated by the original model. Dashed lines represent the true boundaries, while the boldest line indicates the reconstructed boundaries.}
    \label{fig:2dbaselines}
\end{figure*}

We also visualize the outcomes of the baseline methods in \Cref{fig:2dbaselines}. In the ablation study, the boundary loss coefficient remains constant and identical across all experiments, and the same scheduler is applied to the eikonal loss. We also adopted the original loss weight ratios from the experiments of DiGS~\cite{ben2022digs} in the fourth column and StEik~\cite{yang2023steik} in the seventh column. All losses, except for the eikonal loss, are applied without a scheduler. The outputs of the original DiGS and StEik models are also shown in the fifth and eighth columns, respectively.

All other models make some errors in topology. Even when their topologies are correct, details like the Target and House shapes in the first two columns are missing. When using a relatively smaller learning rate in the ablation study for DiGS and StEik, the outputs become overly flat, despite maintaining the same ratio among the loss weights from their paper. This can be interpreted as a limitation of their derivative-based losses, which, as a corollary of the eikonal equation, only serve as a necessary condition for the equation, encouraging the condition $|\nabla u| = c$, where $c$ can be any constant. With a small learning rate, their losses trap the outputs at $|\nabla u| = 0$.

\begin{figure*}
    \centering
    \includegraphics[width=1.0\linewidth]{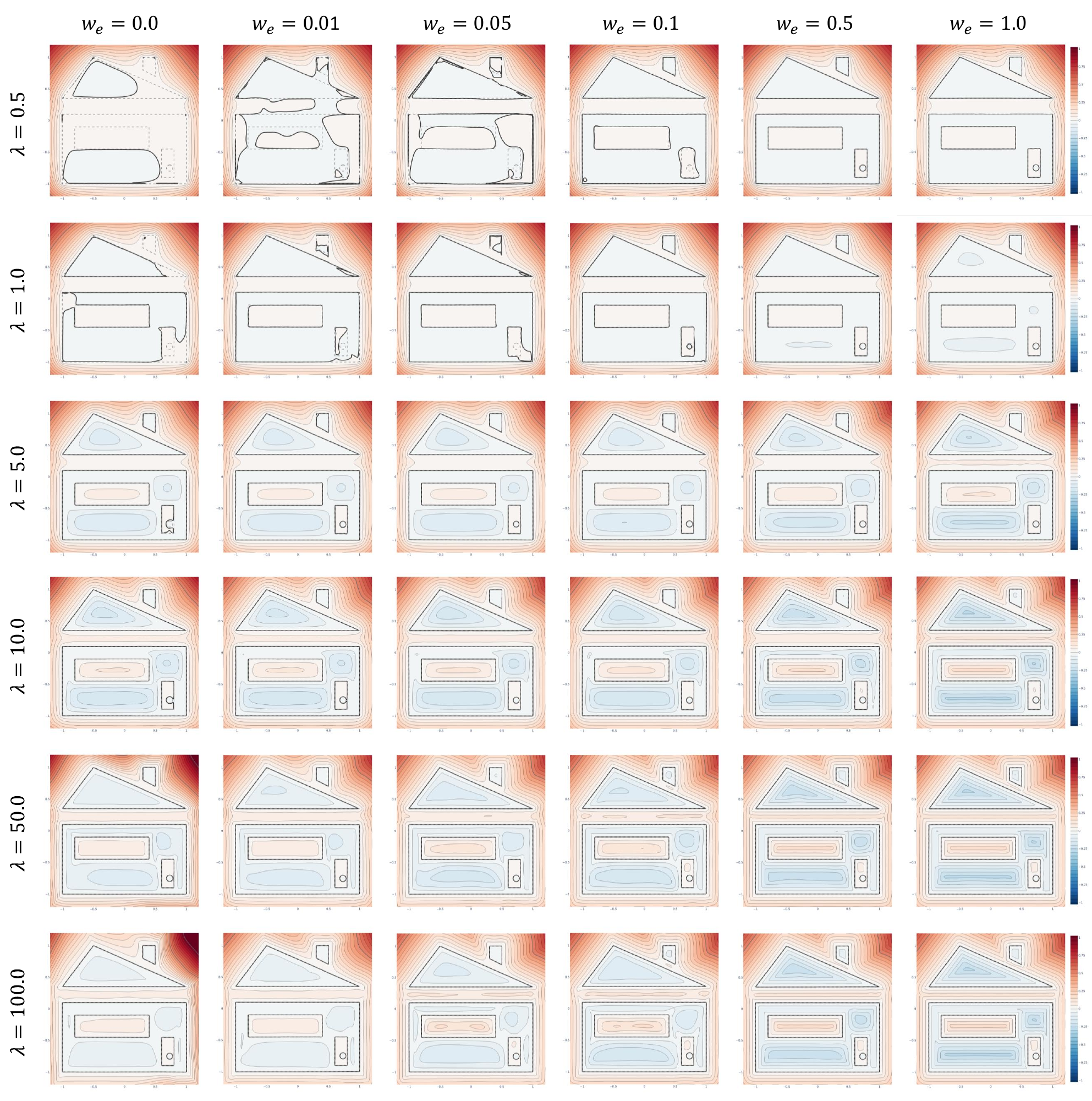}
    \caption{Signed distance function reconstructions of the vector image House with varying values of $\lambda$ and $w_e$.}
    \label{fig:differentlambdas_supp}
\end{figure*}

In our framework, we analyze the influence of different values of $\lambda$ and various coefficients of the eikonal loss, with visualizations presented in \Cref{fig:differentlambdas_supp}. This experiment replicates the settings from the ablation study, except for the values of $\lambda$, $w_e$, and the iteration number, with all schedulers removed. To ensure full convergence, we extend the training iterations from 20k to 200k.

From \Cref{fig:differentlambdas} and \Cref{fig:differentlambdas_supp}, we observe the influence of different $\lambda$. When $\lambda$ is very small, heat from the boundaries diffuses to distant regions, causing $h \approx 1$ almost everywhere in the test region. As a result, $u \approx 0$ across the domain, leading to an overly flat signed distance function with many extra boundaries. As $\lambda$ increases, the outputs become more regular. Notably, even without the eikonal loss, setting $\lambda = 10$ yields outputs that are more regular than several baselines in \Cref{fig:2dbaselines}. However, as $\lambda$ continues to increase, the factor $e^{-\lambda |u|}$ in the loss computation diminishes rapidly, especially where $|u|$ is large initially or grows during training. This makes optimization without the eikonal loss increasingly challenging and less effective. For instance, in the $\lambda = 50$ and $\lambda = 100$ subfigures with $w_e = 0$, the values in the upper-right region remain greater than $1.0$ even after 200k iterations. Although our neural network provides some output values in these regions, they are significantly larger than the ground truth.

Incorporating the eikonal loss stabilizes the training process and promotes a more regular field. When $\lambda$ is small, the approximation from \Cref{eq:heat_to_distance} is weakly achieved, but the eikonal loss helps regulate the output and prevents it from becoming overly flat. When $\lambda$ is large, the eikonal loss dominates in regions where $\lambda|u|$ is substantial. However, if the eikonal loss is overly strong, extra boundaries and local optima may re-emerge.

This does not mean that users must carefully balance the hyperparameters $w_e$, $w_h$, and $\lambda$.
We have found effective ways to choose them.
In our subsequent experiments, the schedulers for $\lambda$ and $w_e$ ensure robust shaping capabilities across various shapes and distance ranges in the ShapeNet dataset~\cite{chang2015shapenet}. Mimicking a real annealing process, we gradually increase $\lambda$ and $w_e$, allowing the heat loss to shape and stretch most regions first, helping the optimization escape local optima. As the heat field cools due to the increasing absorption coefficient $\lambda$, the eikonal loss maintains the stretching and assumes control in remote regions.

\begin{table*}
    [ht!]
    \centering
    \begin{tabular}{ l|c|c|c|c|c|c|c|c|c }
        \hline
            & \multicolumn{3}{c}{IoU $\uparrow$} & \multicolumn{3}{c}{Chamfer Distance $\downarrow$} & \multicolumn{3}{c}{Hausdorff Distance $\downarrow$} \\
        \hline
            & mean & median & std & mean & median & std & mean & median & std \\
            \hline
            DiGS \cite{ben2022digs} & 0.7882 & 0.9359 & 0.2803 & 0.0055 & 0.0037 & 0.0046 & 0.1267 & 0.1350 & 0.1088 \\
            StEik \cite{yang2023steik} & 0.6620 & 0.7305 & 0.3224 & 0.0073 & 0.0051 & 0.0068 & 0.1425 & 0.1654 & 0.1146 \\
            Ours & \textbf{0.9870} & \textbf{0.9888} & \textbf{0.0083} & \textbf{0.0014} & \textbf{0.0013} & \textbf{0.0003} & \textbf{0.0153} & \textbf{0.0150} & \textbf{0.0100} \\
        \hline
    \end{tabular}
    \caption{Comparison of 2D dataset reconstruction metrics.}
    \label{tab:2d_surface_metrics}
\end{table*}

\begin{table*}
    [ht!]
    \centering
    \begin{tabular}{ l|c|c|c|c|c|c|c|c|c }
        \hline
            & \multicolumn{3}{c}{RMSE $\downarrow$} & \multicolumn{3}{c}{MAE $\downarrow$} & \multicolumn{3}{c}{SMAPE $\downarrow$} \\
        \hline
            & mean & median & std & mean & median & std & mean & median & std \\
            \hline
            DiGS \cite{ben2022digs} & 0.0597 & 0.0504 & 0.0511 & 0.0315 & 0.0253 & 0.0351 & 0.3363 & 0.2355 & 0.3414 \\
            StEik \cite{yang2023steik} & 0.0725 & 0.0335 & 0.0903 & 0.0419 & 0.0108 & 0.0574 & 0.4222 & 0.2223 & 0.4409 \\
            Ours & \textbf{0.0199} & \textbf{0.0189} & \textbf{0.0130} & \textbf{0.0101} & \textbf{0.0072} & \textbf{0.0060} & \textbf{0.0699} & \textbf{0.0693} & \textbf{0.0226} \\
        \hline
    \end{tabular}
    \caption{Comparison of 2D dataset distance queries.}
    \label{tab:2d_distance_metrics}
\end{table*}

\subsection{ShapeNet}

\begin{figure*}
    \centering
    \includegraphics[width=1.0\linewidth]{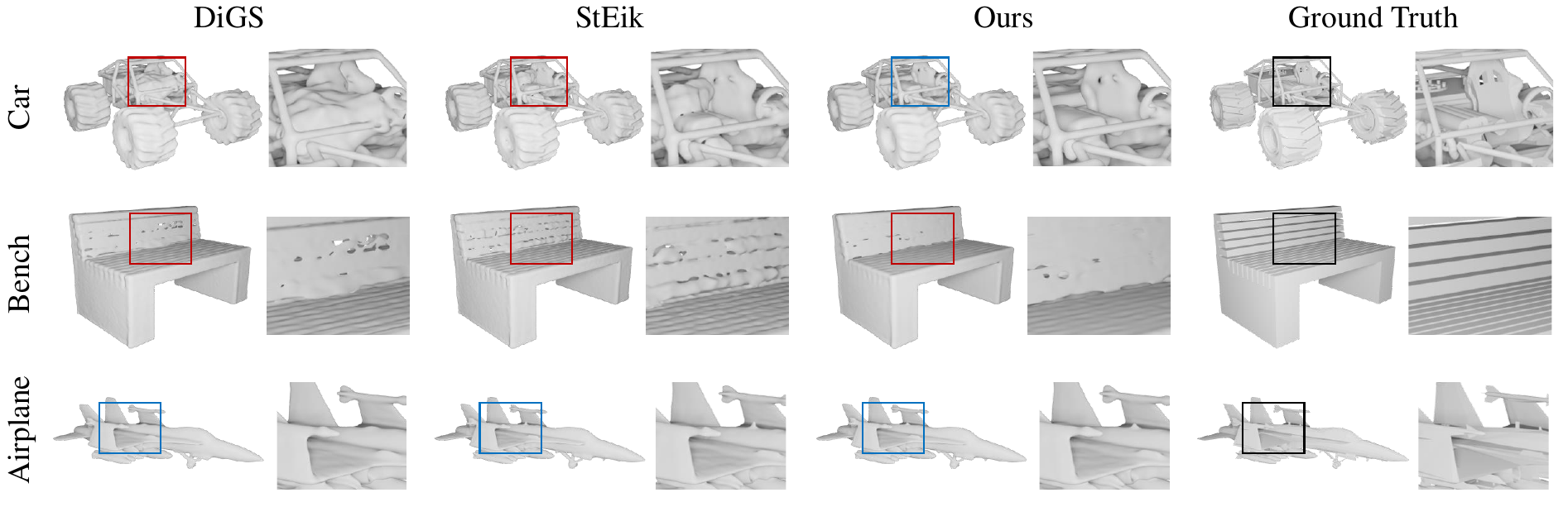}
    \caption{Additional results from the ShapeNet dataset~\cite{chang2015shapenet}. We present the bench, where none of the methods achieves a satisfactory surface reconstruction, and the airplane, where all methods successfully capture detailed structures.}
    \label{fig:enter-label}
\end{figure*}

We show full metrics for the ShapeNet dataset in Tables \ref{tab:shapenet_surface_metrics}, \ref{tab:shapenet_distance_metrics}, and \ref{tab:shapenet_distance_metrics_near_surface}. We compare our method with the state-of-the-art methods, including SAL~\cite{atzmon2020sal}, SIREN without normalization~\cite{sitzmann2020implicit}, Neural-Singular-Hessian~\cite{wang2023neural}, Neural-Pull~\cite{ma2020neural}, DiGS~\cite{ben2022digs}, and StEik~\cite{yang2023steik}. Our method outperforms all other methods in terms of surface reconstruction metrics, including IoU, Chamfer distance, and Hausdorff distance. In terms of distance query metrics, our method achieves near-top performance across RMSE, MAE, and SMAPE.

Notably, in near-surface regions, our approach outperforms all other methods, including SAL~\cite{atzmon2020sal} and Neural-Singular-Hessian~\cite{wang2023neural}, across these metrics. As discussed in the main text, while approximating the SDF using the closest point information may be reasonably accurate for distant regions, it proves inadequate for near-surface regions. This limitation is particularly critical for sphere tracing—one of the most important applications of SDF—as it heavily relies on accurate field values in near-surface regions due to the high density of queries in these areas.

In contrast, as we introduced at the end of \Cref{sec:bg}, after the neural network interpolates these points due to spectral bias~\cite{Rahaman:2019:SBN}, our loss remains effective in faithfully capturing the distance to the interpolated surface. Our experiments on sphere tracing further substantiate this claim, demonstrating the infeasibility of approximating the SDF using only the closest point information and highlighting the superiority of our proposed model.

\begin{table*}
    [ht!]
    \centering
    \begin{tabular}{ lccccccccc }
        \hline
            & \multicolumn{3}{c}{IoU $\uparrow$} & \multicolumn{3}{c}{Chamfer Distance $\downarrow$} & \multicolumn{3}{c}{Hausdorff Distance $\downarrow$} \\
        \hline
            & mean & median & std & mean & median & std & mean & median & std \\
            \hline
            SAL \cite{atzmon2020sal} & 0.7400 & 0.7796 & 0.2231 & 0.0074 & 0.0065 & 0.0048 & 0.0851 & 0.0732 & 0.0590 \\
            SIREN wo/ n \cite{sitzmann2020implicit} & 0.4874&0.4832&0.4030 & 0.0051&0.0038&0.0036 & 0.0558&0.0408&0.0511  \\
            NSH \cite{wang2023neural} & 0.7293&0.9285&0.3538 & 0.0036&0.0033&0.0017 & \textbf{0.0324}&0.0231&\textbf{\underline{0.0286}} \\
            Neural-Pull \cite{ma2020neural} & 0.7300&0.7972&0.2229 & 0.0114&0.0073&0.0161 & 0.1334&0.0840&0.1287 \\
            DiGS \cite{ben2022digs} & 0.9636&0.9831& \textbf{0.0903} & \textbf{0.0031}&\textbf{\underline{0.0028}}&\textbf{0.0016} & 0.0435&\textbf{0.0168}&0.0590 \\
            StEik \cite{yang2023steik} & \textbf{0.9641} & \textbf{\underline{0.9848}} & 0.1052 & 0.0032 & \textbf{\underline{0.0028}} & 0.0028 & 0.0368 & 0.0172 & 0.0552 \\
            Ours & \textbf{\underline{0.9796}} & \textbf{0.9842} & \textbf{\underline{0.0203}} & \textbf{\underline{0.0029}} & \textbf{\underline{0.0028}} & \textbf{\underline{0.0012}} & \textbf{\underline{0.0250}} & \textbf{\underline{0.0153}} & \textbf{0.0360} \\
        \hline
    \end{tabular}
    \caption{Surface reconstruction metrics on ShapeNet \cite{chang2015shapenet}. Bold and underlined data: optimal; bold only: suboptimal. Same below.}
    \label{tab:shapenet_surface_metrics}
\end{table*}

\begin{table*}
    [ht!]
    \centering
    \begin{tabular}{ lccccccccc }
        \hline
            & \multicolumn{3}{c}{RMSE $\downarrow$} & \multicolumn{3}{c}{MAE $\downarrow$} & \multicolumn{3}{c}{SMAPE $\downarrow$} \\
        \hline
            & mean & median & std & mean & median & std & mean & median & std \\
            \hline
            SAL \cite{atzmon2020sal} & \textbf{0.0251} & \textbf{0.0197} & 0.0270 & \textbf{0.0142} & \textbf{0.0116} & 0.0108 & 0.1344 & 0.1064 & 0.1032 \\
            SIREN wo/ n \cite{sitzmann2020implicit} & 0.5009 & 0.4842 & 0.1769 & 0.4261 & 0.4027 & 0.1811 & 1.2694 & 0.9859 & 0.5195\\
            NSH \cite{wang2023neural} & 0.3486&0.2469&0.2566 & 0.2891&0.1780&0.2508 & 0.7386&0.4897&0.5424 \\
            Neural-Pull \cite{ma2020neural} & \textbf{\underline{0.0093}}&\textbf{\underline{0.0067}}&\textbf{\underline{0.0121}} & \textbf{\underline{0.0060}}&\textbf{\underline{0.0053}}&\textbf{\underline{0.0042}} & \textbf{0.0673}&\textbf{\underline{0.0505}}&\textbf{0.0612} \\
            DiGS \cite{ben2022digs} & 0.1194&0.1107&0.0597 & 0.0725&0.0644&0.0423 & 0.2140&0.2162&0.0935 \\
            StEik \cite{yang2023steik} & 0.0387 & 0.0338 & 0.0229 & 0.0248 & 0.0222 & 0.0142 & 0.0931 & 0.0843 & 0.0748 \\
            Ours & 0.0281 & 0.0259 & \textbf{0.0136} & 0.0176 & 0.0160 & \textbf{0.0082} & \textbf{\underline{0.0540}} & \textbf{0.0514} & \textbf{\underline{0.0243}} \\
        \hline
    \end{tabular}
    \caption{Overall distance query metrics on ShapeNet \cite{chang2015shapenet}.}
    \label{tab:shapenet_distance_metrics}
\end{table*}

\begin{table*}
    [ht!]
    \centering
    \begin{tabular}{ lccccccccc }
        \hline
            & \multicolumn{3}{c}{RMSE near surface $\downarrow$} & \multicolumn{3}{c}{MAE near surface $\downarrow$} & \multicolumn{3}{c}{SMAPE near surface $\downarrow$} \\
        \hline
            & mean & median & std & mean & median & std & mean & median & std \\
            \hline
            SAL \cite{atzmon2020sal} & 0.0245 & 0.0252 & 0.0075 & 0.0182 & 0.0189 & 0.0059 & 0.6848 & 0.6890 & 0.2488 \\
            SIREN wo/ n \cite{sitzmann2020implicit} & 0.0513&0.0401&0.0404 & 0.0382&0.0206&0.0351 & 0.8858&0.5406&0.7483 \\
            NSH \cite{wang2023neural} & 0.0876&0.0798&0.0383 & 0.0686&0.0601&0.0342 & 0.8830&0.7254&0.4744 \\
            Neural-Pull \cite{ma2020neural} & \textbf{0.0123}&\textbf{0.0098}&0.0088 & 0.0087&0.0076&0.0047 & 0.3856&0.3459&0.1439 \\
            DiGS \cite{ben2022digs} & 0.0152&0.0135&0.0081 & \textbf{0.0081}&\textbf{0.0074}&\textbf{0.0037} & \textbf{0.1760}&\textbf{0.1657}&\textbf{0.0660} \\
            StEik \cite{yang2023steik} & 0.0147 & 0.0130 & \textbf{0.0070} & \textbf{0.0081} & \textbf{0.0074} & 0.0041 & 0.1770 & 0.1664 & 0.0859 \\
            Ours & \textbf{\underline{0.0094}} & \textbf{\underline{0.0078}} & \textbf{\underline{0.0049}} & \textbf{\underline{0.0047}} & \textbf{\underline{0.0042}} & \textbf{\underline{0.0020}} & \textbf{\underline{0.1206}} & \textbf{\underline{0.1163}} & \textbf{\underline{0.0313}} \\
        \hline
    \end{tabular}
    \caption{Distance function query metrics for near surface region on ShapeNet \cite{chang2015shapenet}.}
    \label{tab:shapenet_distance_metrics_near_surface}
\end{table*}

Our test region is defined in the same way as in DiGS and StEik. Both methods rescaled the circumscribed sphere centered at the geometric center of a point cloud to a unit sphere and designated the circumscribed cube $[-1,1]^3$ as their test region. We evaluate our results within the exact same coordinate system. During training, however, we rescale the point cloud to achieve an adaptive $\lambda$ described in the main text. 

To compute the IoU and ground truth distances, we utilized the Occupancy Network~\cite{mescheder2019occupancy} and a point cloud completion model~\cite{stutz2020learning}. A dense grid was generated within $[-1,1]^3$ to evaluate the metrics. For near-surface queries, we filtered points with a ground truth distance smaller than 0.1 to compute the accuracy, as presented in \Cref{tab:shapenet_distance_metrics_near_surface}. Our method demonstrates a remarkable lead, reducing losses by more than one-third compared to the second-best model.

\begin{figure*}
    \centering
    \includegraphics[width=1.0\linewidth]{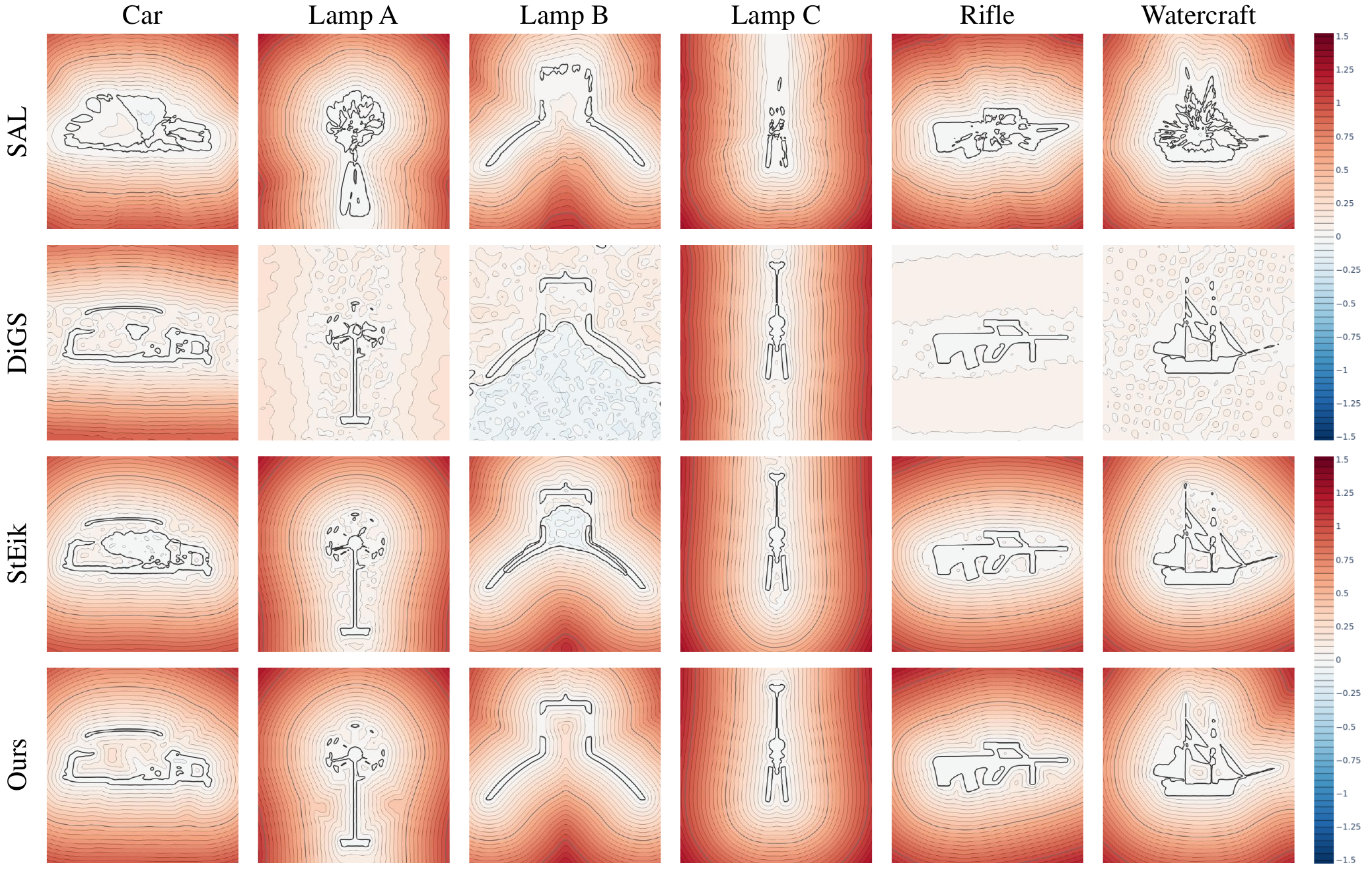}

    \caption{Sectional views of ShapeNet \cite{chang2015shapenet} signed distance function reconstruction results. Car, Lamp A, and Lamp B are shown in \Cref{fig:shapenet_visual}, while Lamp C, Rifle, and Watercraft are presented in \Cref{fig:sphere_tracing} and \Cref{fig:large_st}. By cross-verifying with them, \MethodName~achieves reconstructions with fewer extra boundaries, more regularized level sets, and accurate topologies.}
    \label{fig:shapenetsectional}
\end{figure*}

To illustrate the improved quality of our level sets, we also provide sectional views from different models for comparison in \Cref{fig:shapenetsectional}.

Our level set near the surface is smoother and more regular, offering significant advantages for downstream tasks such as sphere tracing.

\subsection{Complex Topology Reconstruction}

\begin{figure*}
    \centering
    \includegraphics[width=1.0\linewidth]{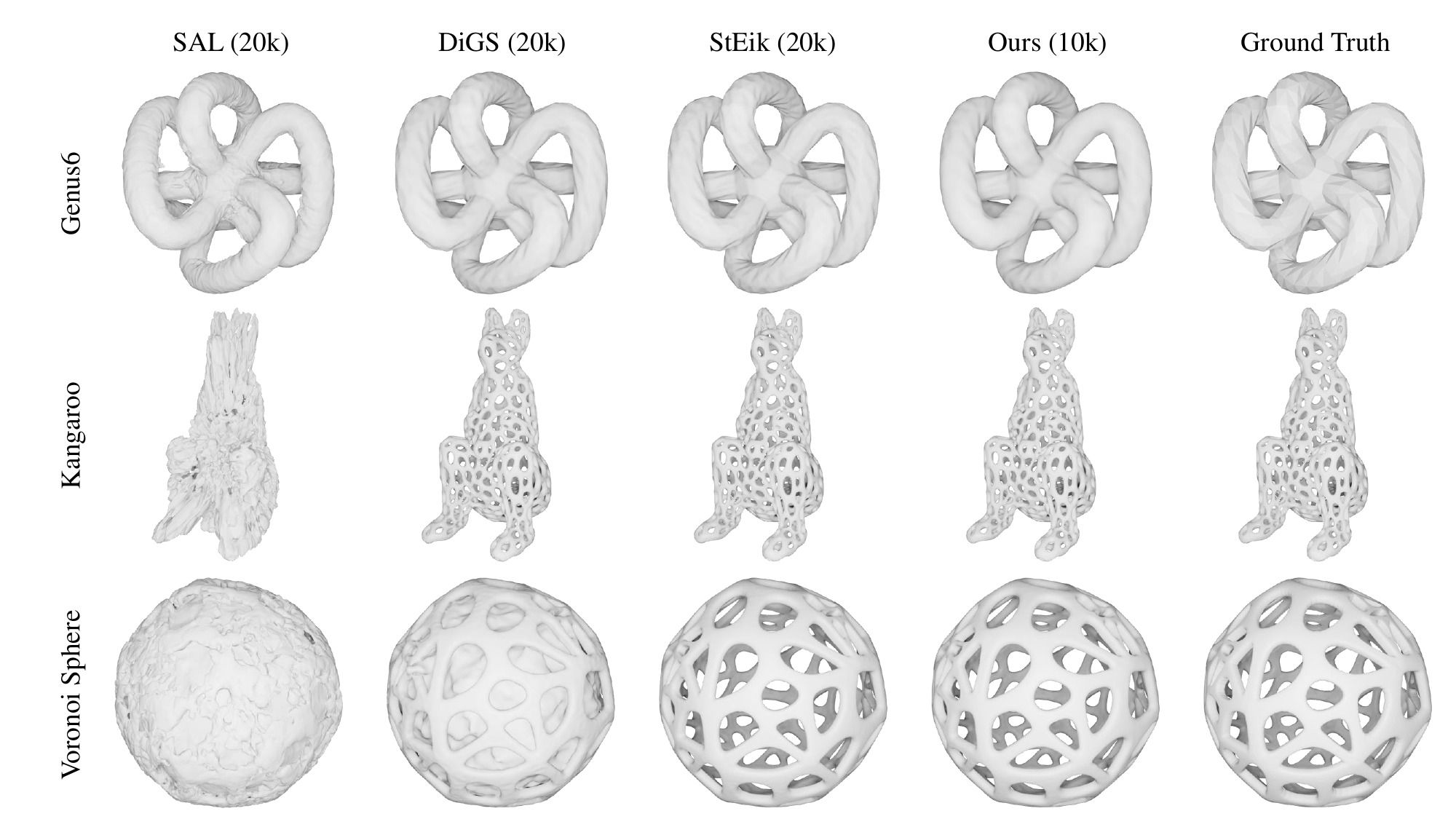}
    \caption{Visualizations of Genus6, Kangaroo, and Voronoi Sphere from Mehta et al~\cite{Mehta:2022:LST}. We show the number of iterations used in training in the parentheses. We achieve excellent results with only half the iterations, successfully capturing the correct topologies.}
    \label{fig:origin_nie}
\end{figure*}

\begin{figure*}
    \centering
    \includegraphics[width=1.0\linewidth]{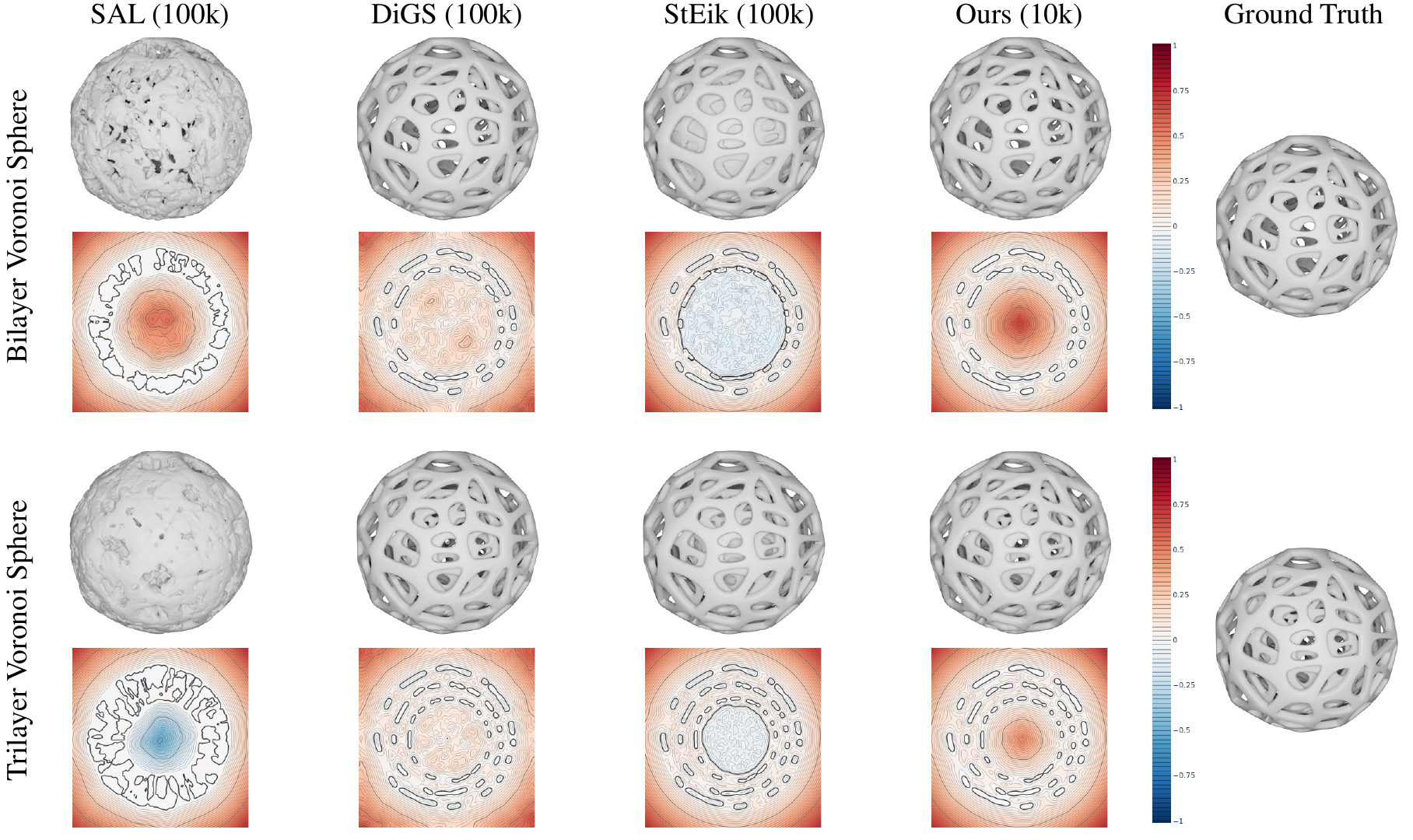}
    \caption{Visualizations of bilayer and trilayer Voronoi Spheres. Beyond achieving accurate topology reconstructions, the sectional views reveal that our level set is the only one free from chaotic and noisy interiors, ensuring meaningful representations.}
    \label{fig:multilayer}
\end{figure*}

We adopt five high-genus geometries from Mehta et al.~\cite{Mehta:2022:LST} and generate 3D point clouds for them, including Bunny, Genus6, VSphere, Dino, and Kangaroo.
Additionally, we use the mesh of VSphere to create bilayer and trilayer VSphere, resulting in a total of seven shapes with complex topologies.
We compare our method with SAL~\cite{atzmon2020sal}, DiGS~\cite{ben2022digs}, and StEik~\cite{yang2023steik} on these shapes, presenting the visual results in \Cref{fig:teaser}, \Cref{fig:nie}, \Cref{fig:origin_nie}, and \Cref{fig:multilayer}.
Our method runs for 10k iterations.
To ensure sufficient convergence and minimize extra boundaries, we run the other methods for 20k iterations on Bunny, Genus6, VSphere, Dino, and Kangaroo, and for 100k iterations on the bilayer and trilayer VSphere.
Despite the increased iterations, the other methods fail to reconstruct the correct topology and generate extra boundaries, whereas our method successfully reconstructs the correct topology for all shapes.

\subsection{Surface Reconstruction Benchmark (SRB)}

SRB consists of 5 noisy scans, each containing point cloud and normal data. 
We compare our method against the current state-of-the-art methods on this benchmark without using the normal data.
The results are presented in \Cref{tab:srb_surface_metrics}, where we report the Chamfer ($d_C$) and Hausdorff ($d_H$) distances between the reconstructed meshes and the ground truth meshes. 

Additionally, we provide the corresponding one-sided distances ($d_{\vec{C}}$ and $d_{\vec{H}}$) between the reconstructed meshes and the input noisy point cloud. It is worth noting that one-sided distances are used here to maintain consistency with the historical choice of previous methods.

\begin{figure}
    \centering
    \includegraphics[width=1.0\linewidth]{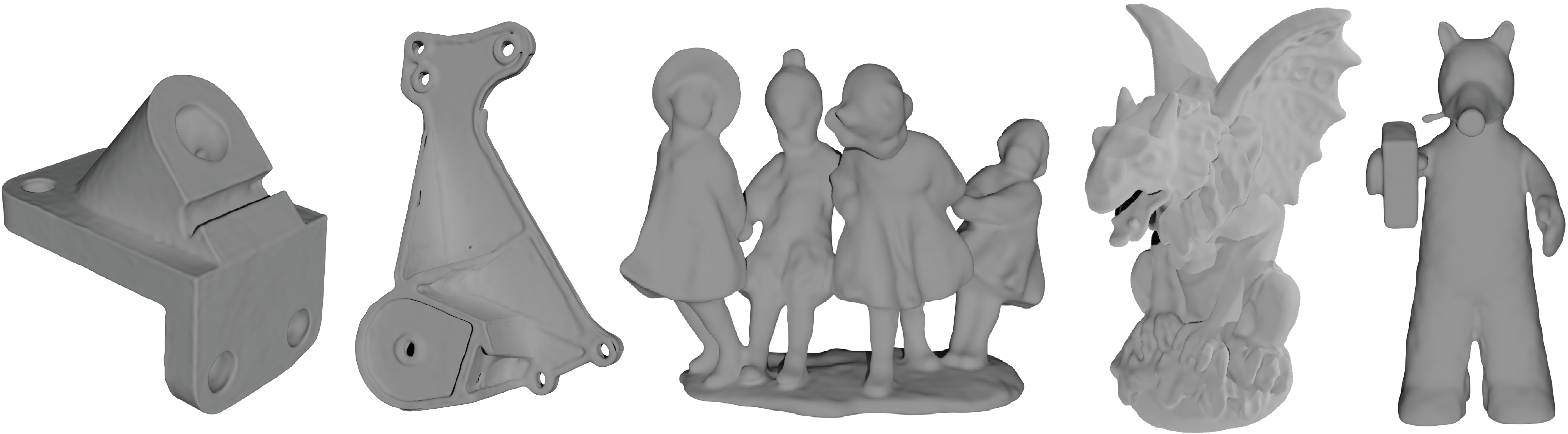}
    \caption{Visualization results on the SRB dataset. From left to right: Anchor, Daratech, DC, Garagoyle, and Lord Quas.}
    \label{fig:srb}
\end{figure}

Our improvement is less pronounced compared to prior methods, as the SRB dataset represents a relatively simple benchmark without complex structures. 
Additional visual results are provided in \Cref{fig:srb}.

\begin{table}[ht!]
    \centering
    \begin{tabular}{ lcccc }
        \toprule
        Compare with                    & \multicolumn{2}{c}{GT} & \multicolumn{2}{c}{Scans} \\
        \midrule
        Method                          & $d_{C}$ & $d_{H}$ & $d_{\vec{C}}$ & $d_{\vec{H}}$ \\
        \midrule
        IGR wo n                        & 1.38 & 16.33 & 0.25 & 2.96 \\
        SIREN wo n                      & 0.42 & 7.67  & \textbf{0.08} & \textbf{1.42} \\
        SAL \cite{atzmon2020sal}        & 0.36 & 7.47  & 0.13 & 3.50 \\
        IGR+FF \cite{Lipman:2021:PTD}   & 0.96 & 11.06 & 0.32 & 4.75 \\
        PHASE+FF \cite{Lipman:2021:PTD} & 0.22 & 4.96  & \textbf{\underline{0.07}} & 1.56 \\
        DiGS \cite{ben2022digs}         & \textbf{0.19} & 3.52  & \textbf{0.08} & 1.47 \\
        StEik \cite{yang2023steik}      & \textbf{\underline{0.18}} & \textbf{\underline{2.80}} & 0.10 & 1.45 \\
        Ours  & \textbf{0.19} & \textbf{3.17} & 0.09 &  \textbf{\underline{1.36}} \\
        \bottomrule
    \end{tabular}
    \caption{Surface reconstruction metrics on SRB \cite{berger2013benchmark}.}
    \label{tab:srb_surface_metrics}
\end{table}

\subsection{Sphere Tracing}

For implementation details, we adopt the sphere tracing algorithm~\cite{Hart:1996:STG}, as implemented in Yariv et al.'s work~\cite{Yariv:2020:MNS}. For each pixel, the algorithm advances along the ray by the signed distance function value at the current point, repeating this process until one of the following conditions is satisfied: \textit{convergence}, where the SDF value falls below a threshold of $5.0 \times 10^{-5}$; \textit{divergence}, where the ray steps outside the unit sphere; or the maximum step limit of $30$ is reached.

For signed distance functions, we use trained models from SAL~\cite{atzmon2020sal}, DiGS~\cite{ben2022digs}, StEik~\cite{yang2023steik}, and our proposed method. The evaluation is conducted on five randomly selected objects (airplane, car, watercraft, rifle, and lamp) from ShapeNet~\cite{chang2015shapenet}. For each object, we generate ten camera poses arranged in a circular trajectory around the central object, with a radius of $1.0$ and a height of $0.5$. The rendered images have a resolution of $500 \times 500$ pixels.

\begin{figure*}
    \centering
    \includegraphics[width=1.0\linewidth]{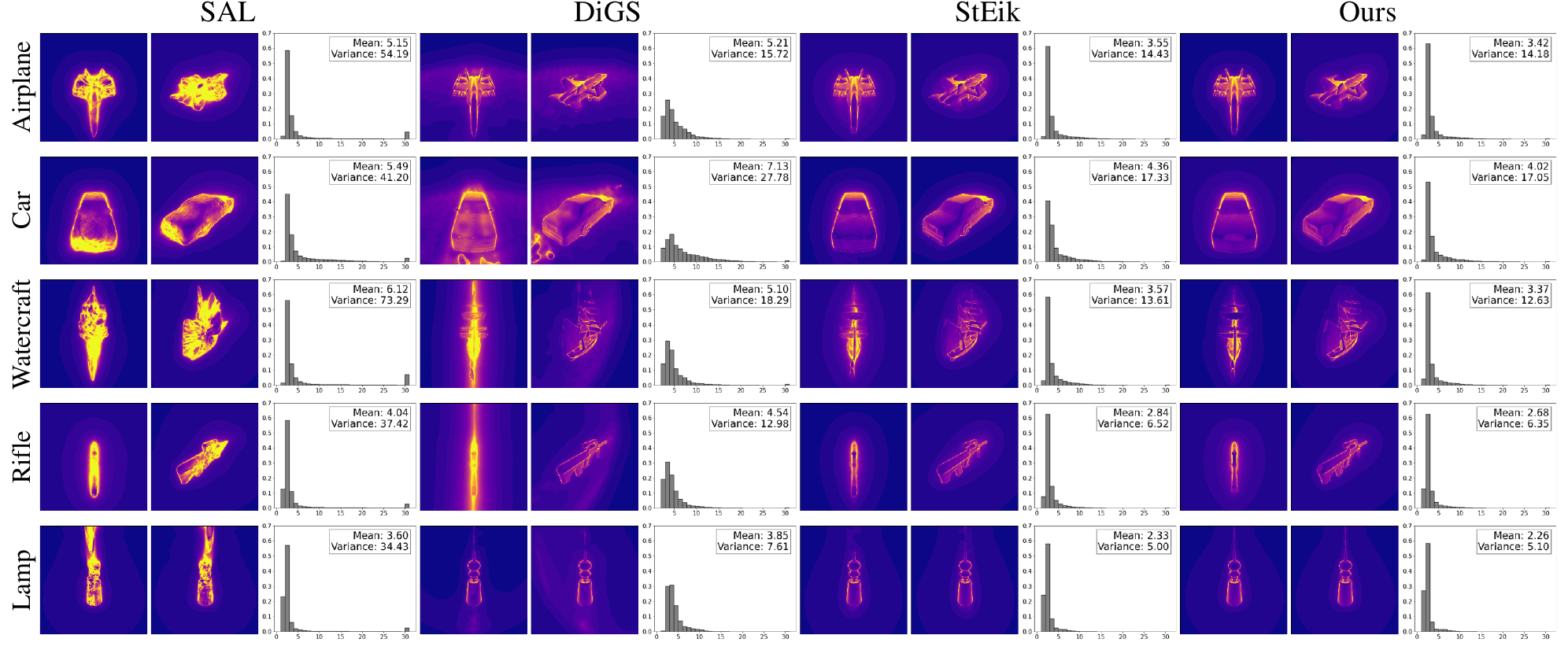}
    \\ [1ex]
    \caption{Visualization of iteration counts for each pixel and their corresponding histograms. The iteration count images are taken from two arbitrary poses, while the histograms gather outcomes across all ten rendered poses. Brighter pixels indicate a higher number of queries required for convergence before termination, while darker pixels signify fewer queries. The histograms clearly show that when rendering with our output, most pixels require fewer iterations to determine the surface.}
    \label{fig:large_st}
\end{figure*}

\Cref{fig:sphere_tracing} and \Cref{fig:large_st} illustrate the number of steps required for each pixel until ray marching terminates under one of the three conditions described above. Brighter pixels correspond to rays that are harder to converge, requiring more queries, whereas our model produces relatively darker results compared to other models. Furthermore, the histograms for our model are more skewed to the left, highlighting its efficiency. These observations demonstrate that our model excels at early divergence detection in non-intersected regions and requires fewer steps to locate the surface in intersected regions. This efficiency stems from the smoothness and accuracy of our signed distance function, particularly its high quality near the surface, which significantly enhances the rendering performance of sphere tracing.

\begin{figure*}[htbp]
    \centering
    \begin{subfigure}[b]{0.49\linewidth}
        \centering
        \includegraphics[width=\linewidth]{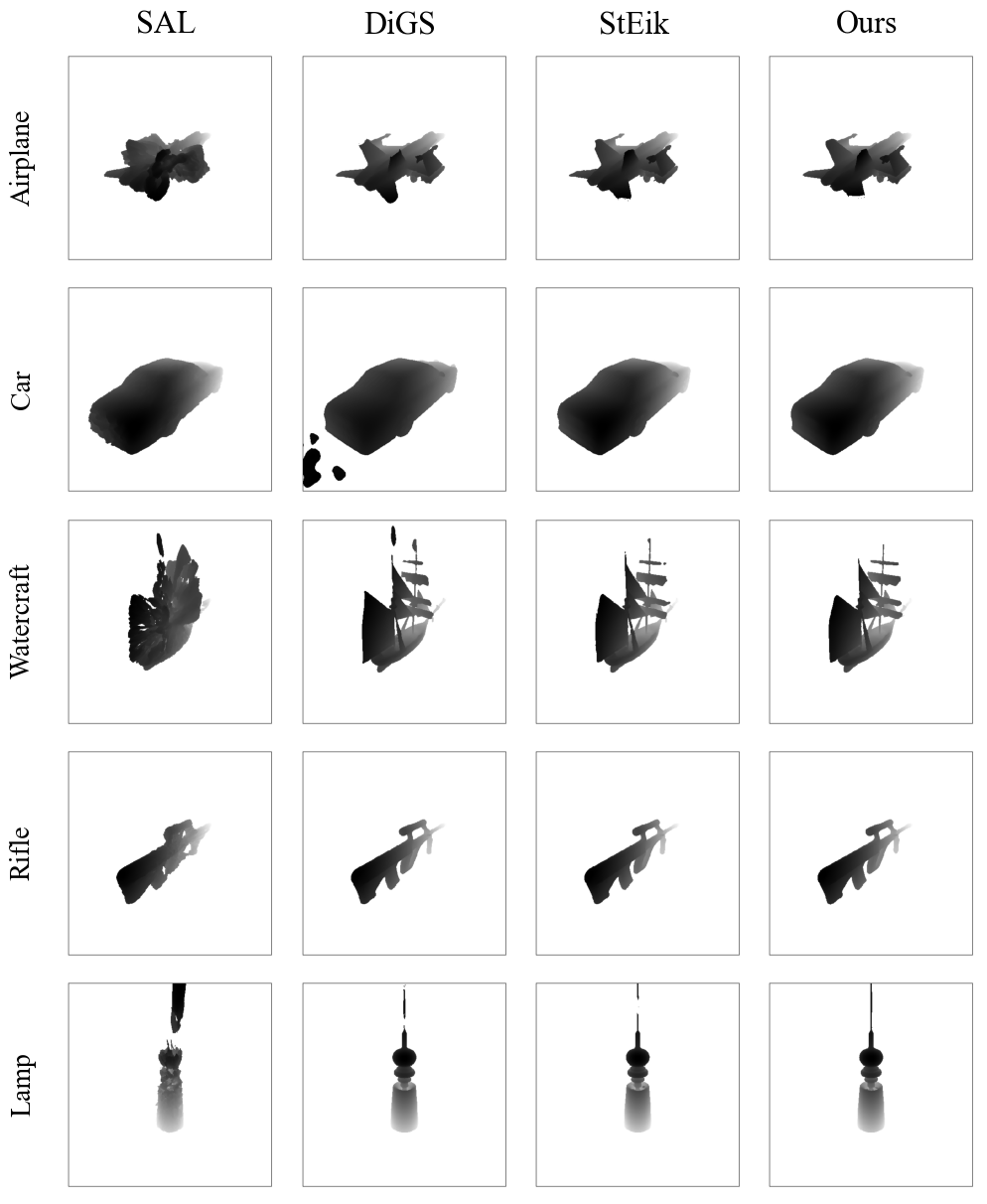}
        \caption{Depth maps showcasing the distance from the camera to the surface. }
        \label{fig:depth}
    \end{subfigure}
    \hfill
    \begin{subfigure}[b]{0.49\linewidth}
        \centering
        \includegraphics[width=\linewidth]{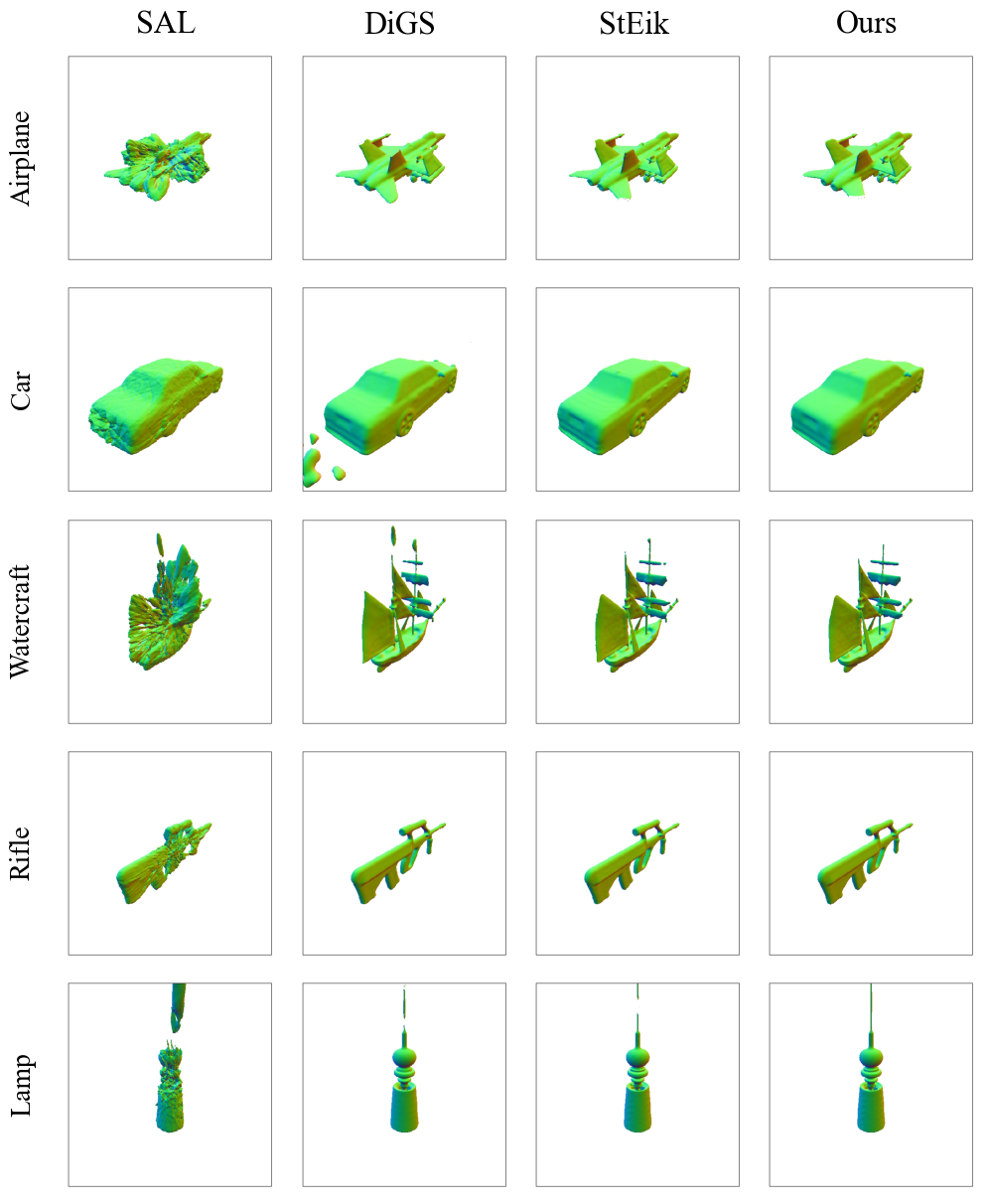}
        \caption{Normal maps illustrating surface orientation at each point.}
        \label{fig:normal}
    \end{subfigure}
    \caption{Rendering results with sphere tracing algorithm.}
    \label{fig:combined_spheretracing}
\end{figure*}

Because \Cref{fig:sphere_tracing} and \Cref{fig:large_st} only visualize the computation costs, to verify the accuracy of the object shapes in rendering, we visualize the depth and surface normals at the intersections. For rays that do not converge within 30 iterations, the intersection is approximated by identifying sign transitions at 100 equally spaced sampling points along the ray. The normal vector at the intersection, denoted by $\hat{x}$, is computed using \Cref{eq:normal} to enhance the visualization of the geometries. Non-intersected areas are masked in white. \Cref{fig:combined_spheretracing} demonstrates that our model accurately detects and represents the object's surface in this downstream application, while other models show artifacts and distortions.

\begin{equation}
\label{eq:normal}
\hat{n}(\theta) = \frac{\nabla f(\hat{x}(\theta), \theta)}{\|\nabla f(\hat{x}(\theta), \theta)\|_2}.
\end{equation}

These results demonstrate that in sphere tracing rendering, the distance query accuracy of our model effectively guides the ray toward the object's surface, enhancing rendering efficiency. Additionally, its precision at the zero level set ensures an accurate representation of the object.

\section{Relation to PHASE~\cite{Lipman:2021:PTD}}

While derived from very different mathematical principles and a different motivation, our method turns out to have a close relation to the PHASE model proposed by Lipman. 
The PHASE model essentially simulates two fluids finding equilibrium in a container by minimizing an energy functional.
A smooth approximation of the indicator of different fluids is involved and interpreted as an \emph{occupancy function} $o(\bm{x})$ such that $o$ outputs $1$ when outside of the object, $-1$ when inside the object, and $0$ when exactly on the surface.
Lipman further adds a reconstruction loss to encourage $o(\bm{x})$ to vanish at the boundary and adapts the Van der Waals-Cahn-Hilliard theory of phase transitions~\cite{modica1987gradient, sternberg1988effect}, resulting in the following functional to minimize:
\begin{equation}
    \mathcal{F} (o) = w_b \mathcal{L}(o) + \int_\Omega \epsilon \| \nabla o \|^2 + W(o),
    \label{eq:phase_log_transform}
\end{equation}
where $w_b$ is the weight of the reconstruction loss, $\mathcal{L}(o)$ is the reconstruction loss that encourages $o$ to be zero at the boundary, $\epsilon$ is a small positive constant, and $W(o)$ is a potential.
Lipman showed that by choosing a double-well potential $W(o) = o^2 - 2|o| + 1$, the minimizer can be converted into an approximated signed distance function $s$ through a log transform:
\begin{equation}
s = -\sqrt{\epsilon} \ln\left(1 - \left|o\right|\right) \text{sign}(o).
\label{eq:phase}
\end{equation}

It turns out that our heat field $h$ when optimized under our heat loss is closely related to the regularized occupancy function $o$. If we set $h = 1 - |o|$ and $\lambda = \epsilon^{-\frac{1}{2}}$ and solve for the screened Poisson equation (\Cref{eq:heat}), we would obtain PHASE's regularized occupancy. 

However, there are several crucial differences between our approach and PHASE where our theory and derivation has led to additional insights and huge differences in implementations and results. 

First, based on a theoretical framework aiming the area minimization, PHASE's theory promotes setting the weight $w_b$ of the boundary loss (\Cref{eq:boundary_loss}) to $w_b = \epsilon^{\alpha}$ where $\alpha \in (\frac{1}{4}, \frac{1}{2})$.
However, we found that finding the exact minimal area is detrimental to the optimization, as discussed in \Cref{sec:bg}.
On the contrary, according to our theory and experiments, we find it necessary to maintain boundaries between points and set the weight $w_b$ to a much higher value to ensure the boundary condition of the screened Poisson equation is satisfied (\Cref{eq:heat}), as states in main text.

Theorem 2 in the PHASE paper states that to achieve minimal area, $w_b$ should converge to $0$ as $\epsilon$ approaches an infinitesimal value.
However, this interpretation does not align with the nature of the signed distance function reconstruction task which should complete the manifold and connect points, as discussed at the end of \Cref{sec:bg}, and may lead to unintended consequences as follows:
When the boundary weight $w_b$ is overly small, the signed distance function values of boundary points cannot even converge to close to 0. 
When the weight $w_b$ is still not large enough, their output collapses from a surface-based-distance signed distance function to a point-cloud-based-distance \emph{unsigned} distance function upon reaching the target where the boundaries are only point clouds and the area becomes almost zero.
In contrast, our model demonstrates robust and accurate performance, effectively connecting points and interpolating boundaries by taking advantages of neural network's spectral bias~\cite{Rahaman:2019:SBN}, with a large boundary weight $w_b$ and an adaptive absorption $\lambda$, as discussed in \Cref{sec:disc}.

In addition to \Cref{fig:phase-boundary-weight}, we show more empirical results on our 2D dataset supporting our claim here in the supplementary.
In \Cref{fig:phase-boundary-weight-circle}, we show the visual results of PHASE on a single circle across different boundary weight choices, and compute the surface reconstruction and distance metrics.
We show visual results on the rest of the 2D dataset in \Cref{fig:phase-boundary-weight-more}, and the mean metrics for each boundary loss weight over the whole 2D dataset in \Cref{tab:phase-boundary-weight-metrics}.
In these experiments, we use eikonal loss weight $0.1$, as provided in the PHASE paper.
We also show one example from ShapeNet in \Cref{fig:phase-boundary-weight-3d}, where we use eikonal loss weight $1.0$.
From our theory and the examples above, one can clearly see that using a small boundary loss weight is not the best strategy.

\input{figures/phase-boundary-exp-fig}

Second, we design our network to output the signed distance $u$ directly instead of the heat $h$, whereas PHASE's model would directly output the occupancy $o$ and convert to signed distance. 
We show that this can lead to extremely numerically unstable results. 
Consider the case where the occupancy function $o$ outputs $1$ or $-1$, then the log transform (\Cref{eq:phase_log_transform}) would simply output infinity or negative infinity. 
The infinities can be avoided by clamping the occupancy, but how much should we clamp?

For their proposed setting $\epsilon=0.01$, only to get $s=2$, we will need to set $o = 1 - e^{-20} \approx 0.9999999979$, which already is beyond what a 32-bit floating point number can reliably represent.
Rescaling the scene is unfortunately not going to help, since the parameter $\epsilon$ is scene dependent and needs to be scaled accordingly.

\setlength{\columnsep}{4pt}
\begin{wrapfigure}[11]{r}{0.14\textwidth}
    \vspace{-0pt}
    \includegraphics[width=\linewidth,trim={0 0 0 0},clip]{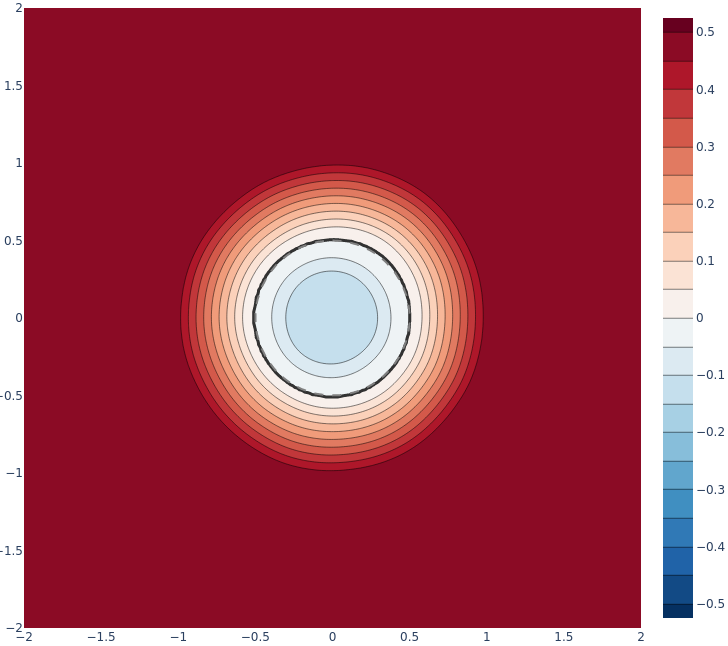}
    \includegraphics[width=\linewidth,trim={0 0 0 0},clip]{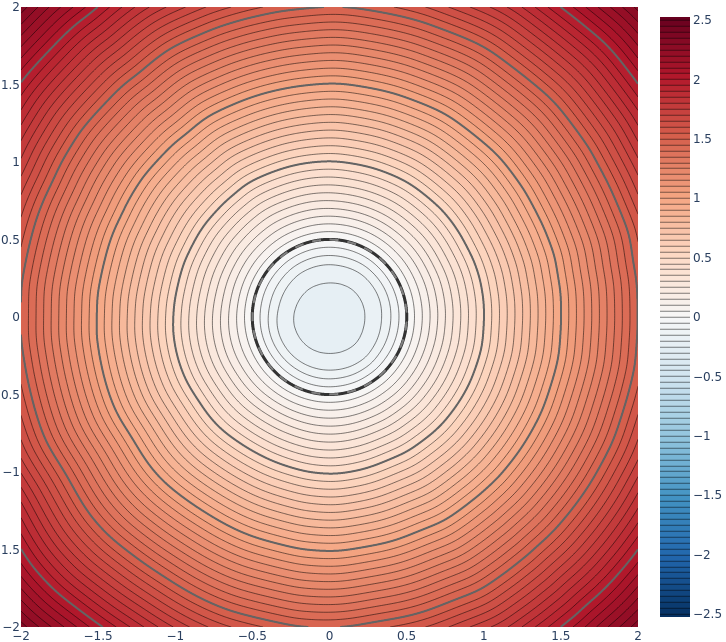}
\end{wrapfigure}
In practice, we verify that when PHASE struggles with queries that are far away from the surfaces, and show visual examples in the inset, where the first figure is PHASE result with $\epsilon = 0.01$ and occupancy clamped at $0.99$, and the second figure is our result, which can represent arbitrary signed distance function values.

Moreover, optimizing the occupancy directly instead of the distance leads to another issue when combined with the eikonal loss.
When backproping the gradient from $s$ to $o$, we have:
\begin{equation}
    \nabla s = \sqrt{\epsilon} \frac{1}{1-|o|} \nabla o
\end{equation}
When optimizing the eikonal loss $| \|\nabla s\| - 1|^p$, this $\frac{1}{1-|o|}$ term is multiplied as an coefficient and becomes unstable in the optimization.

%% file: figures/phase-boundary-exp-fig.tex
\begin{table*}[t]
    \setlength{\tabcolsep}{2pt}
    \centering
    \begin{tabular}{cccccccccc} 
        \includegraphics[width=0.09\textwidth]{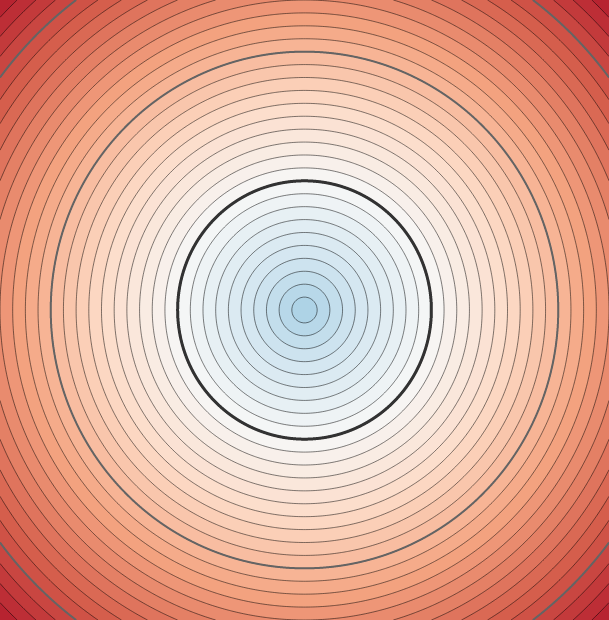} & 
        \includegraphics[width=0.09\textwidth]{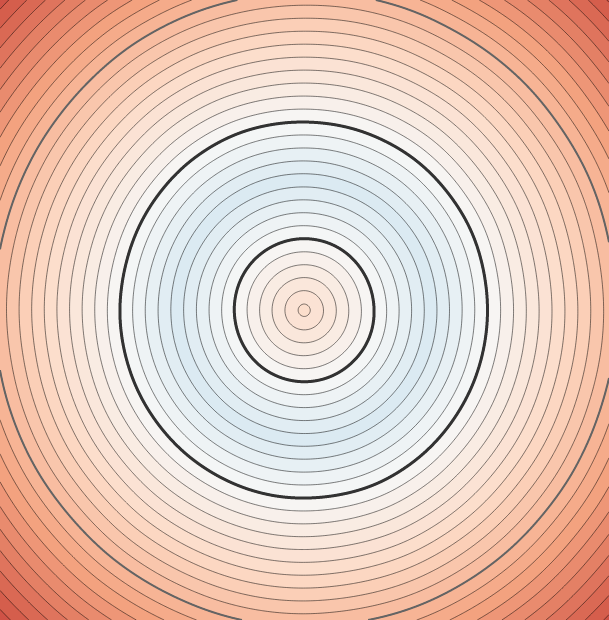} & 
        \includegraphics[width=0.09\textwidth]{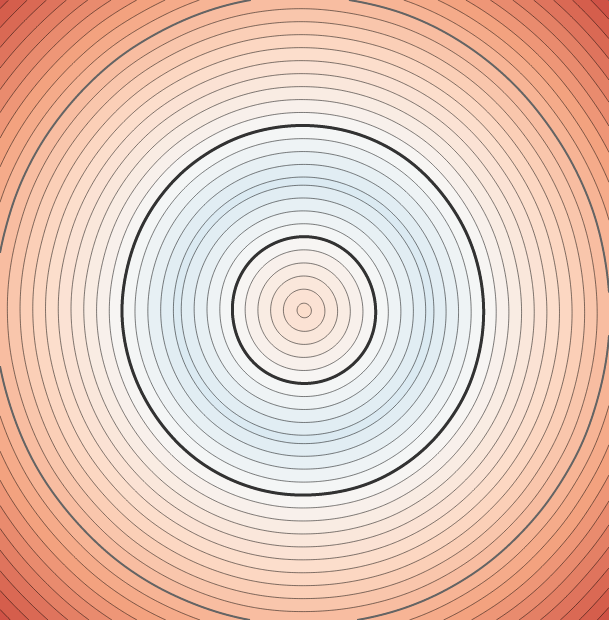} & 
        \includegraphics[width=0.09\textwidth]{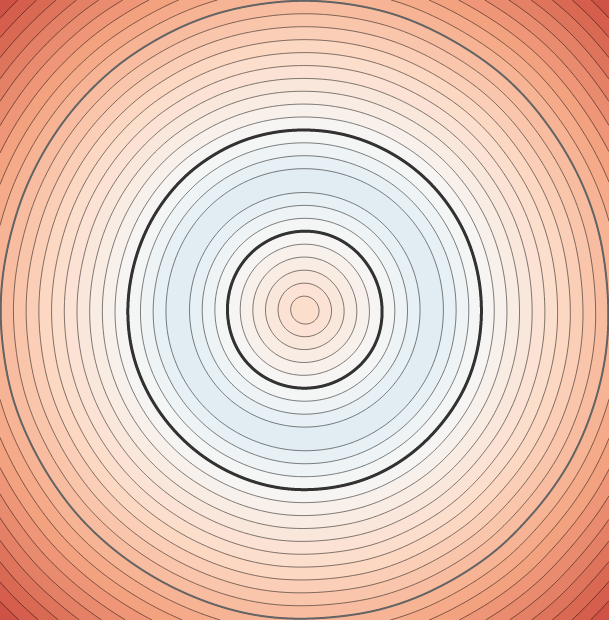} & 
        \includegraphics[width=0.09\textwidth]{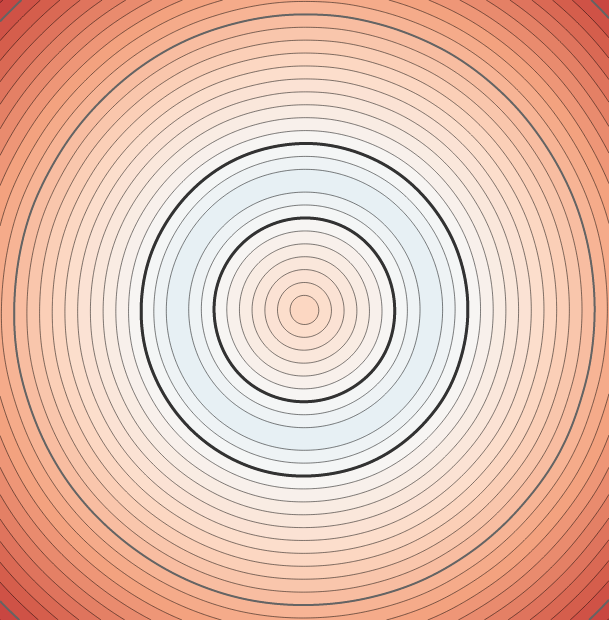} &
        \includegraphics[width=0.09\textwidth]{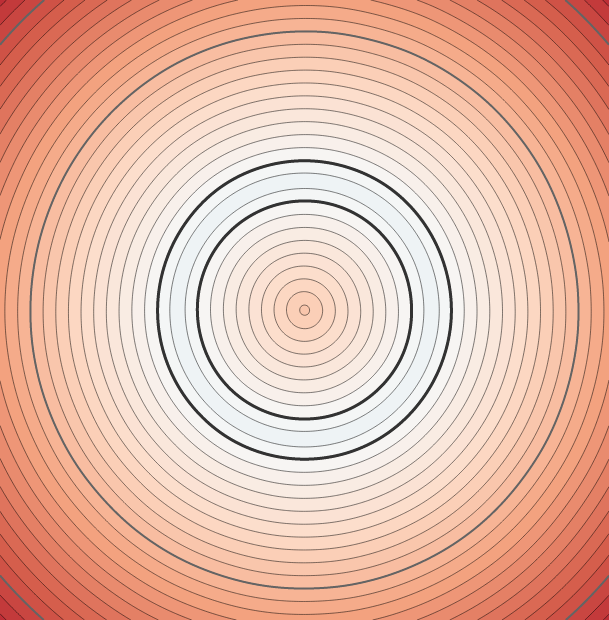} & 
        \includegraphics[width=0.09\textwidth]{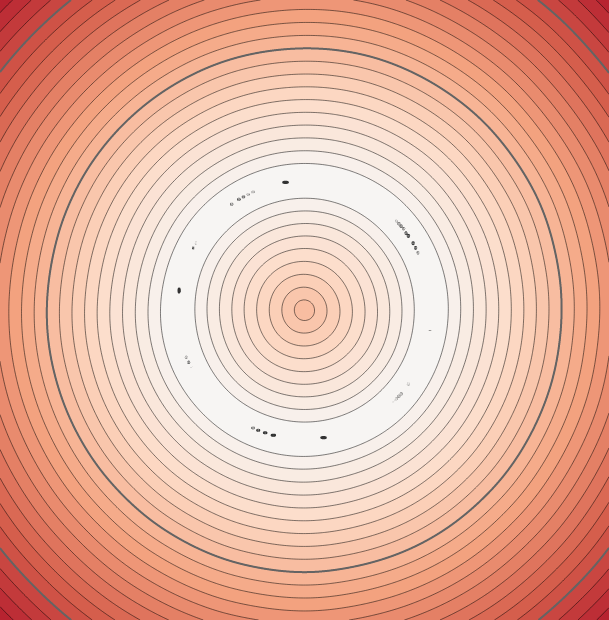} & 
        \includegraphics[width=0.09\textwidth]{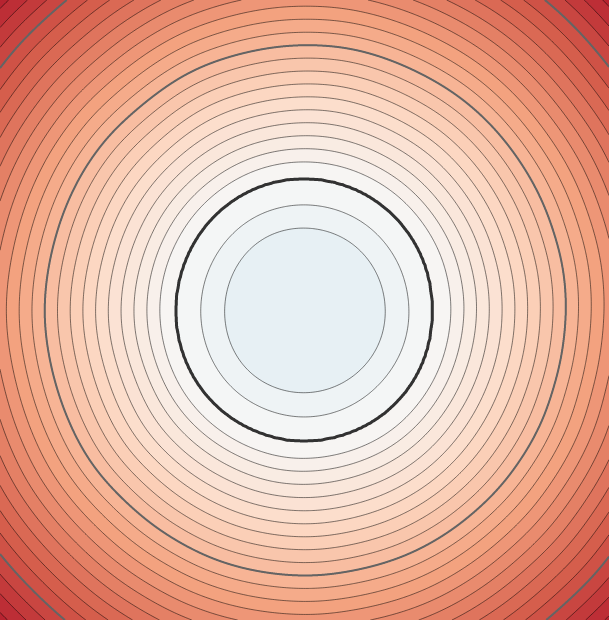} & 
        \includegraphics[width=0.09\textwidth]{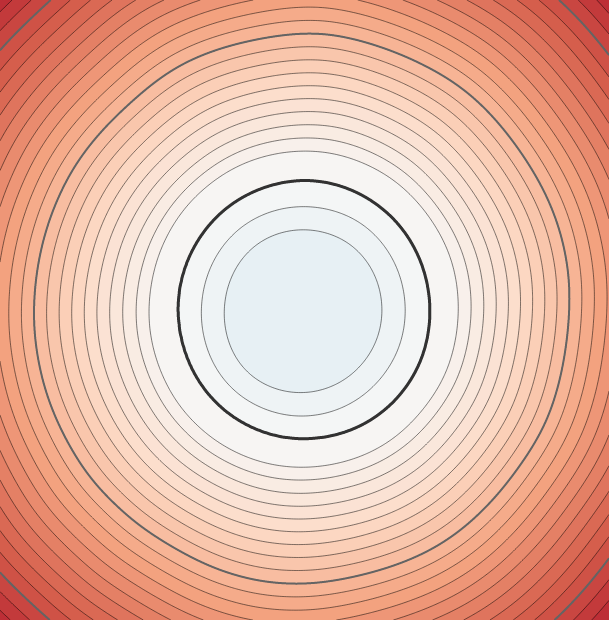} & 
        \includegraphics[width=0.09\textwidth]{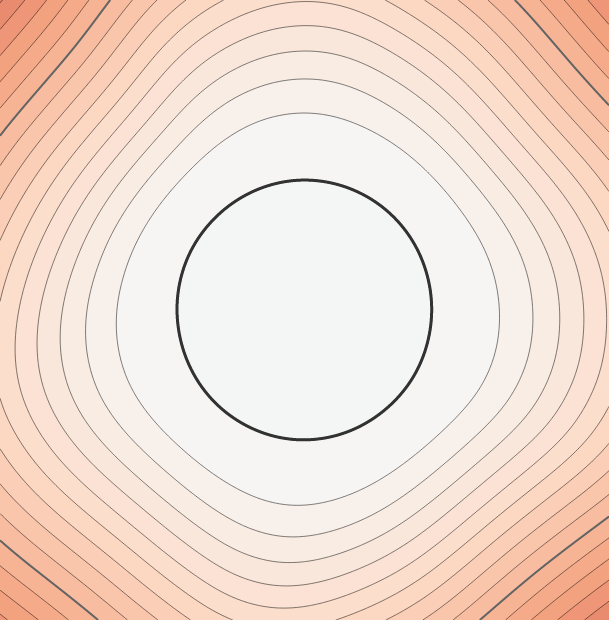} \\ [1ex]
        
        GT & $w_b = 0.1$ & $w_b = 0.2$ & $w_b = 0.3$ & $w_b = 0.5$ & $w_b = 1.0$ & $w_b = 2.0$ & $w_b = 5.0$ & $w_b = 10.0$ & $w_b = 20.0$ \\ [1ex]
        \toprule
        IoU & 0.3288 & 0.3330 & 0.3242 & 0.2975 & 0.2138 & 0.0022 & 0.9693 & 0.9912 & 0.9917 \\
        Chamfer & 0.2244 & 0.2138 & 0.1955 & 0.1434 & 0.0777 & 0.0175 & 0.0080 & 0.0026 & 0.0025 \\
        Hausdorff & 0.2315 & 0.2216 & 0.1991 & 0.1455 & 0.0799 & 0.1584 & 0.0150 & 0.0102 & 0.0105 \\
        RMSE & 0.2318 & 0.2209 & 0.2085 & 0.1740 & 0.1477 & 0.1476 & 0.0481 & 0.0681 & 0.3004 \\
        MAE & 0.2237 & 0.2122 & 0.1982 & 0.1552 & 0.1048 & 0.0556 & 0.0336 & 0.0610 & 0.2744 \\
        SMAPE & 0.9159 & 0.8814 & 0.8404 & 0.7059 & 0.5333 & 0.3396 & 0.2028 & 0.3291 & 1.0863 \\
        \bottomrule
    \end{tabular}
    \caption{Comparison of PHASE results on a circle with different $w_b$ values. The color scale is the same as in \Cref{fig:phase-boundary-weight}.}
    \label{fig:phase-boundary-weight-circle}
\end{table*}

\begin{table*}[t]
    \setlength{\tabcolsep}{4pt}
    \centering
    \begin{tabular}{cccccccccc}
        \toprule
        $w_b$ & $0.1$ & $0.2$ & $0.3$ & $0.5$ & $1.0$ & $2.0$ & $5.0$ & $10.0$ & $20.0$ \\
        \midrule
        IoU & 0.2026 & 0.1843 & 0.2050 & 0.2089 & 0.2505 & 0.2061 & 0.3899 & 0.4838 & 0.4696 \\
        Chamfer & 0.2663 & 0.3122 & 0.2504 & 0.1785 & 0.1067 & 0.0499 & 0.0793 & 0.0888 & 0.1030 \\
        Hausdorff & 0.5692 & 0.7001 & 0.5635 & 0.4191 & 0.3446 & 0.3747 & 0.4567 & 0.4383 & 0.5407 \\
        RMSE & 0.4644 & 0.3314 & 0.2382 & 0.1606 & 0.1111 & 0.0734 & 0.0567 & 0.0747 & 0.1282 \\
        MAE & 0.4332 & 0.3114 & 0.2265 & 0.1500 & 0.0918 & 0.0413 & 0.0405 & 0.0623 & 0.1112 \\
        SMAPE & 1.2558 & 1.1504 & 1.0870 & 0.9475 & 0.8107 & 0.5873 & 0.7191 & 0.8877 & 1.1477 \\
        \bottomrule
    \end{tabular}
    \caption{Mean metrics of PHASE results with different $w_b$ values on the full 2D dataset.}
    \label{tab:phase-boundary-weight-metrics}
\end{table*}

\begin{figure*}[t]
    \setlength{\tabcolsep}{1pt}
    \centering
    \begin{tabular}{ccccccccc}
        \includegraphics[width=0.11\textwidth]{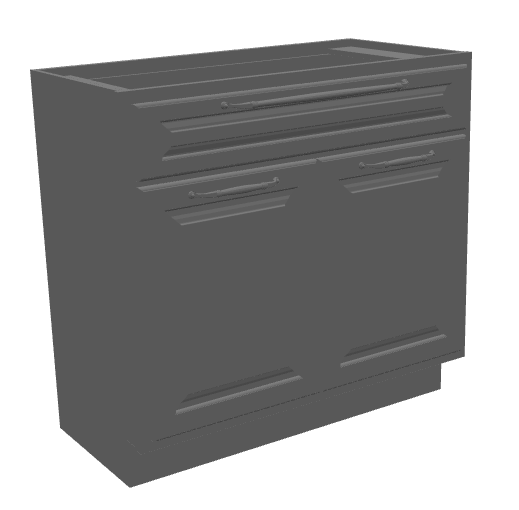} &
        \includegraphics[width=0.11\textwidth]{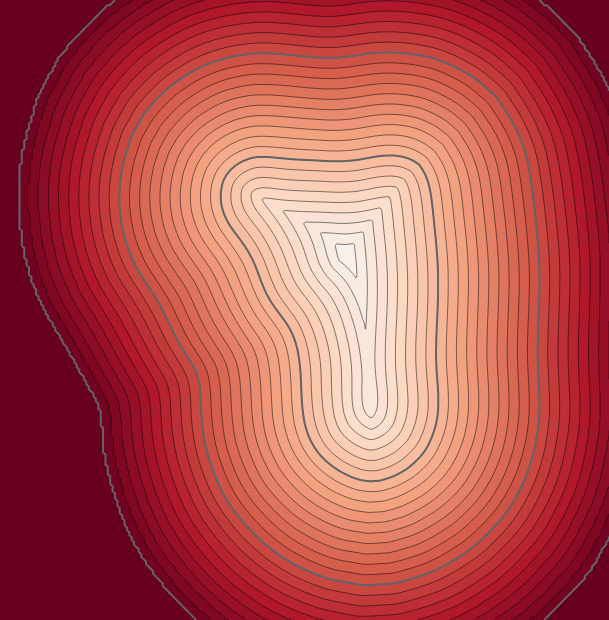} & \includegraphics[width=0.11\textwidth]{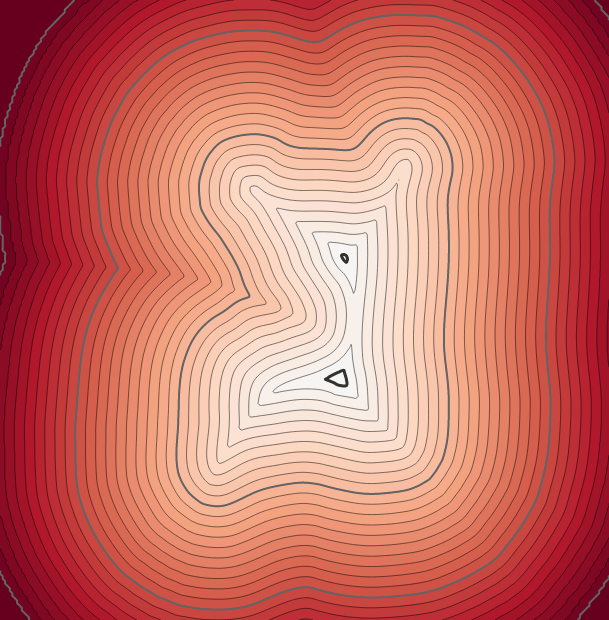} & \includegraphics[width=0.11\textwidth]{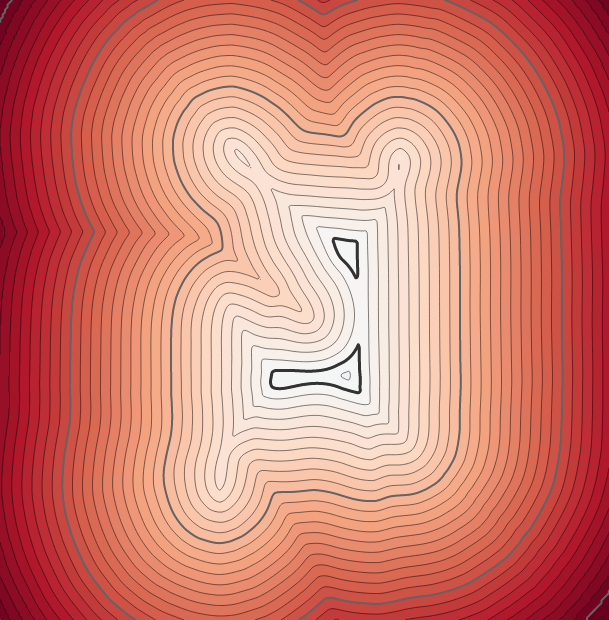} & \includegraphics[width=0.11\textwidth]{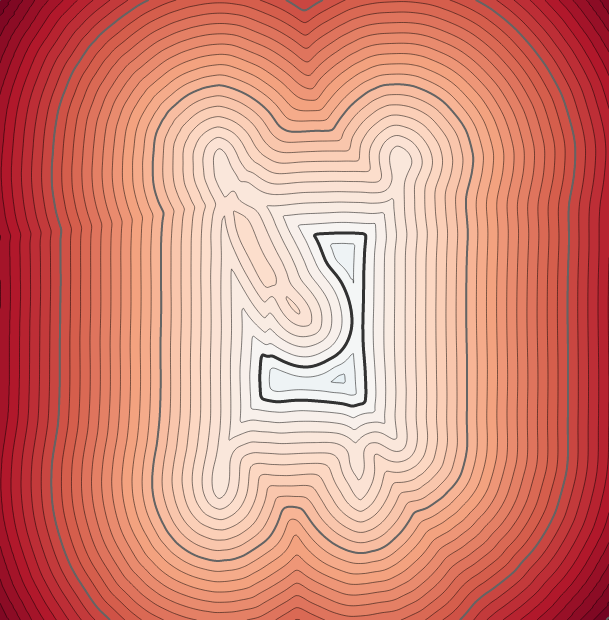} &
        \includegraphics[width=0.11\textwidth]{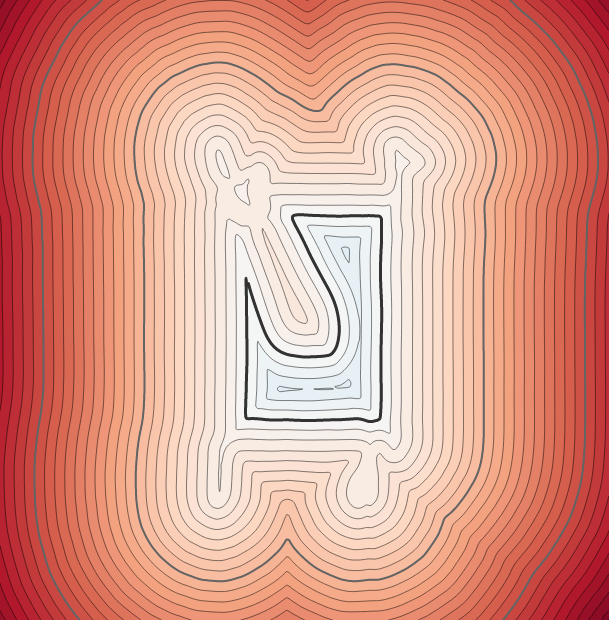} & \includegraphics[width=0.11\textwidth]{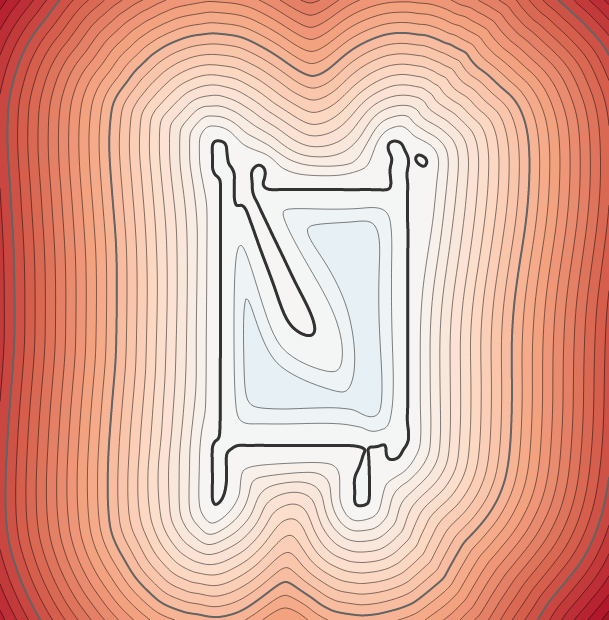} & \includegraphics[width=0.11\textwidth]{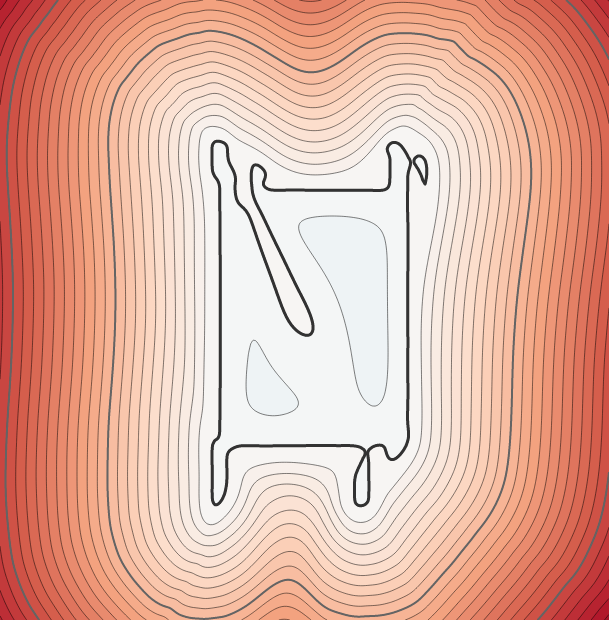} & \includegraphics[width=0.11\textwidth]{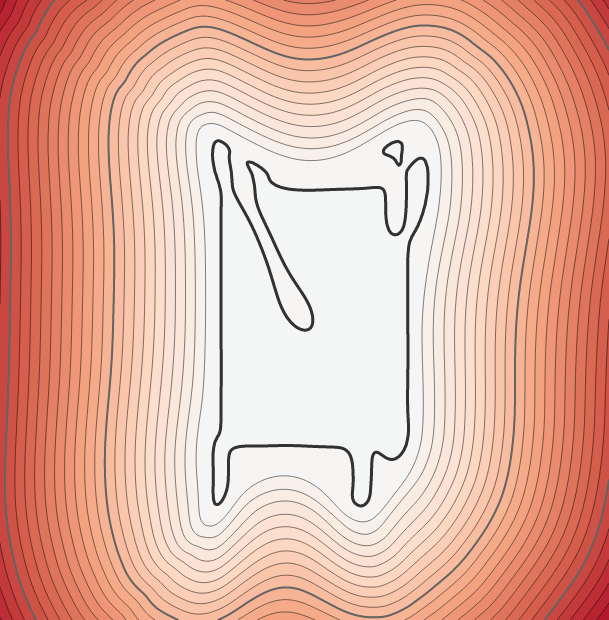} \\
        GT mesh & $w_b = 0.1$ & $w_b = 0.2$ & $w_b = 0.3$ & $w_b = 0.5$ & $w_b = 1.0$ & $w_b = 5.0$ & $w_b = 10.0$ & $w_b = 20.0$ \\
        \toprule
        IoU & 0.0000 & 0.0046 & 0.0387 & 0.1177 & 0.3106 & 0.9742 & 0.9638 & 0.9020 \\
        Chamfer & 0.2056 & 0.2055 & 0.1749 & 0.1441 & 0.0935 & 0.0060 & 0.0065 & 0.0089 \\
        Hausdorff & 0.4905 & 0.4909 & 0.4395 & 0.3855 & 0.3094 & 0.0518 & 0.0545 & 0.0651 \\
        RMSE & 0.8656 & 0.8263 & 0.7944 & 0.7505 & 0.6588 & 0.4578 & 0.4275 & 0.4041 \\
        MAE & 0.8468 & 0.8108 & 0.7769 & 0.7292 & 0.6339 & 0.4004 & 0.3528 & 0.3182 \\
        SMAPE & 1.5123 & 1.5100 & 1.4956 & 1.4608 & 1.3912 & 1.1019 & 0.9778 & 1.0324 \\
        \bottomrule
    \end{tabular}
    \caption{Comparison of PHASE results on a 3D cabinet with different $w_b$ values. SDF values are visualized only on a 2D plane. The color scale is the same as in \Cref{fig:phase-boundary-weight}.}
    \label{fig:phase-boundary-weight-3d}
\end{figure*}

\begin{figure*}[t]
    \setlength{\tabcolsep}{2pt}
    \centering
    \begin{tabular}{cccccccccc}
    \includegraphics[width=0.09\textwidth]{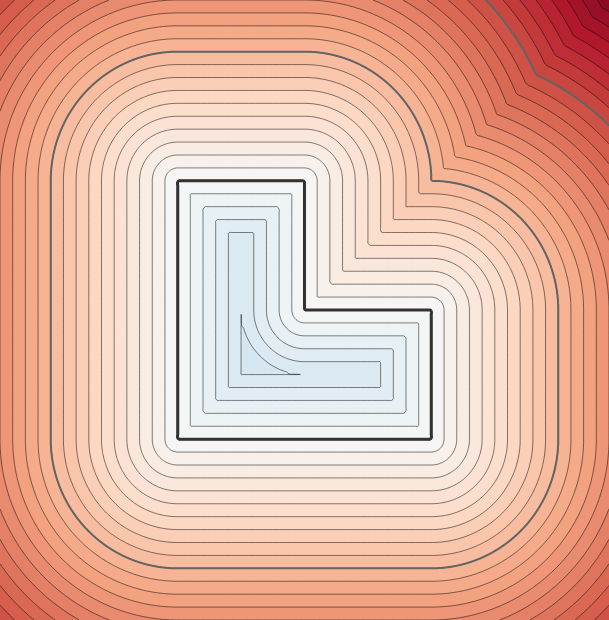} & \includegraphics[width=0.09\textwidth]{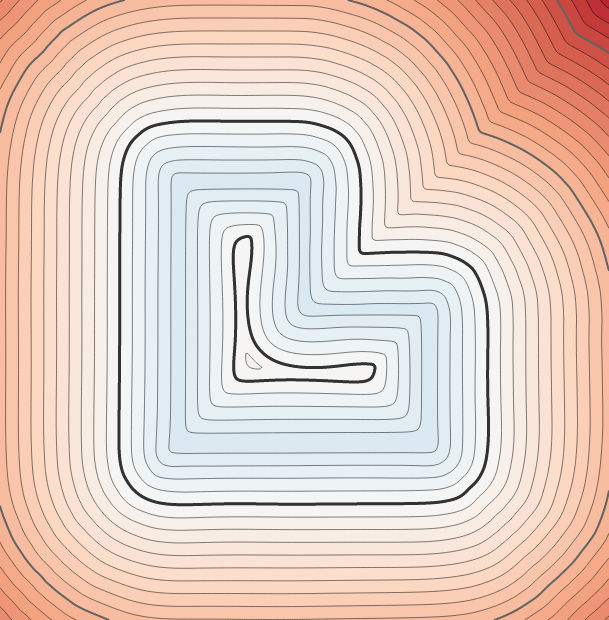} & \includegraphics[width=0.09\textwidth]{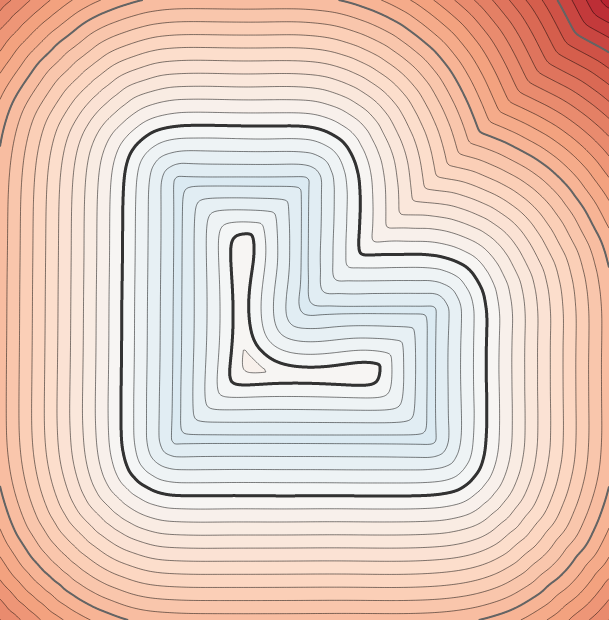} & \includegraphics[width=0.09\textwidth]{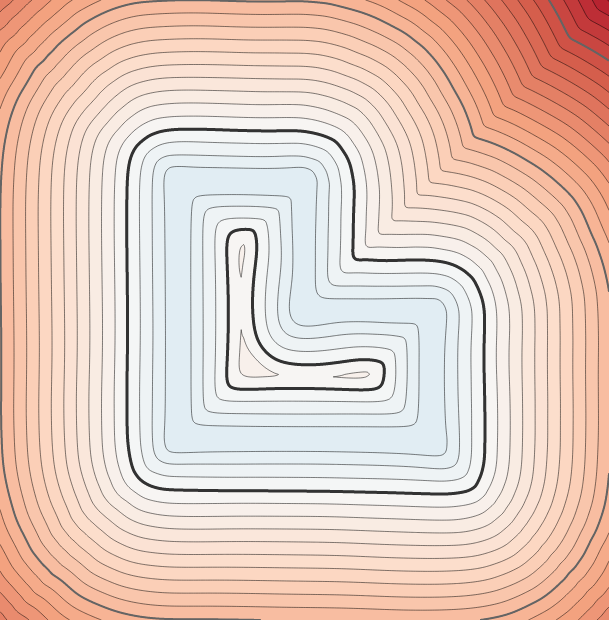} & \includegraphics[width=0.09\textwidth]{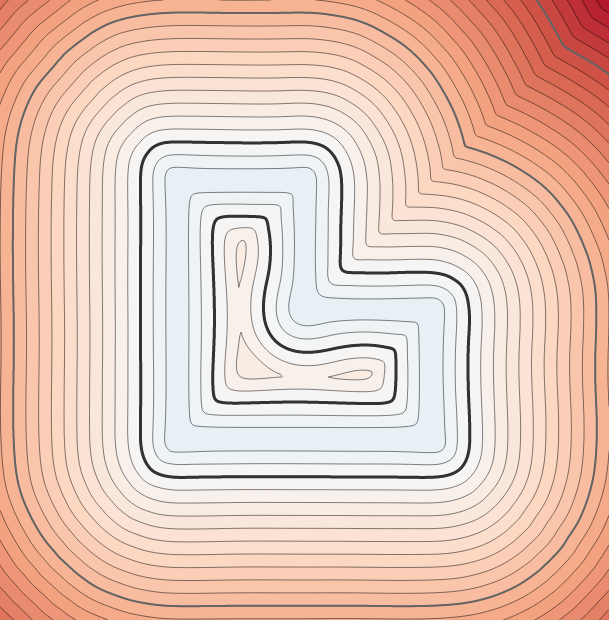} & \includegraphics[width=0.09\textwidth]{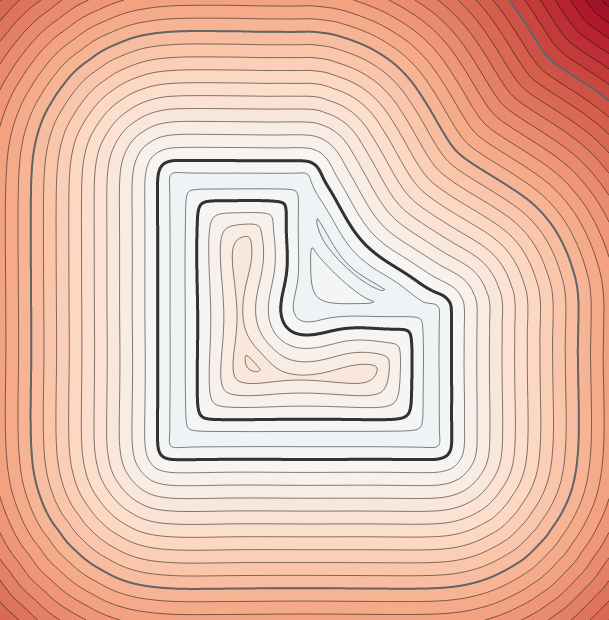} & \includegraphics[width=0.09\textwidth]{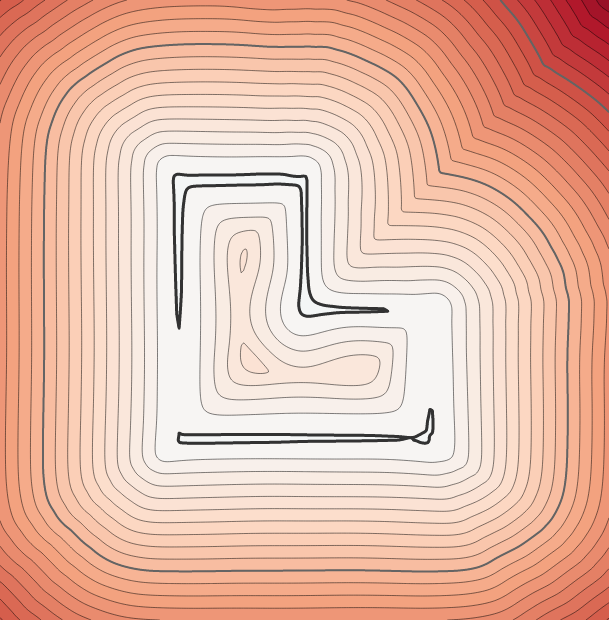} & \includegraphics[width=0.09\textwidth]{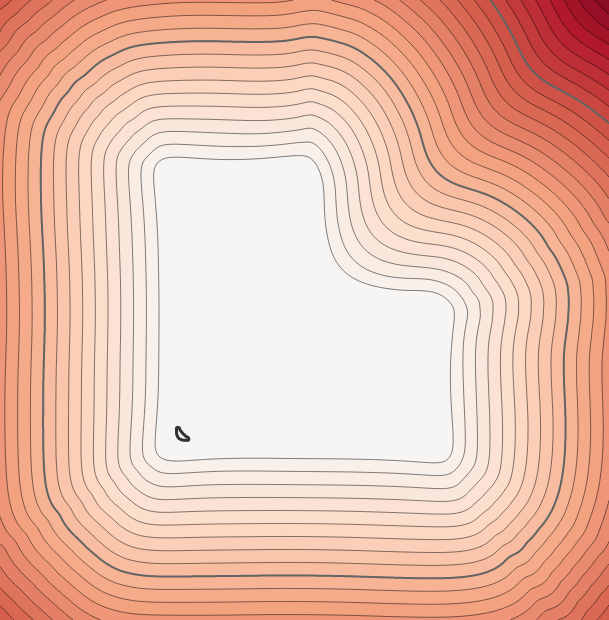} & \includegraphics[width=0.09\textwidth]{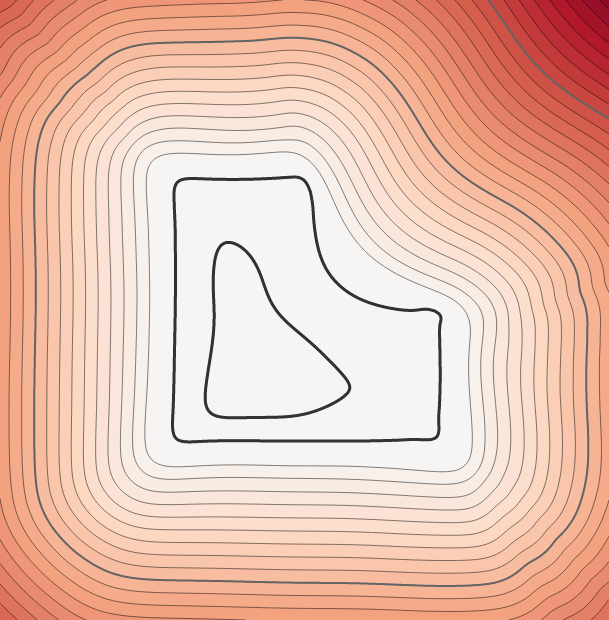} & \includegraphics[width=0.09\textwidth]{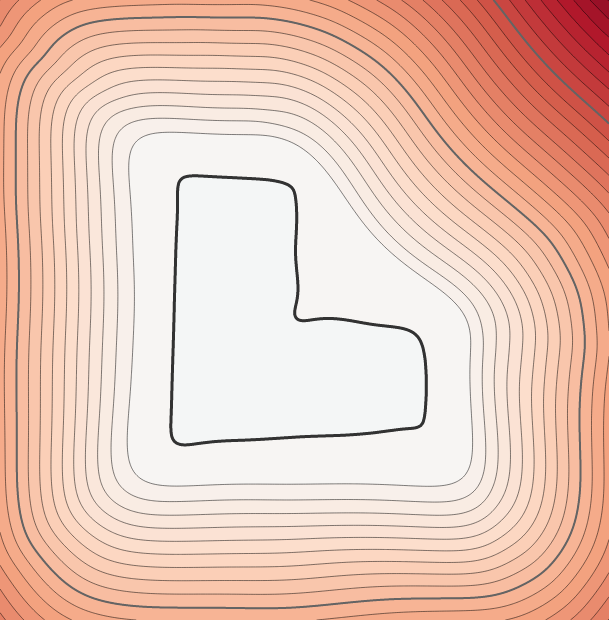} \\
    \includegraphics[width=0.09\textwidth]{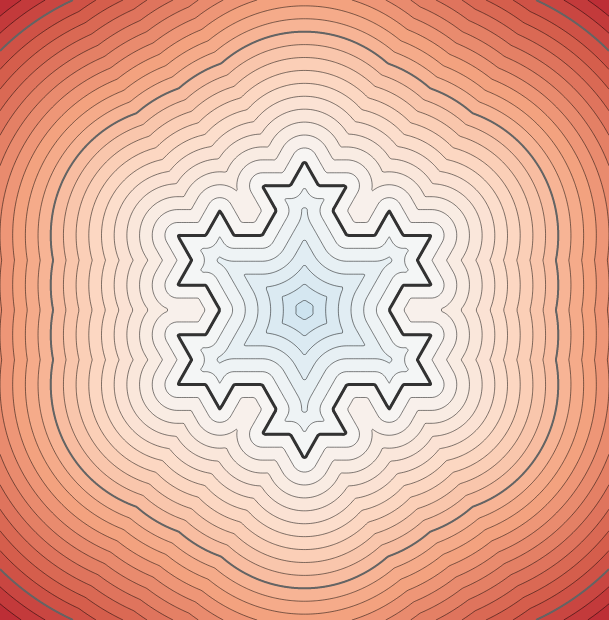} & \includegraphics[width=0.09\textwidth]{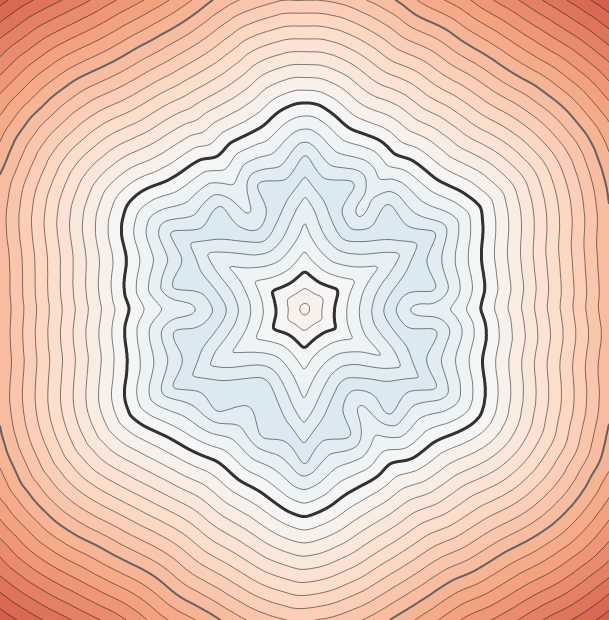} & \includegraphics[width=0.09\textwidth]{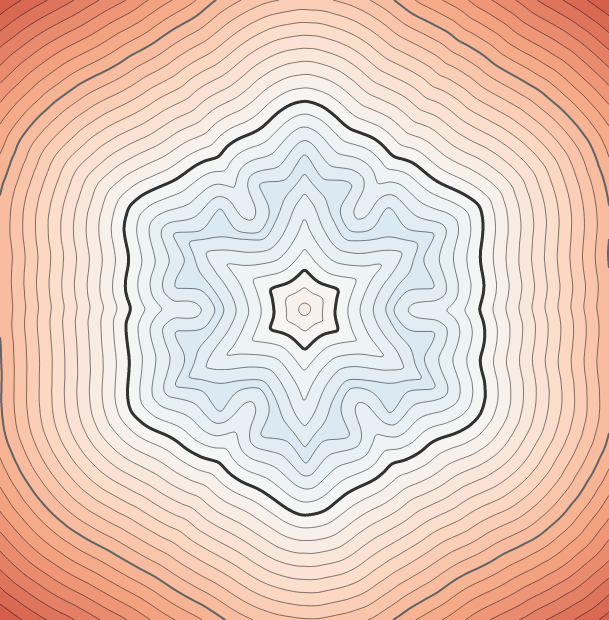} & \includegraphics[width=0.09\textwidth]{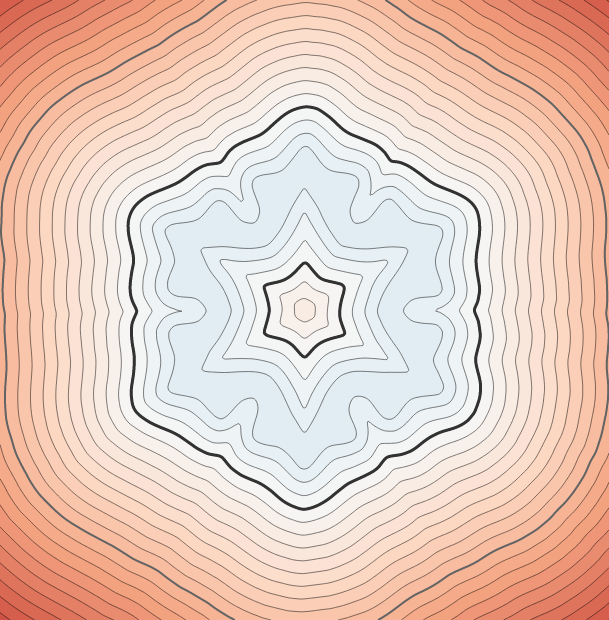} & \includegraphics[width=0.09\textwidth]{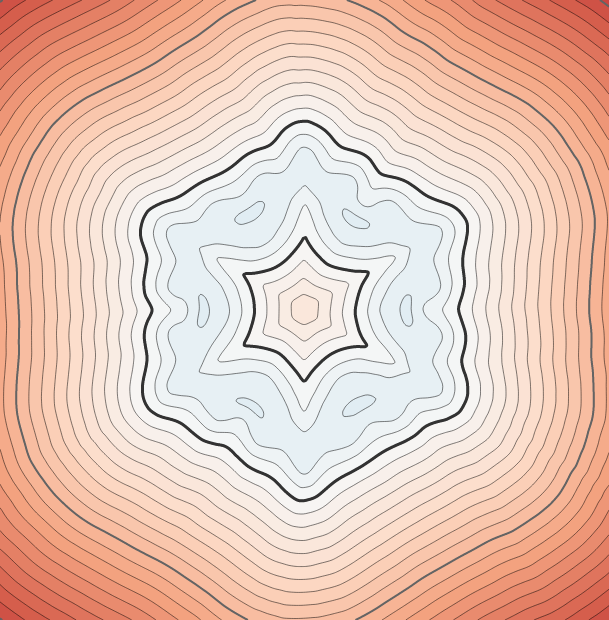} & \includegraphics[width=0.09\textwidth]{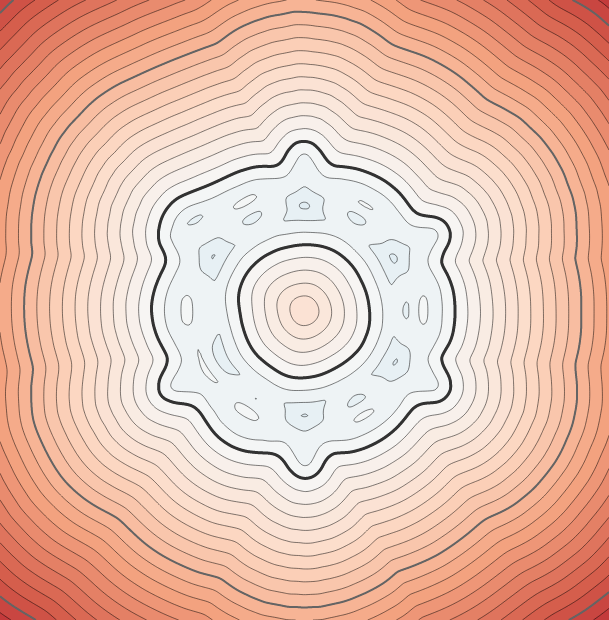} & \includegraphics[width=0.09\textwidth]{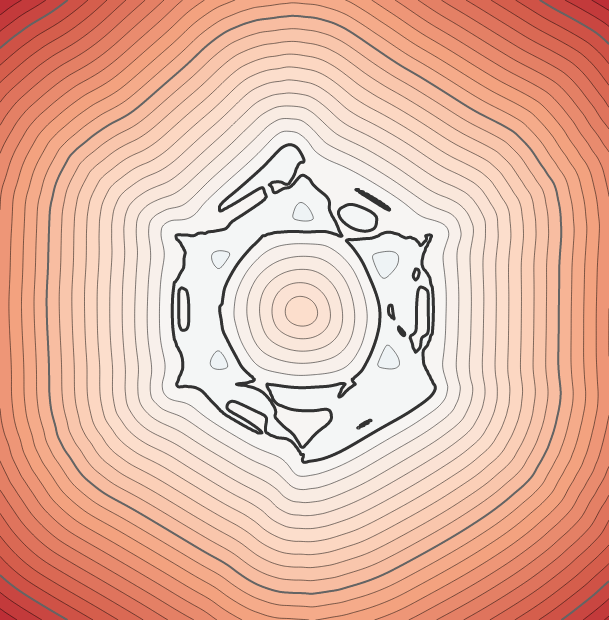} & \includegraphics[width=0.09\textwidth]{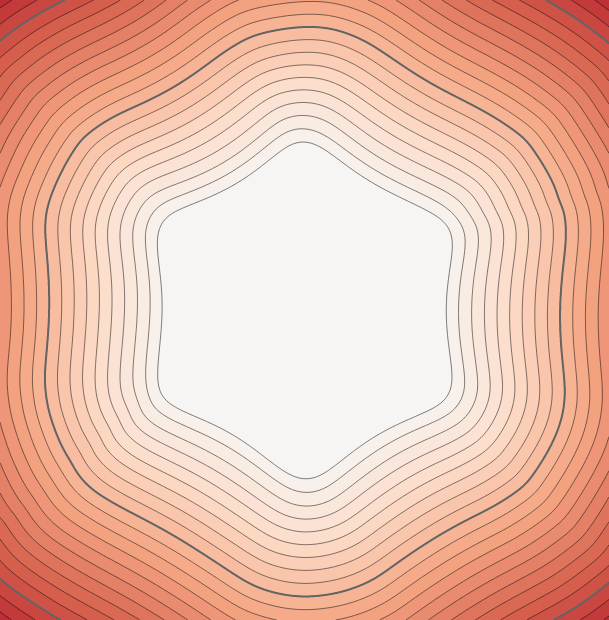} & \includegraphics[width=0.09\textwidth]{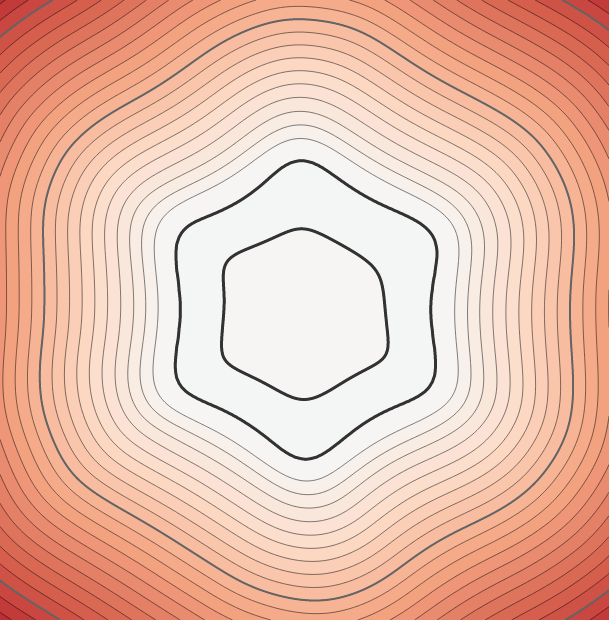} & \includegraphics[width=0.09\textwidth]{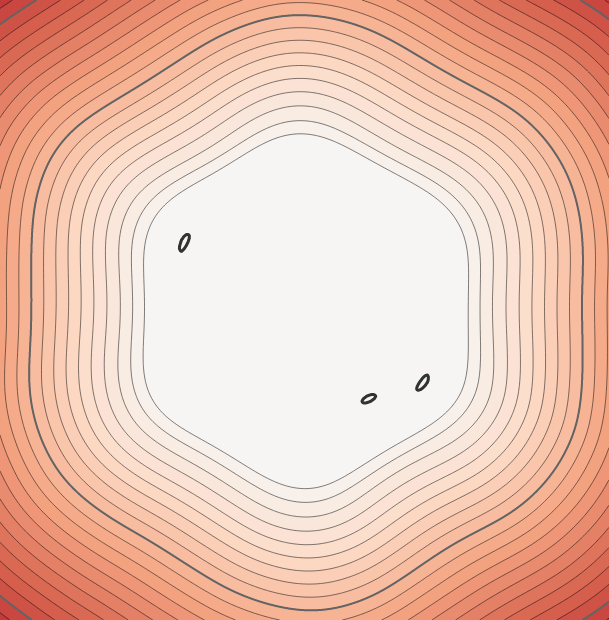} \\
    \includegraphics[width=0.09\textwidth]{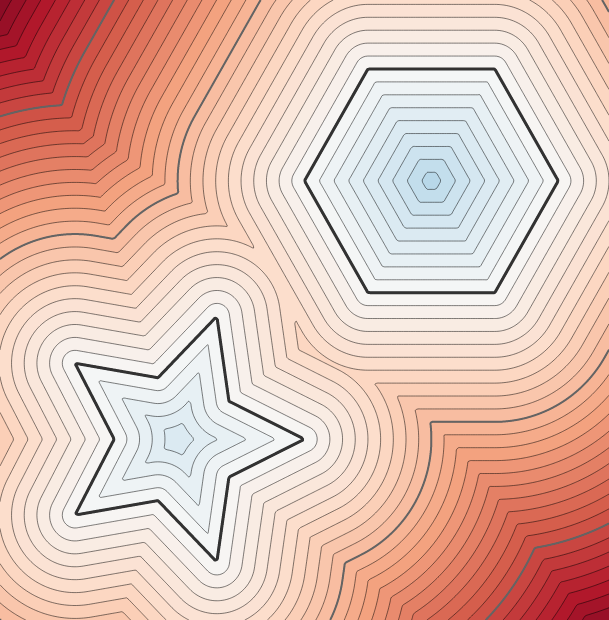} & \includegraphics[width=0.09\textwidth]{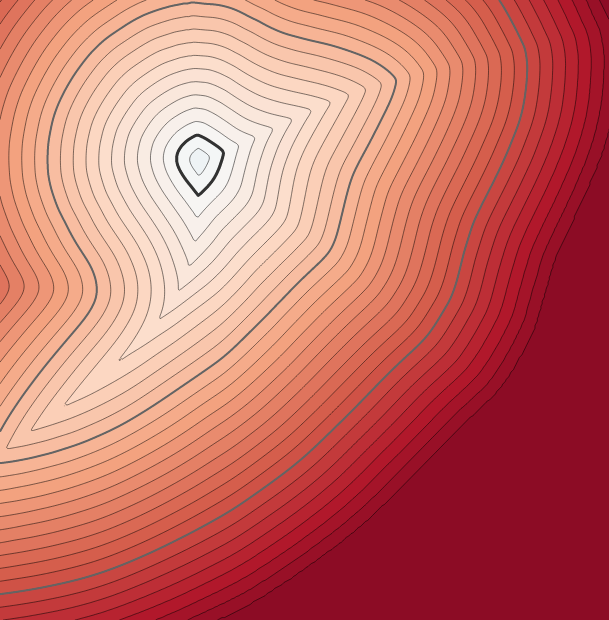} & \includegraphics[width=0.09\textwidth]{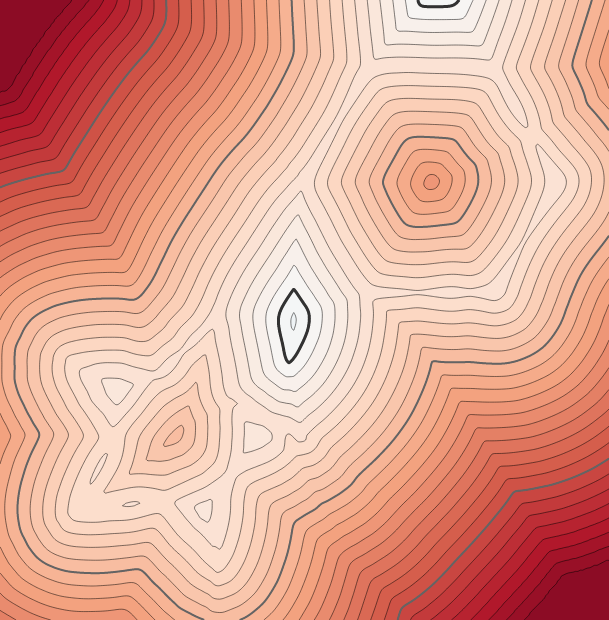} & \includegraphics[width=0.09\textwidth]{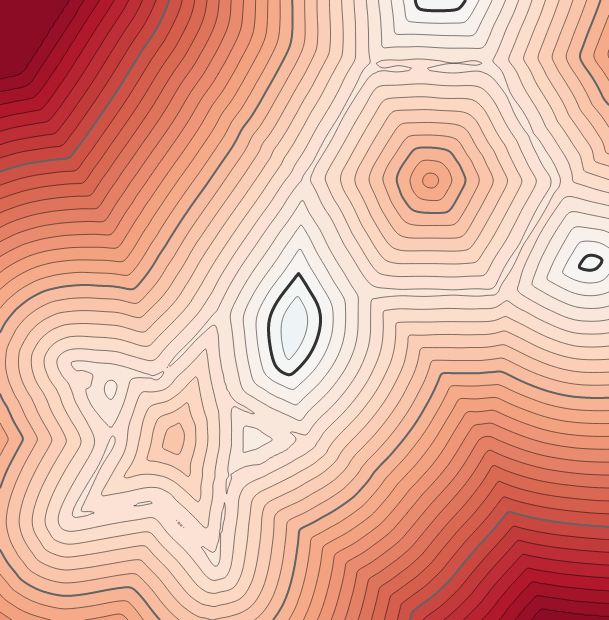} & \includegraphics[width=0.09\textwidth]{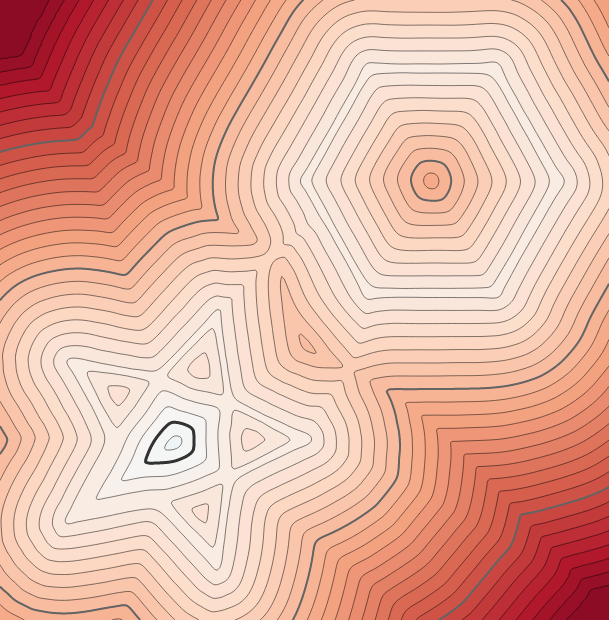} & \includegraphics[width=0.09\textwidth]{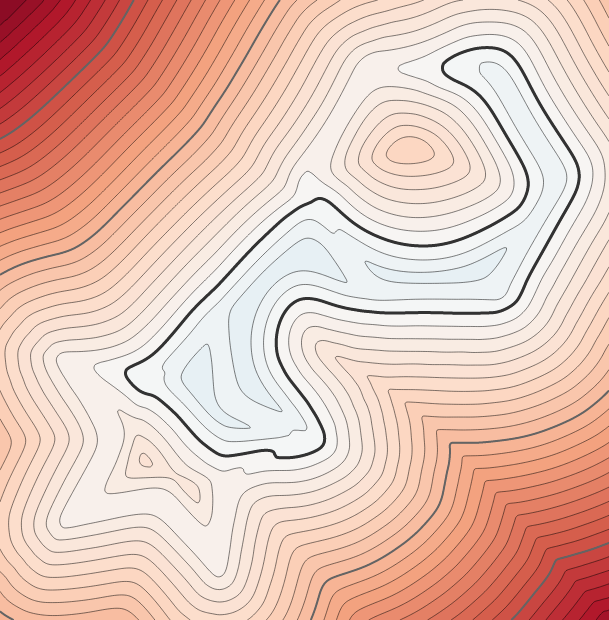} & \includegraphics[width=0.09\textwidth]{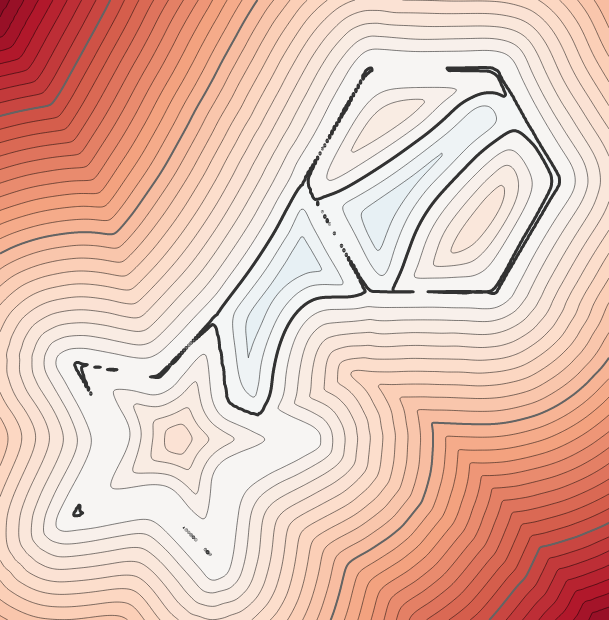} & \includegraphics[width=0.09\textwidth]{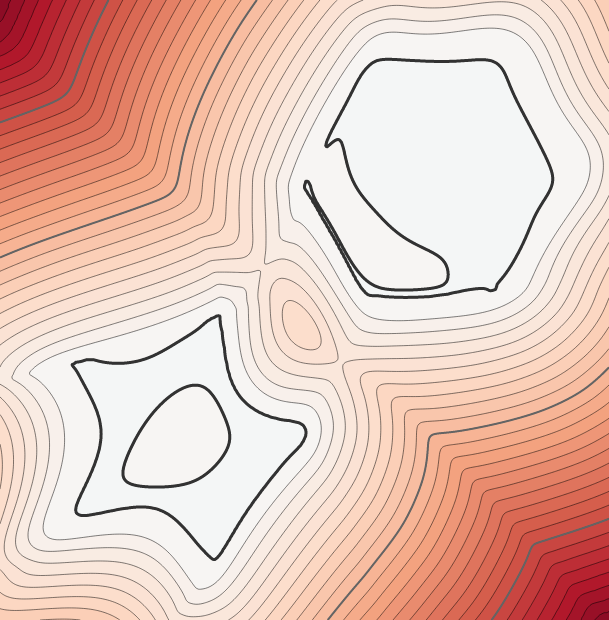} & \includegraphics[width=0.09\textwidth]{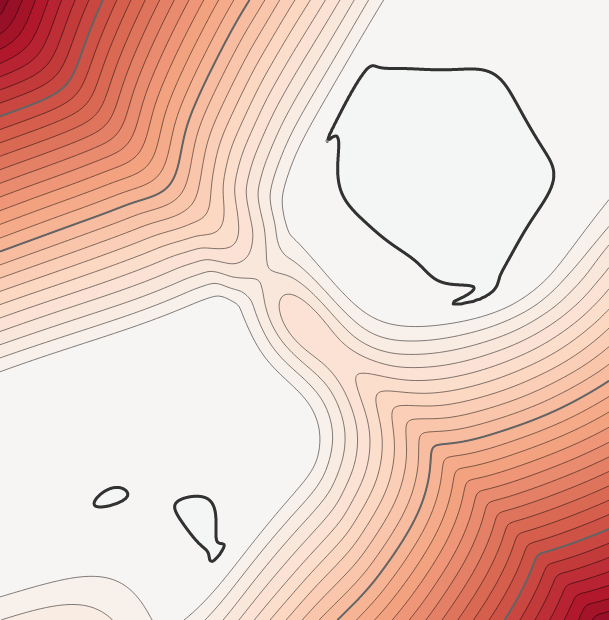} & \includegraphics[width=0.09\textwidth]{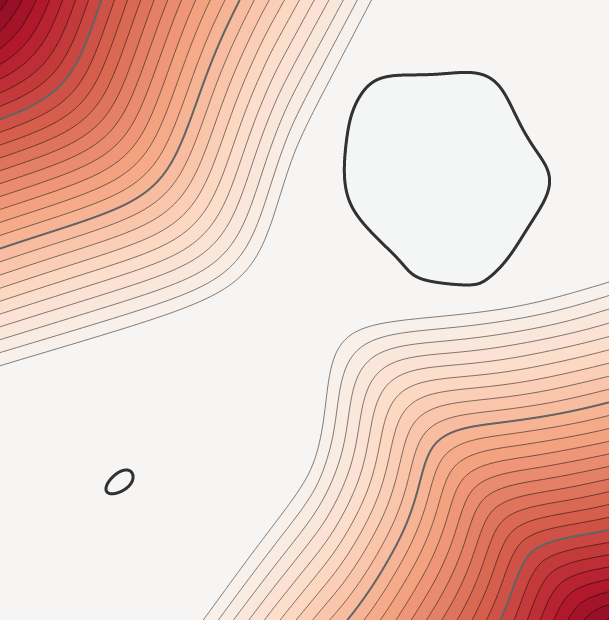} \\
    \includegraphics[width=0.09\textwidth]{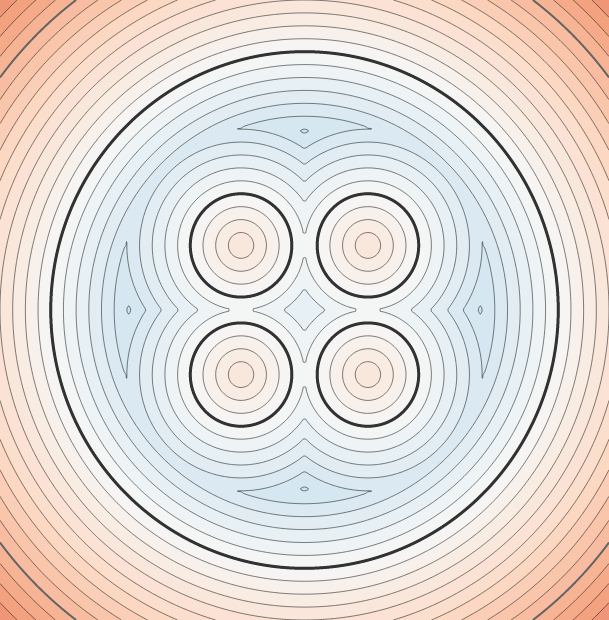} & \includegraphics[width=0.09\textwidth]{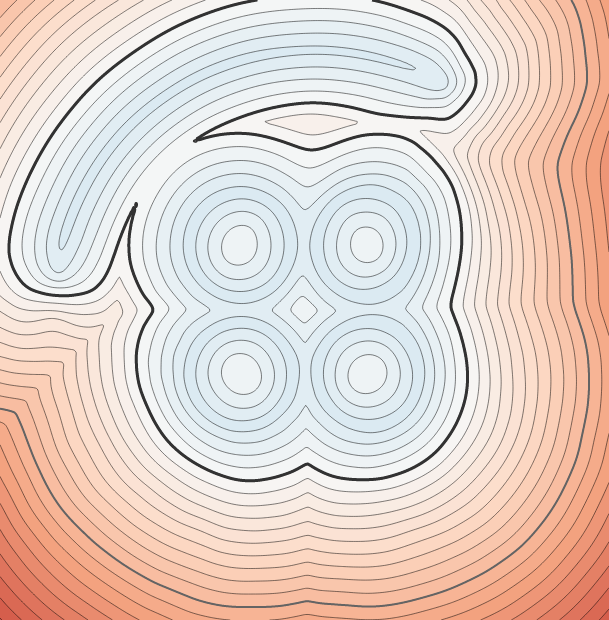} & \includegraphics[width=0.09\textwidth]{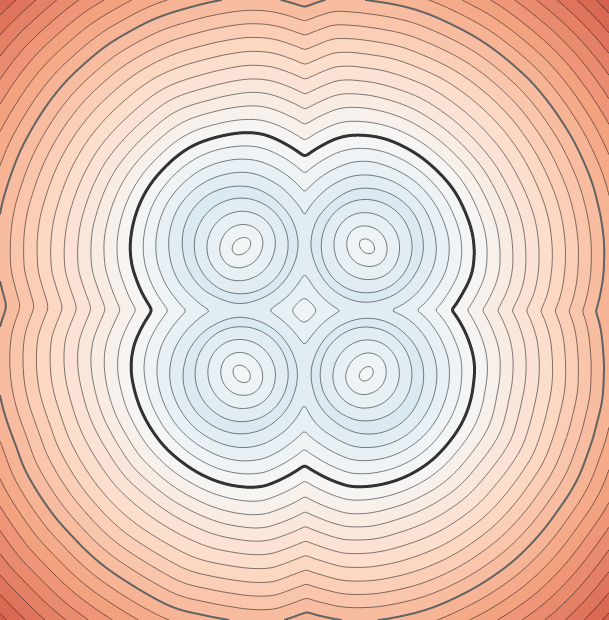} & \includegraphics[width=0.09\textwidth]{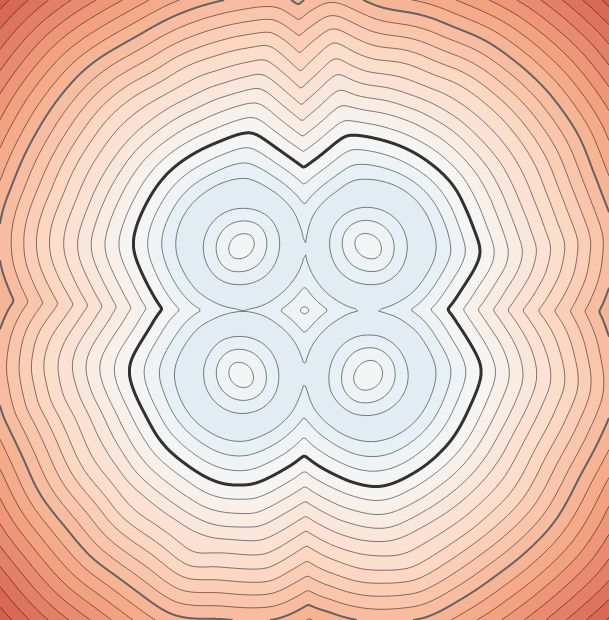} & \includegraphics[width=0.09\textwidth]{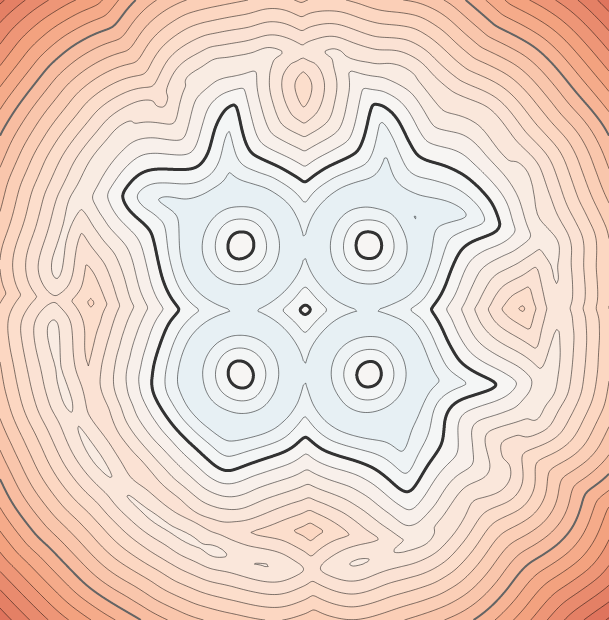} & \includegraphics[width=0.09\textwidth]{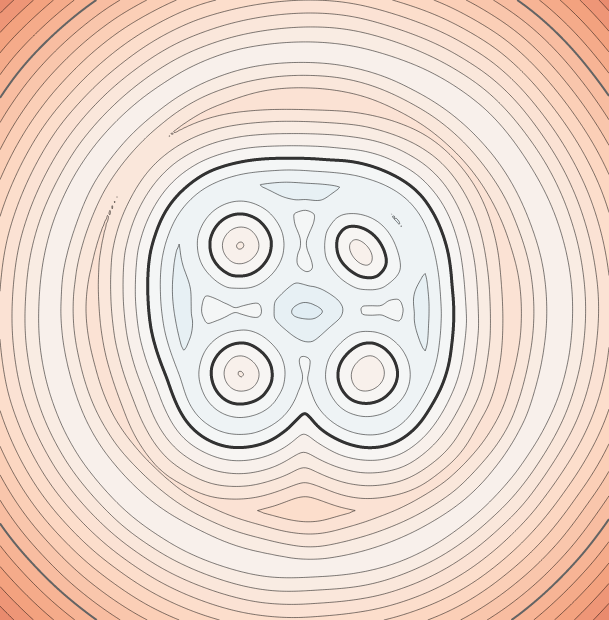} & \includegraphics[width=0.09\textwidth]{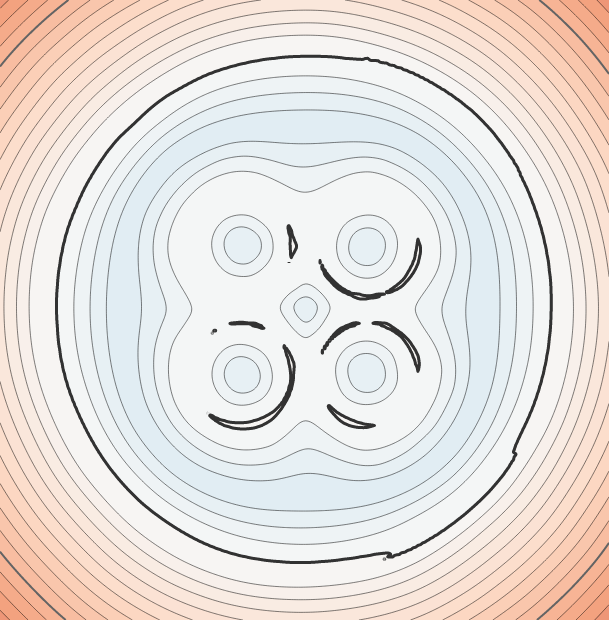} & \includegraphics[width=0.09\textwidth]{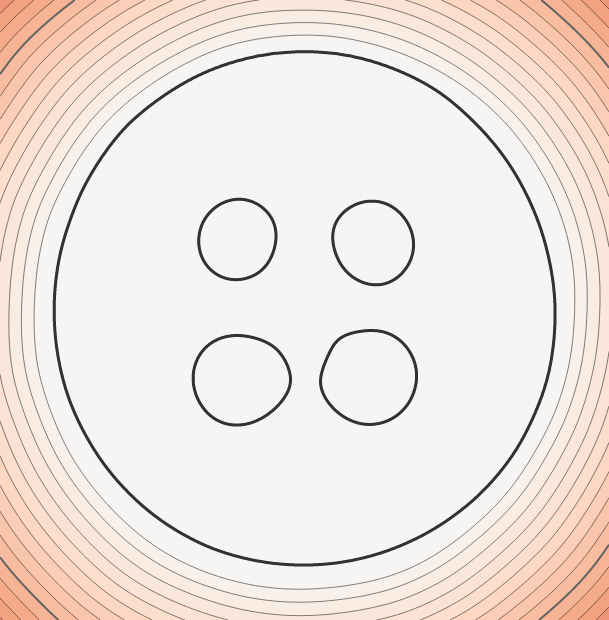} & \includegraphics[width=0.09\textwidth]{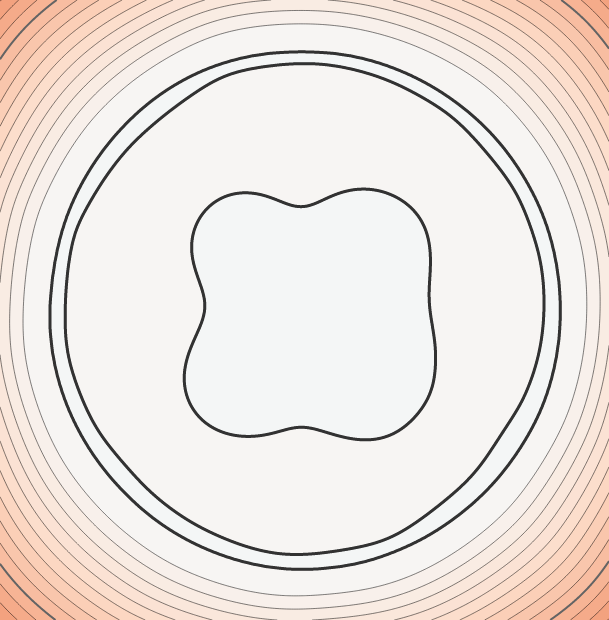} & \includegraphics[width=0.09\textwidth]{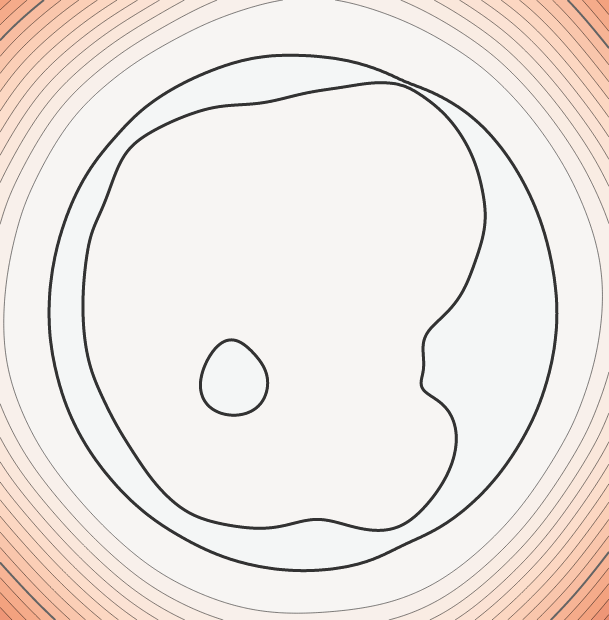} \\
    \includegraphics[width=0.09\textwidth]{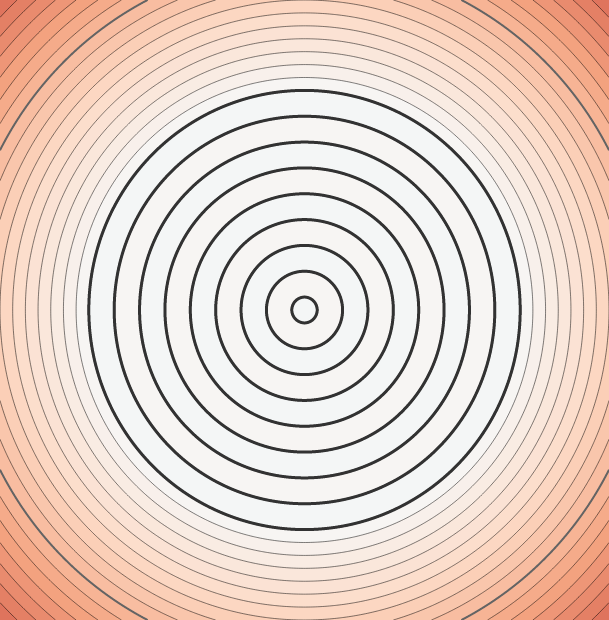} & \includegraphics[width=0.09\textwidth]{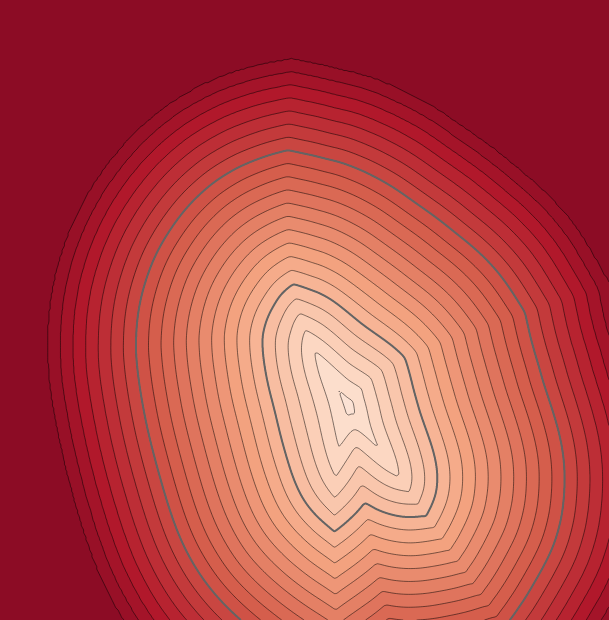} & \includegraphics[width=0.09\textwidth]{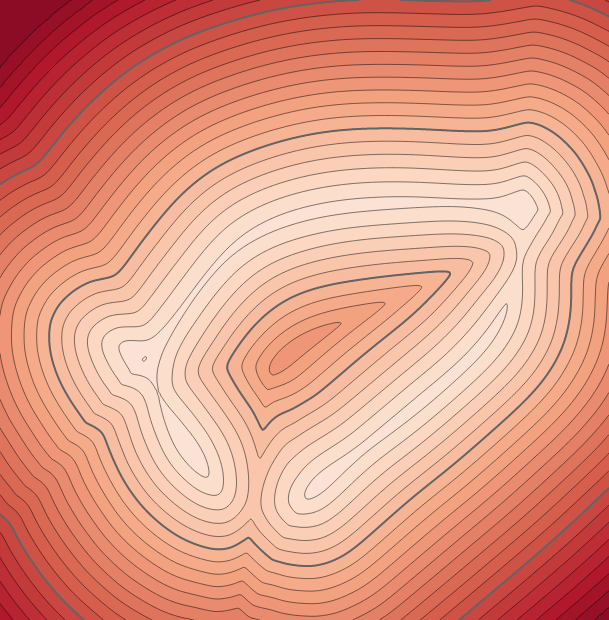} & \includegraphics[width=0.09\textwidth]{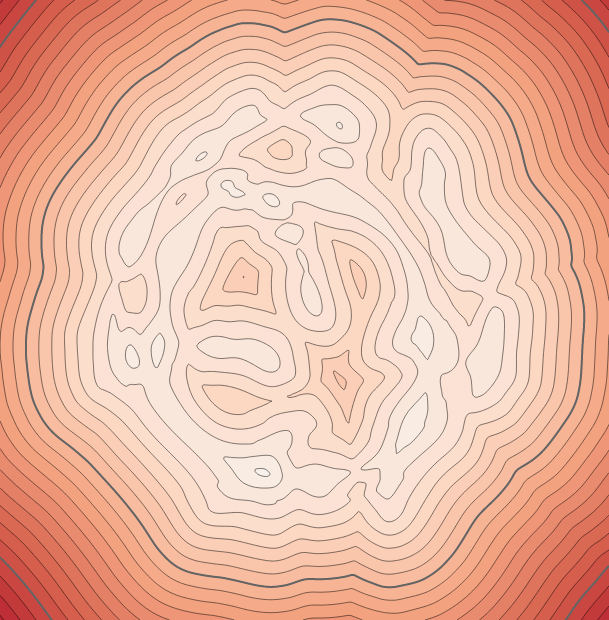} & \includegraphics[width=0.09\textwidth]{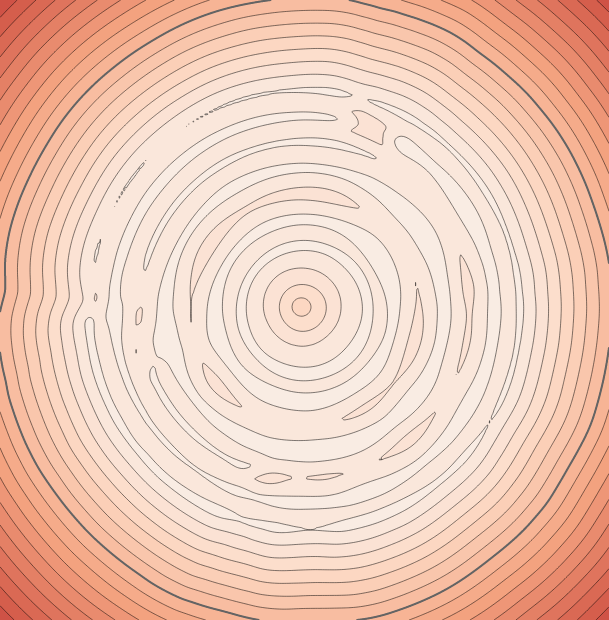} & \includegraphics[width=0.09\textwidth]{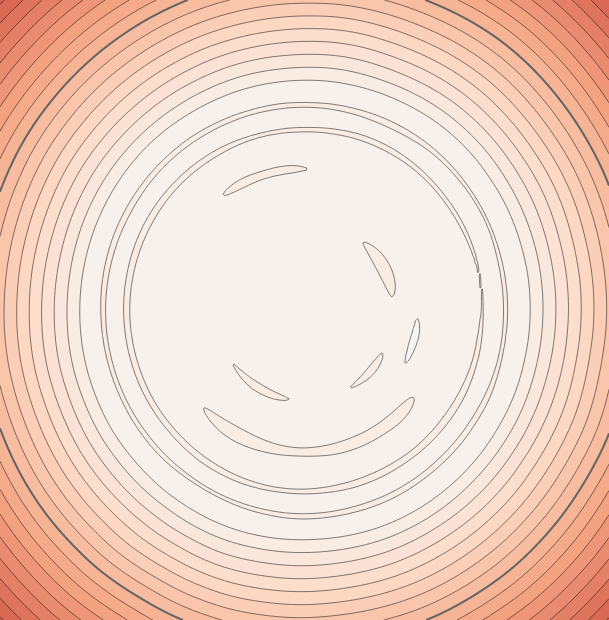} & \includegraphics[width=0.09\textwidth]{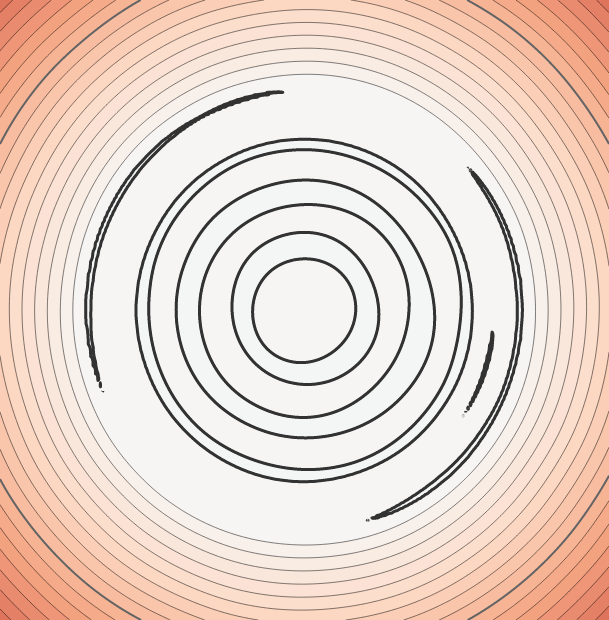} & \includegraphics[width=0.09\textwidth]{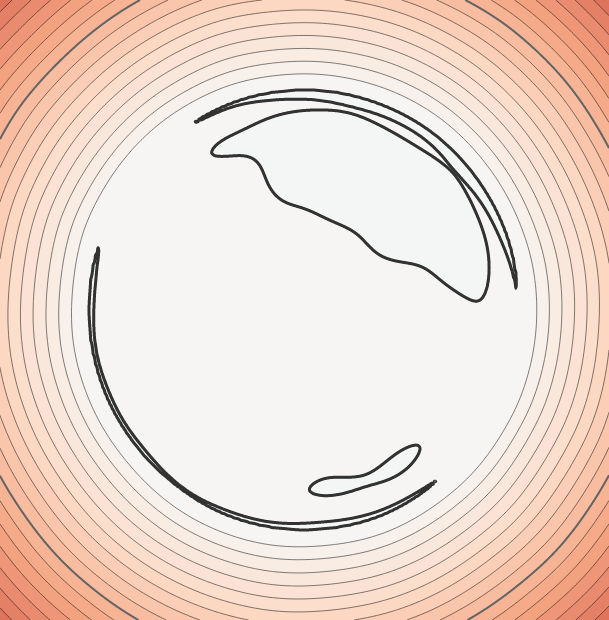} & \includegraphics[width=0.09\textwidth]{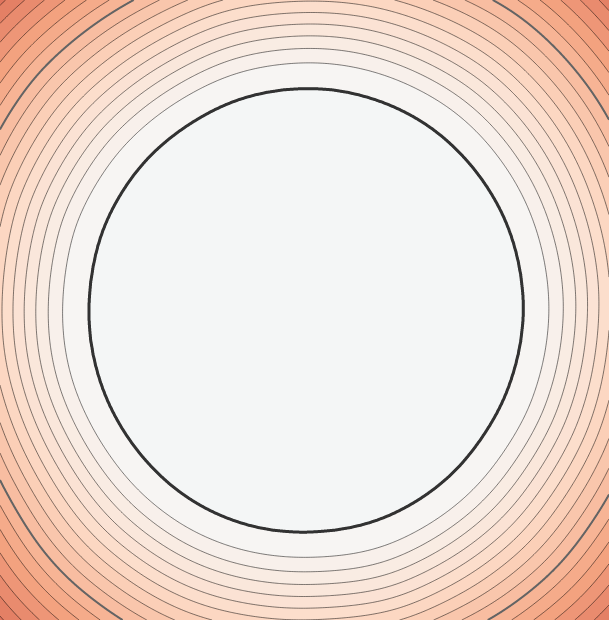} & \includegraphics[width=0.09\textwidth]{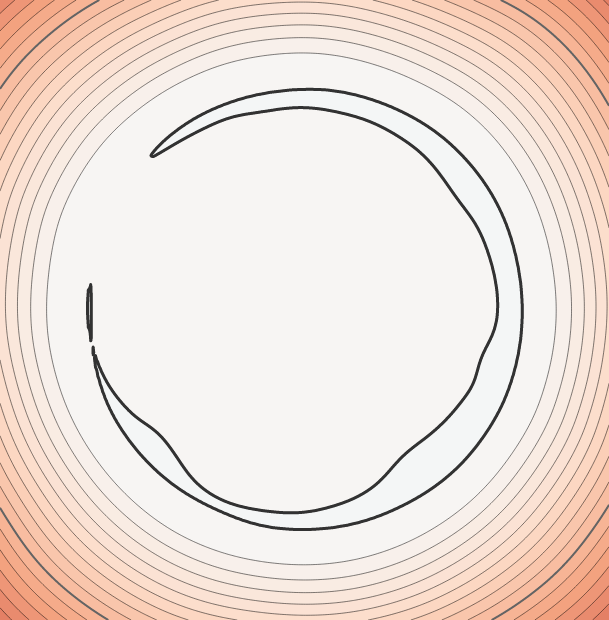} \\
    \includegraphics[width=0.09\textwidth]{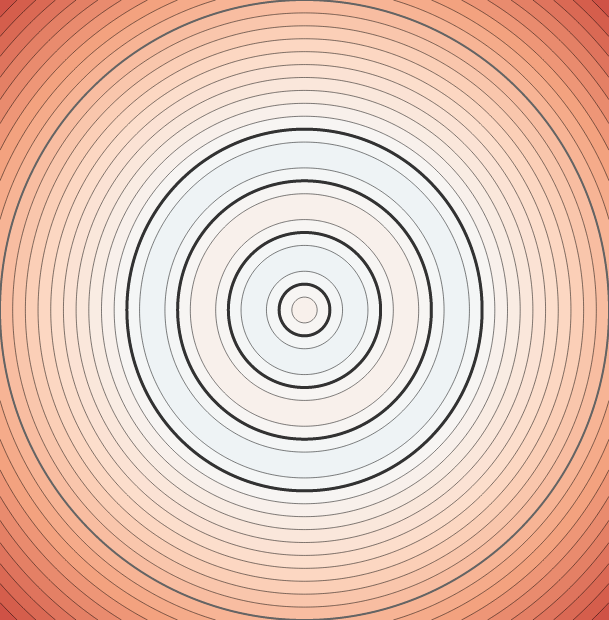} & \includegraphics[width=0.09\textwidth]{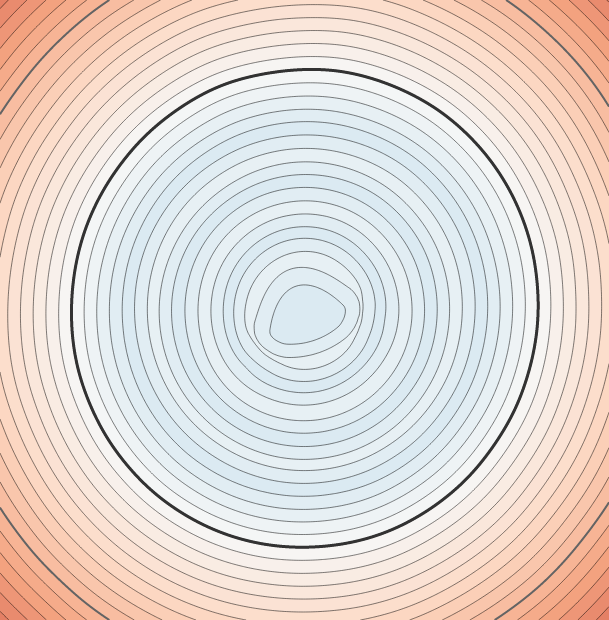} & \includegraphics[width=0.09\textwidth]{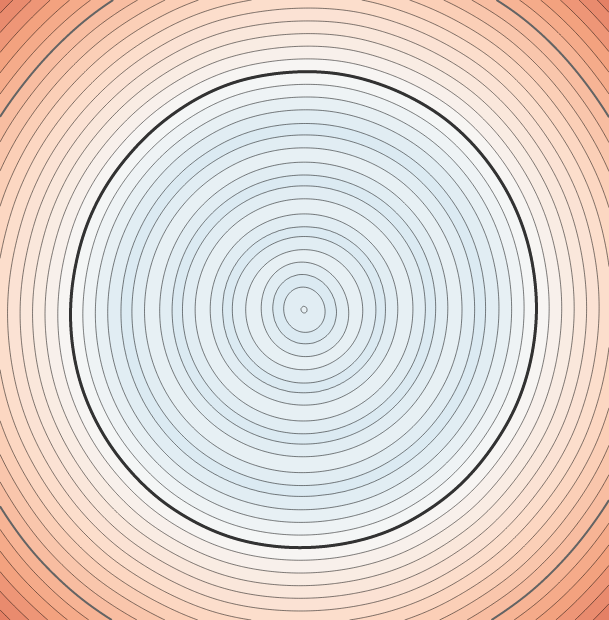} & \includegraphics[width=0.09\textwidth]{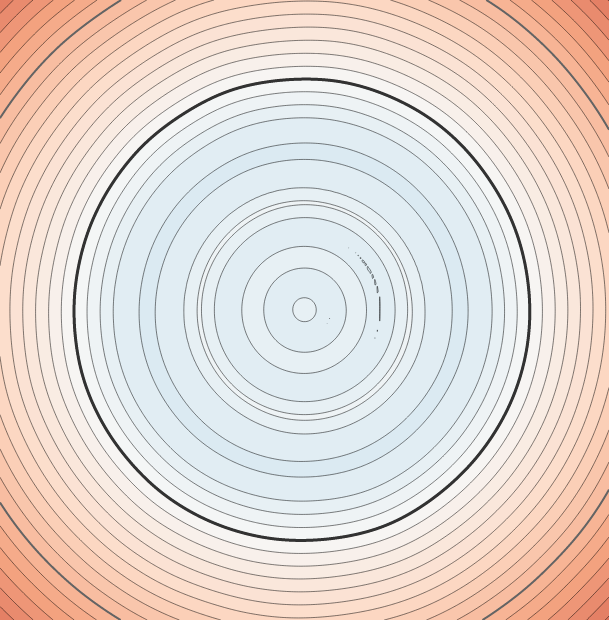} & \includegraphics[width=0.09\textwidth]{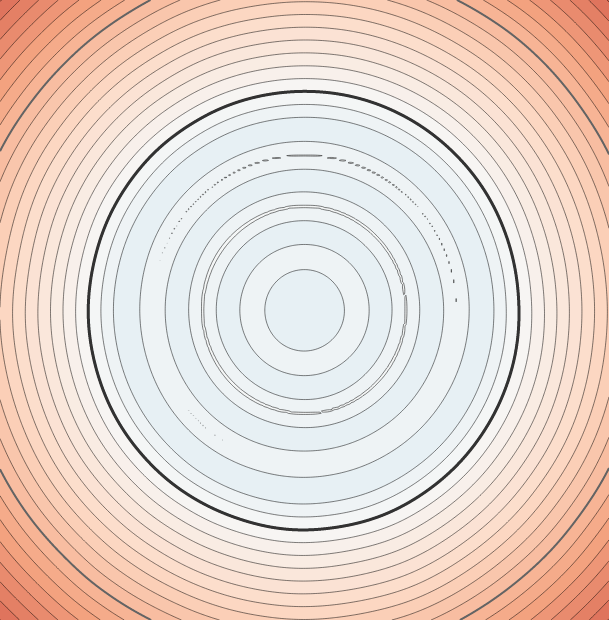} & \includegraphics[width=0.09\textwidth]{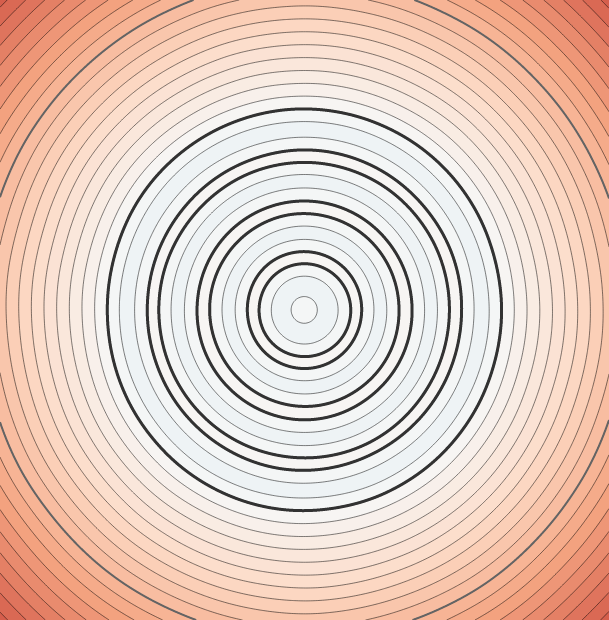} & \includegraphics[width=0.09\textwidth]{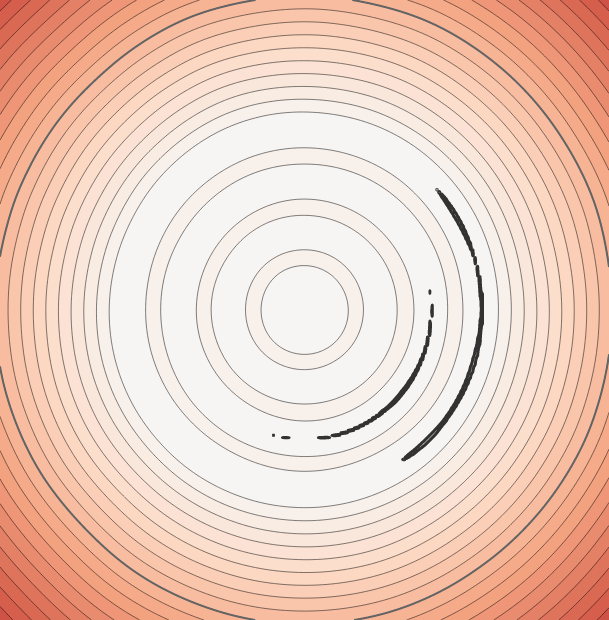} & \includegraphics[width=0.09\textwidth]{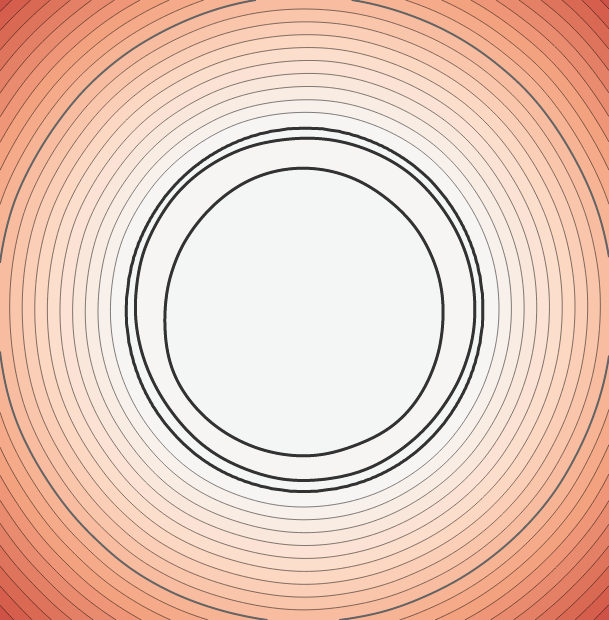} & \includegraphics[width=0.09\textwidth]{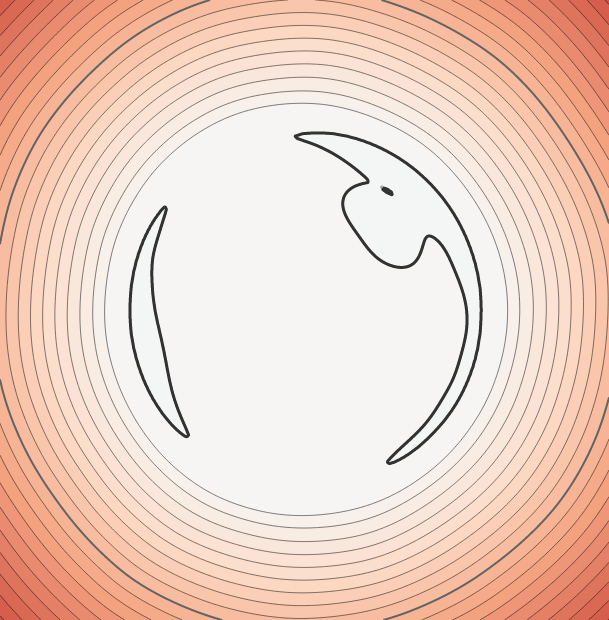} & \includegraphics[width=0.09\textwidth]{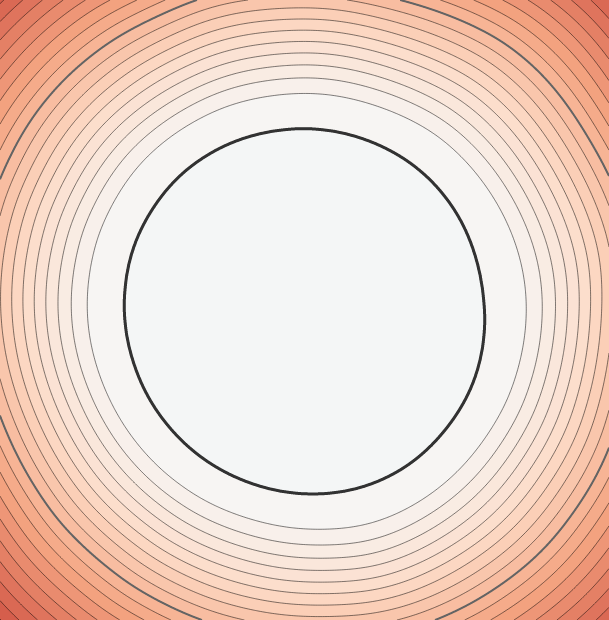} \\
    \includegraphics[width=0.09\textwidth]{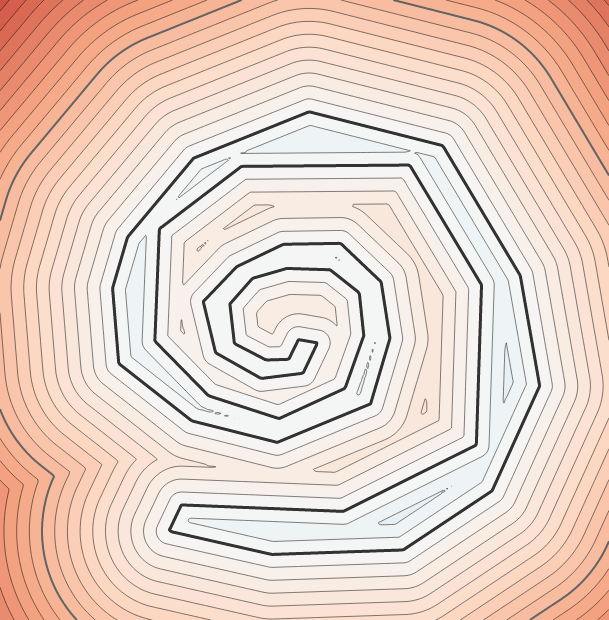} & \includegraphics[width=0.09\textwidth]{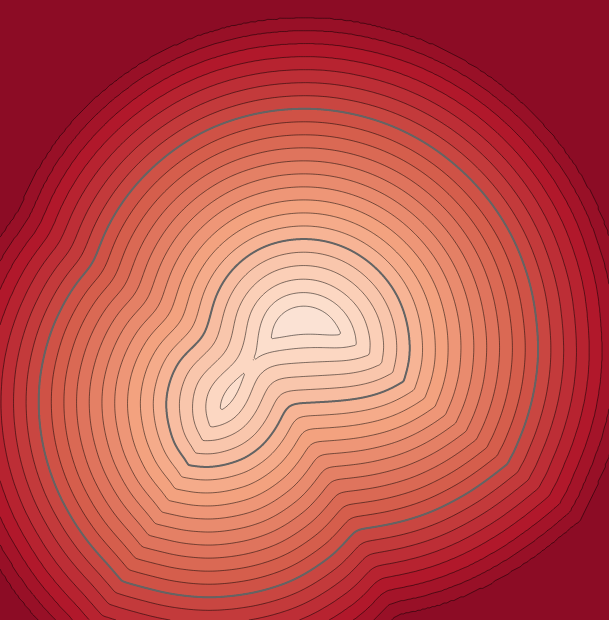} & \includegraphics[width=0.09\textwidth]{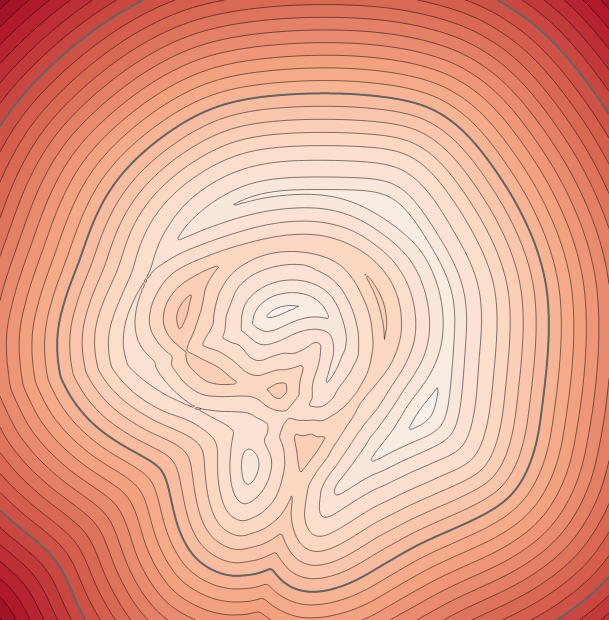} & \includegraphics[width=0.09\textwidth]{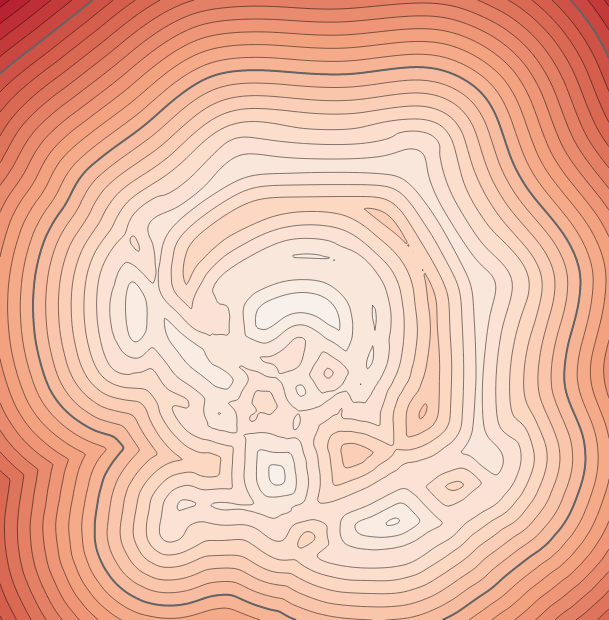} & \includegraphics[width=0.09\textwidth]{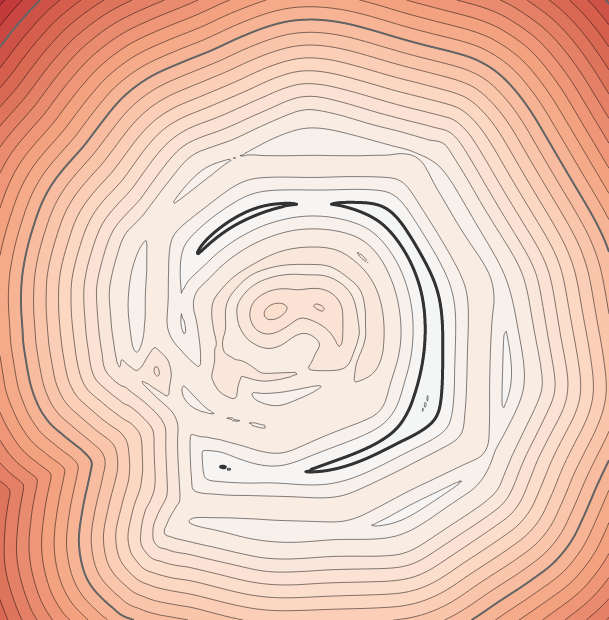} & \includegraphics[width=0.09\textwidth]{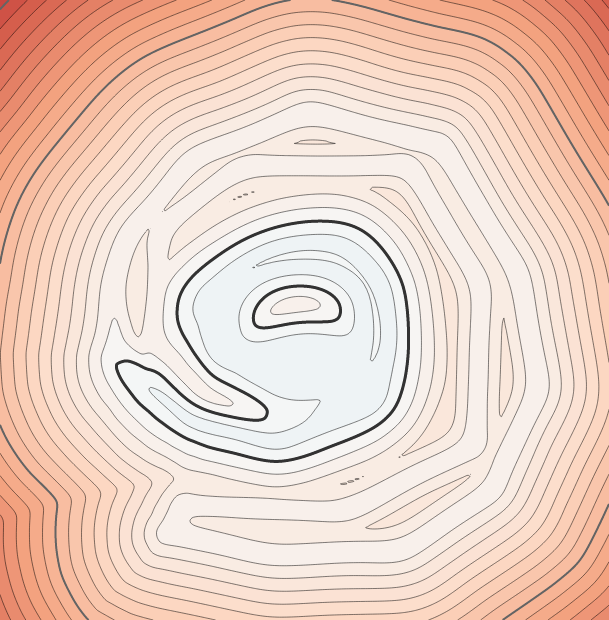} & \includegraphics[width=0.09\textwidth]{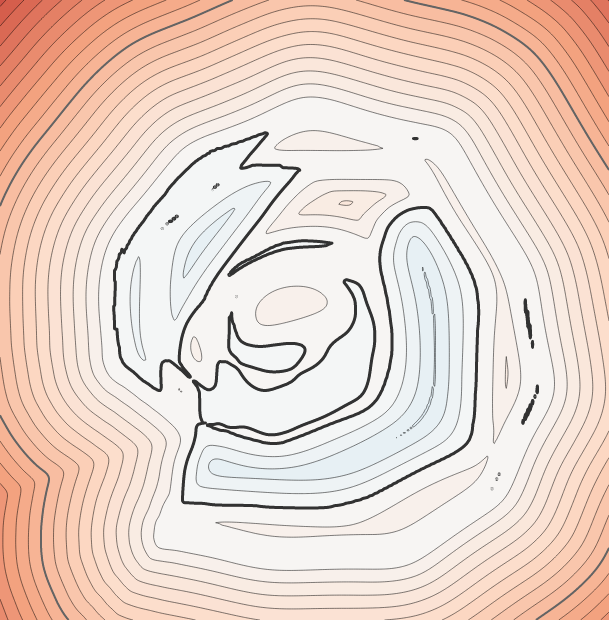} & \includegraphics[width=0.09\textwidth]{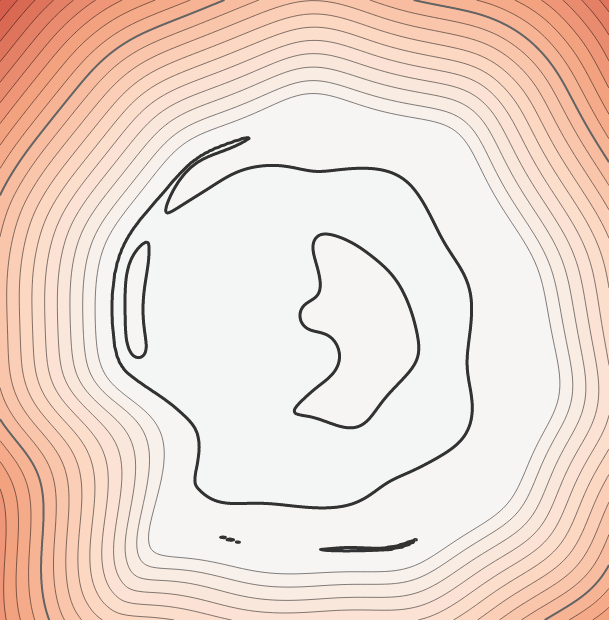} & \includegraphics[width=0.09\textwidth]{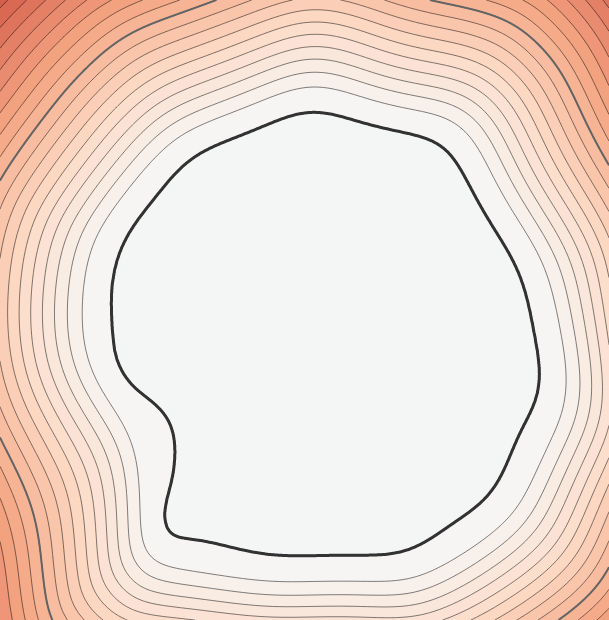} & \includegraphics[width=0.09\textwidth]{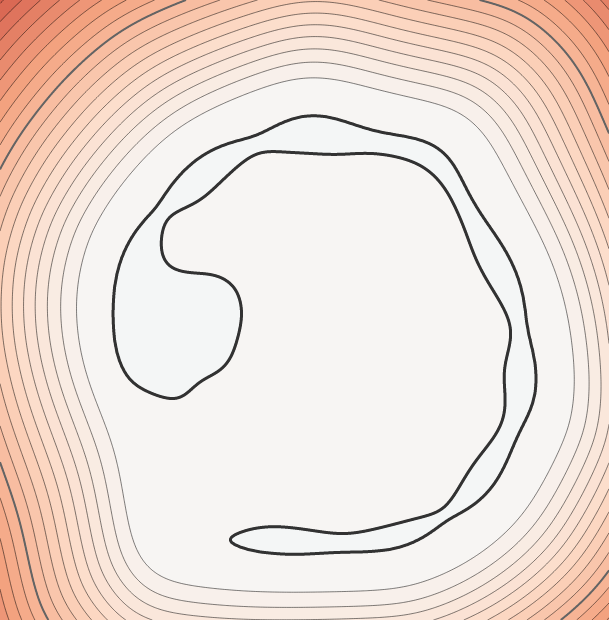} \\
    \includegraphics[width=0.09\textwidth]{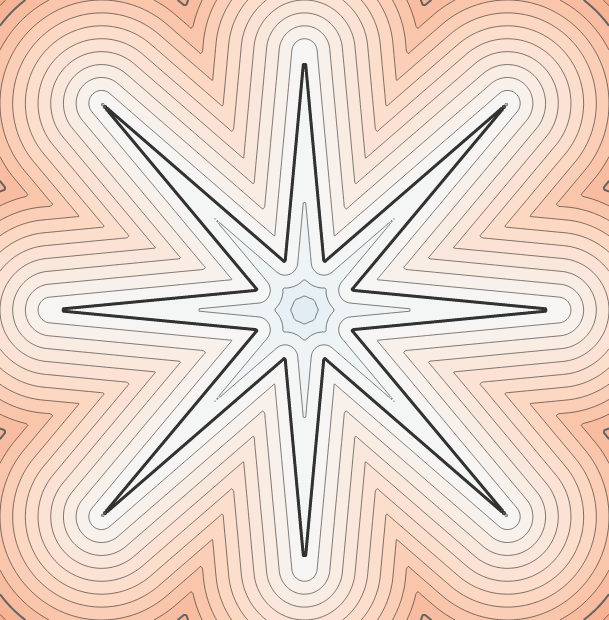} & \includegraphics[width=0.09\textwidth]{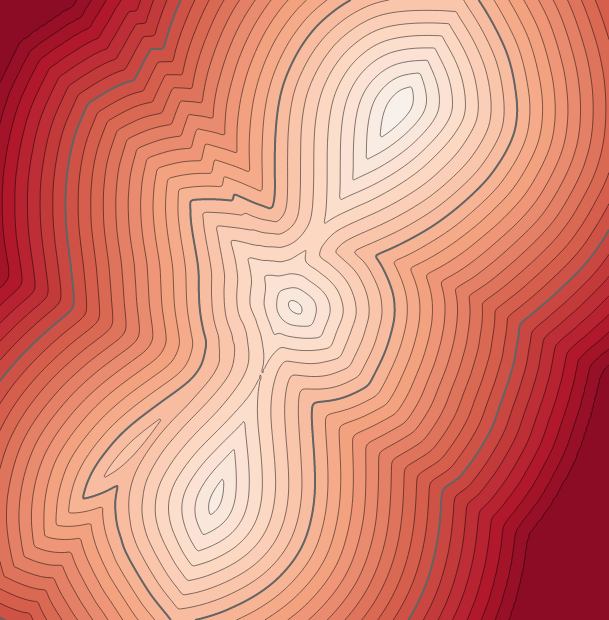} & \includegraphics[width=0.09\textwidth]{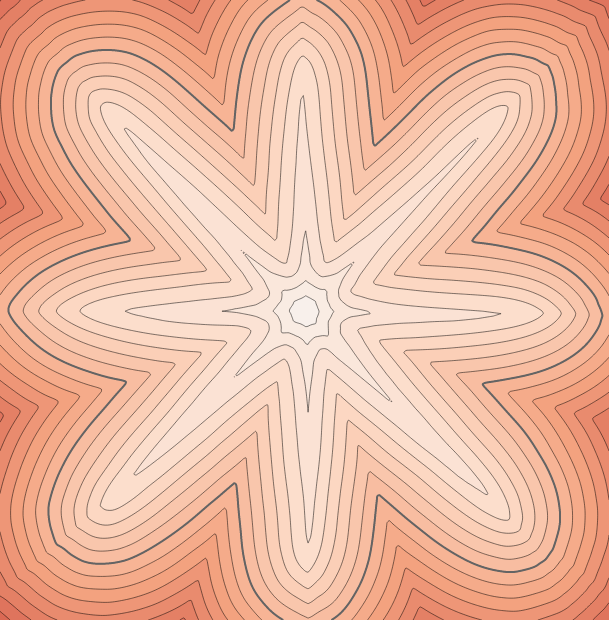} & \includegraphics[width=0.09\textwidth]{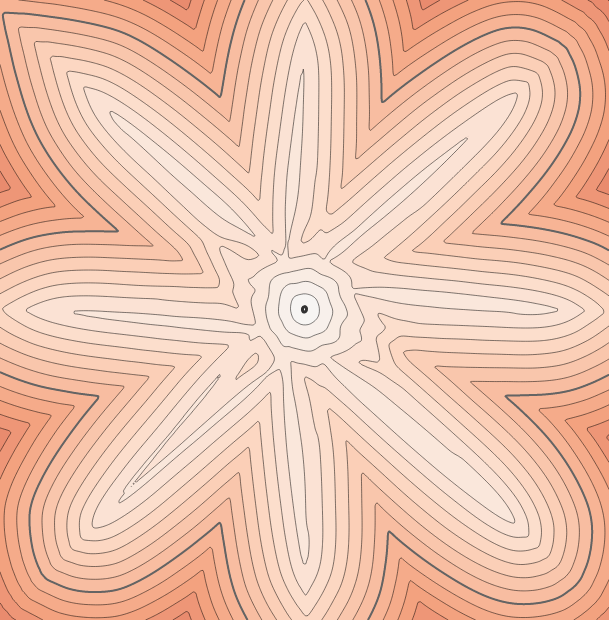} & \includegraphics[width=0.09\textwidth]{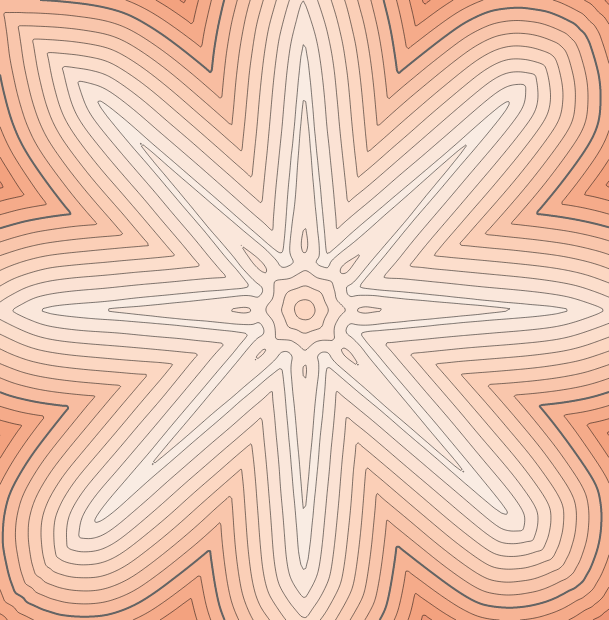} & \includegraphics[width=0.09\textwidth]{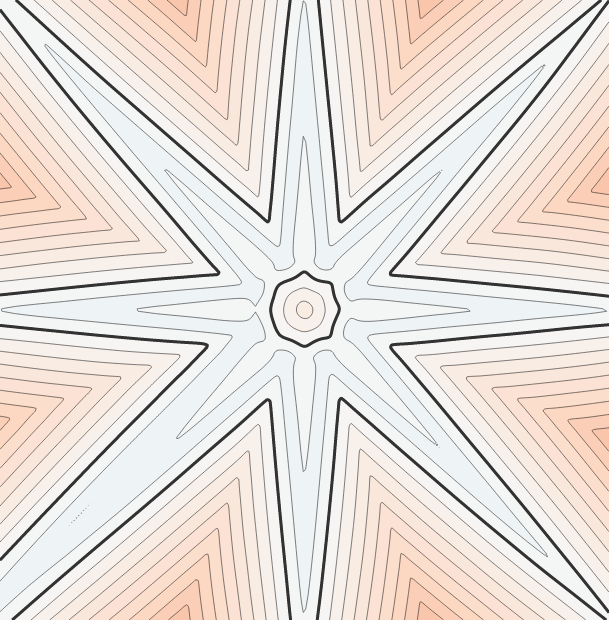} & \includegraphics[width=0.09\textwidth]{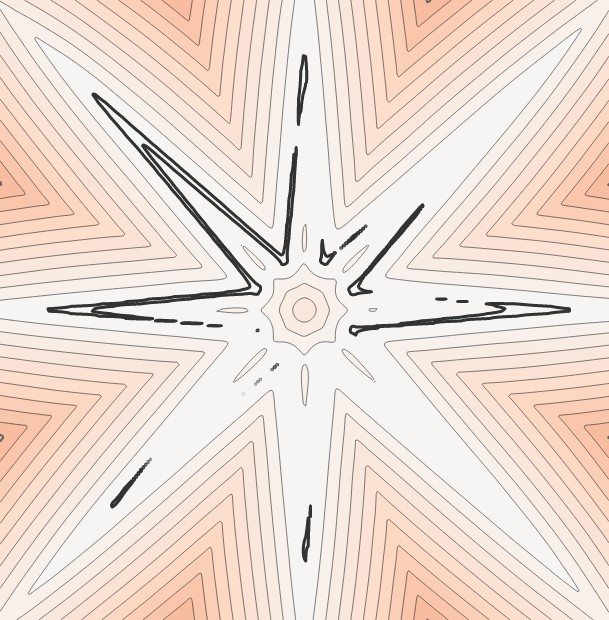} & \includegraphics[width=0.09\textwidth]{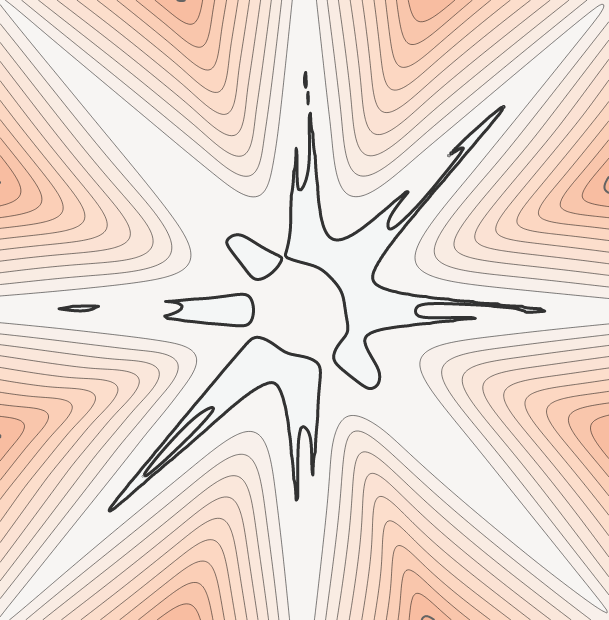} & \includegraphics[width=0.09\textwidth]{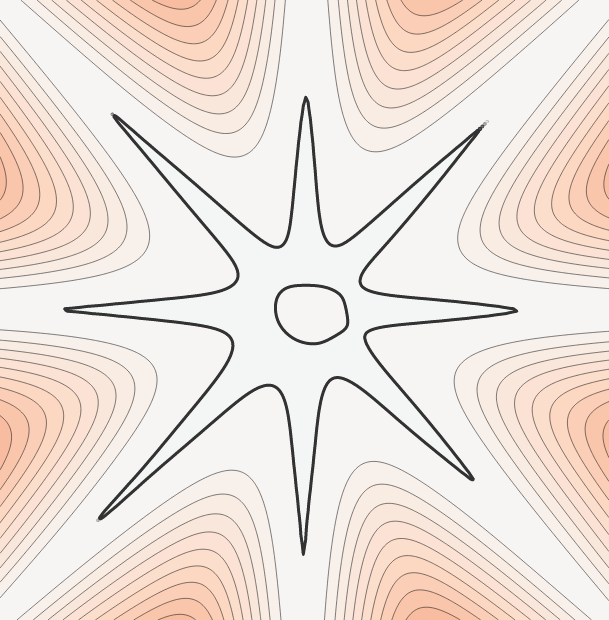} & \includegraphics[width=0.09\textwidth]{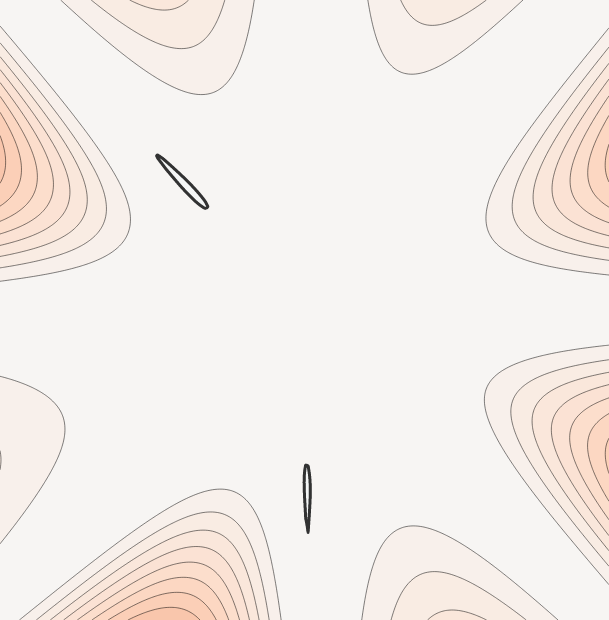} \\
    \includegraphics[width=0.09\textwidth]{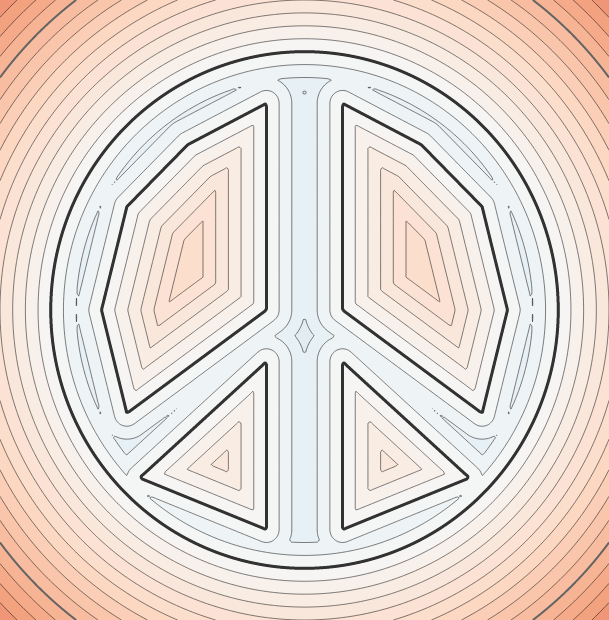} & \includegraphics[width=0.09\textwidth]{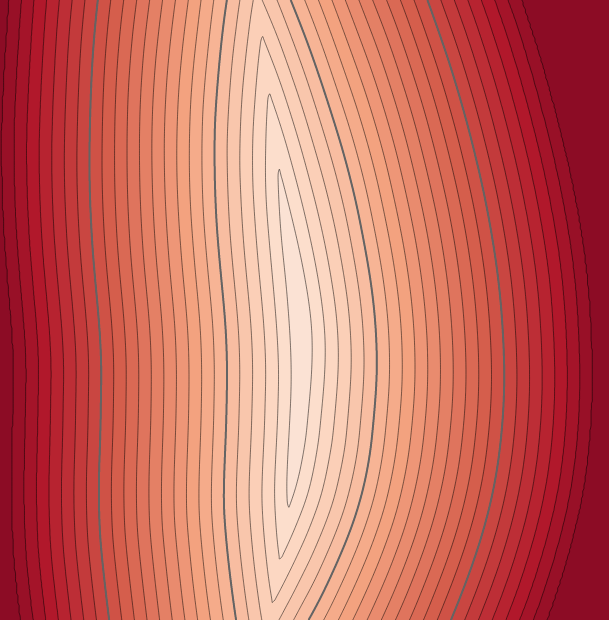} & \includegraphics[width=0.09\textwidth]{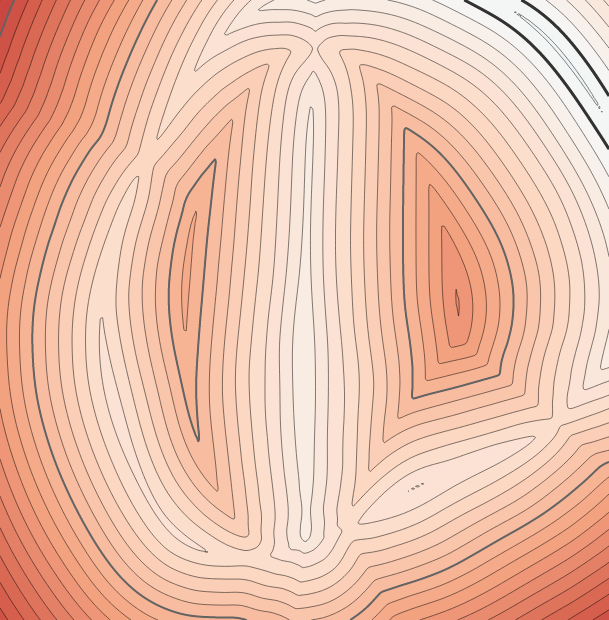} & \includegraphics[width=0.09\textwidth]{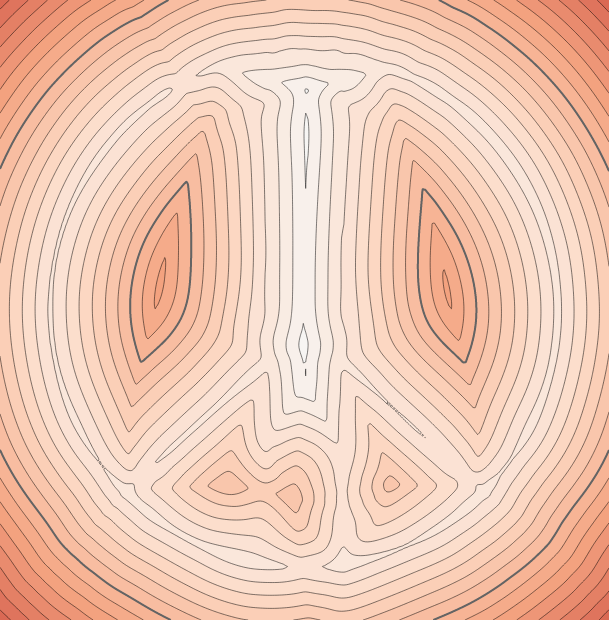} & \includegraphics[width=0.09\textwidth]{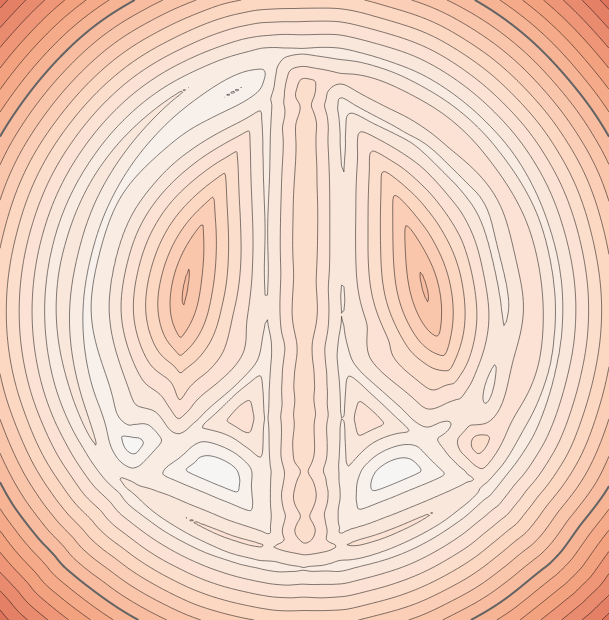} & \includegraphics[width=0.09\textwidth]{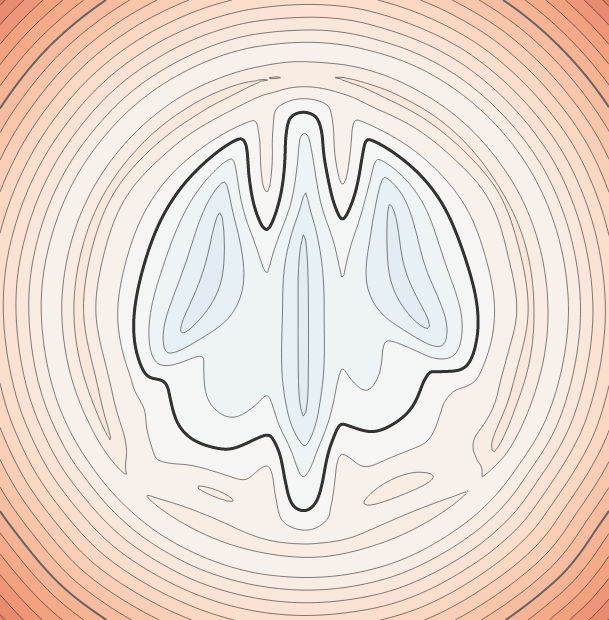} & \includegraphics[width=0.09\textwidth]{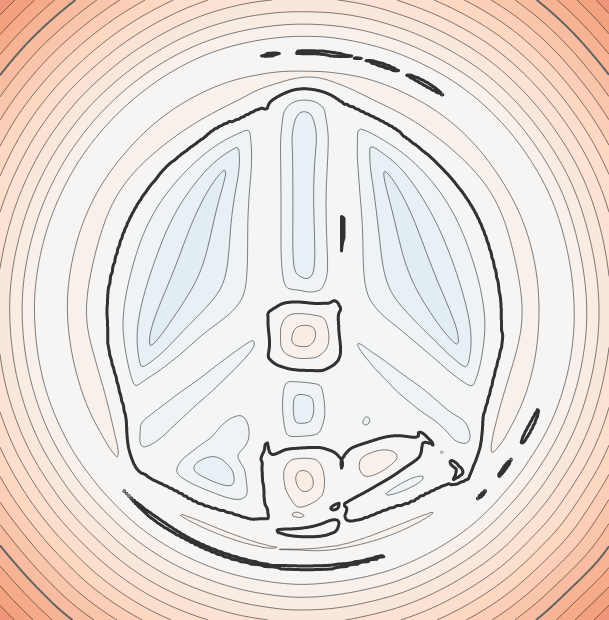} & \includegraphics[width=0.09\textwidth]{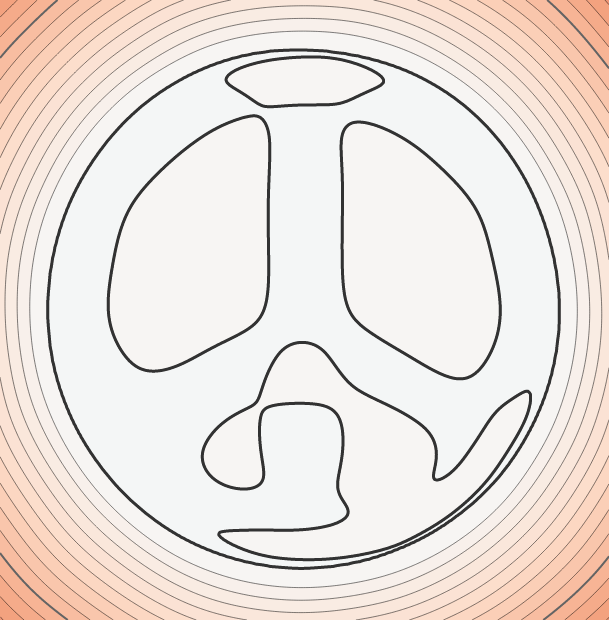} & \includegraphics[width=0.09\textwidth]{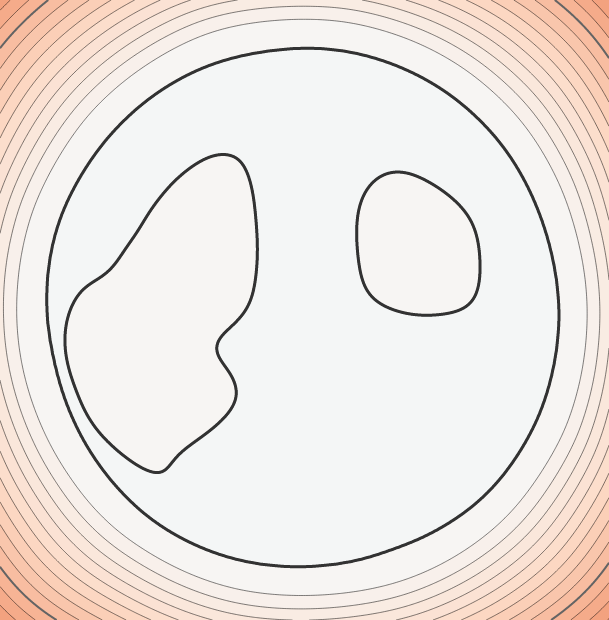} & \includegraphics[width=0.09\textwidth]{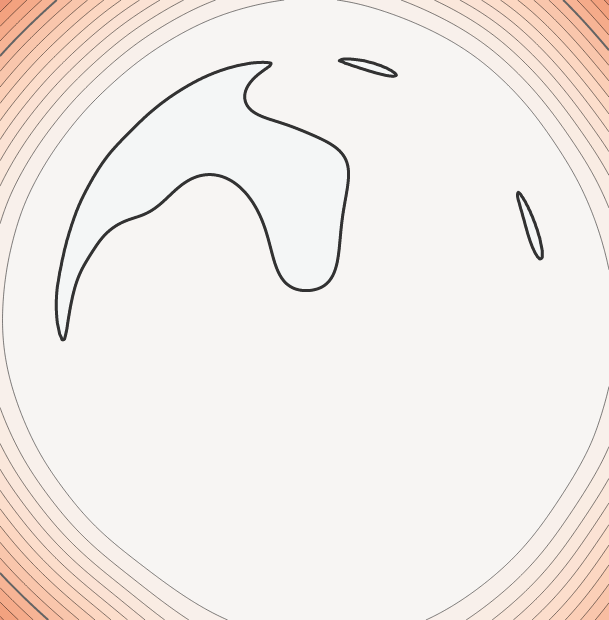} \\
    \includegraphics[width=0.09\textwidth]{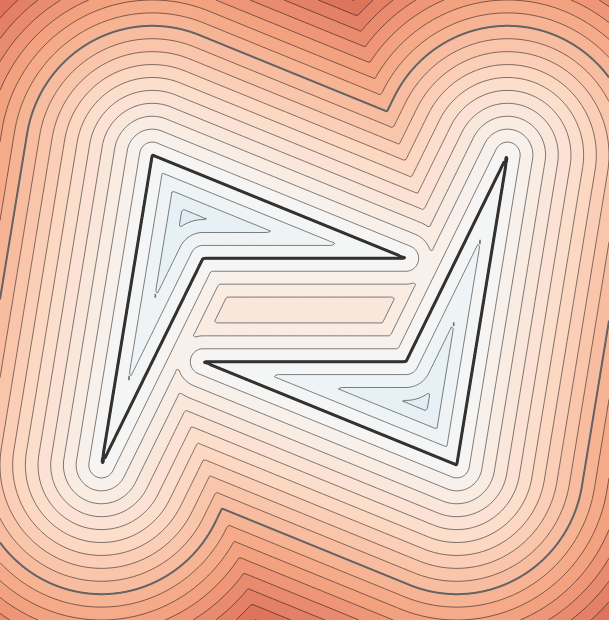} & \includegraphics[width=0.09\textwidth]{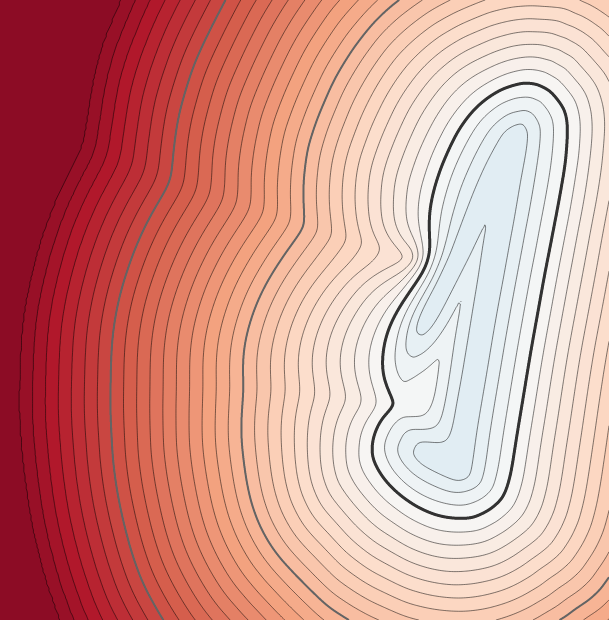} & \includegraphics[width=0.09\textwidth]{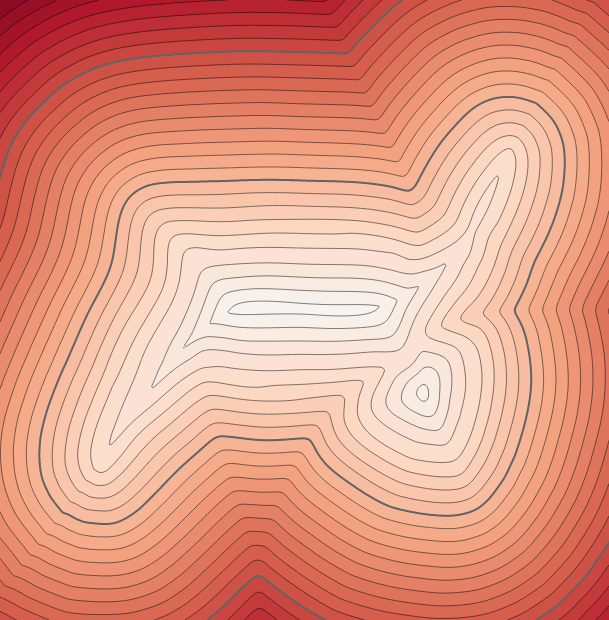} & \includegraphics[width=0.09\textwidth]{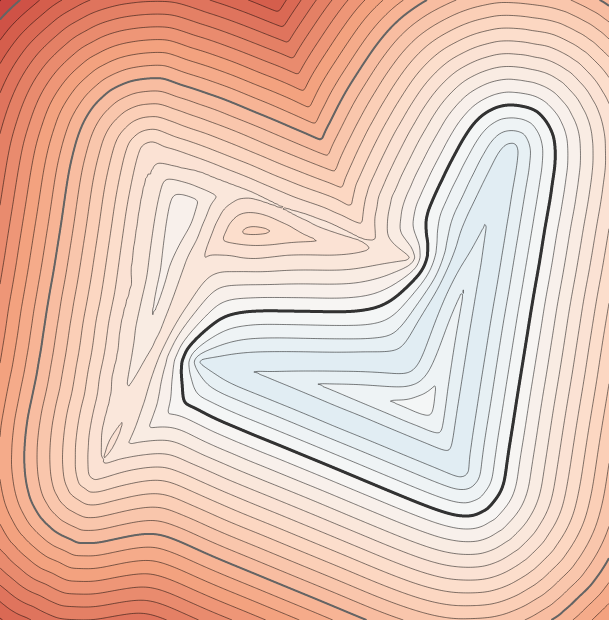} & \includegraphics[width=0.09\textwidth]{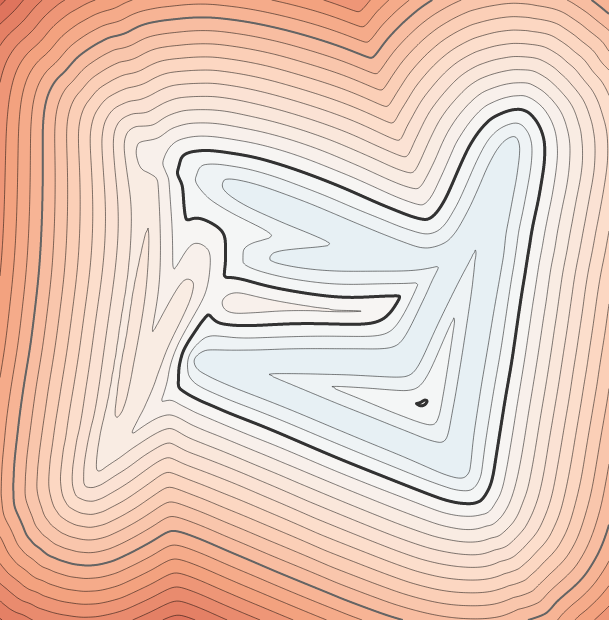} & \includegraphics[width=0.09\textwidth]{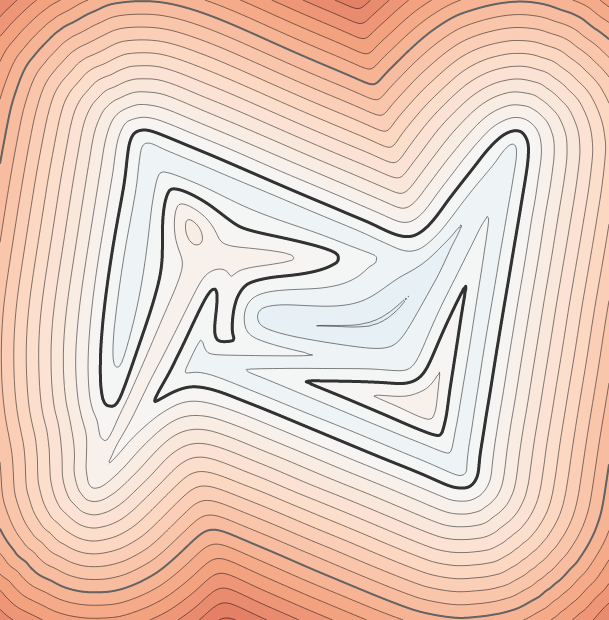} & \includegraphics[width=0.09\textwidth]{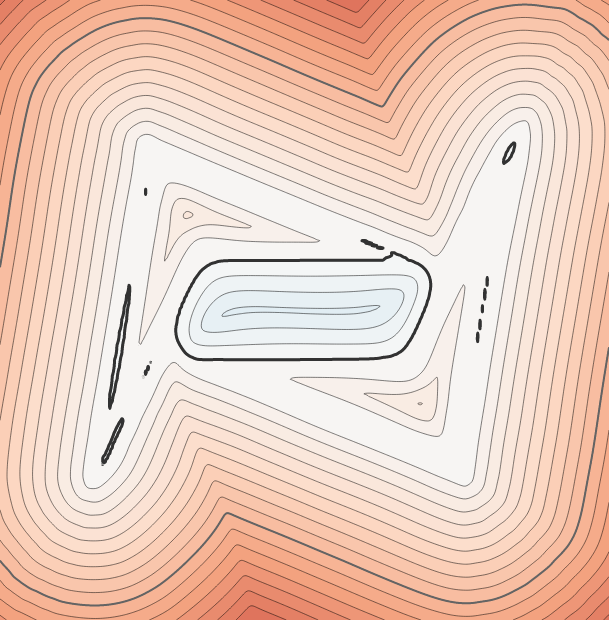} & \includegraphics[width=0.09\textwidth]{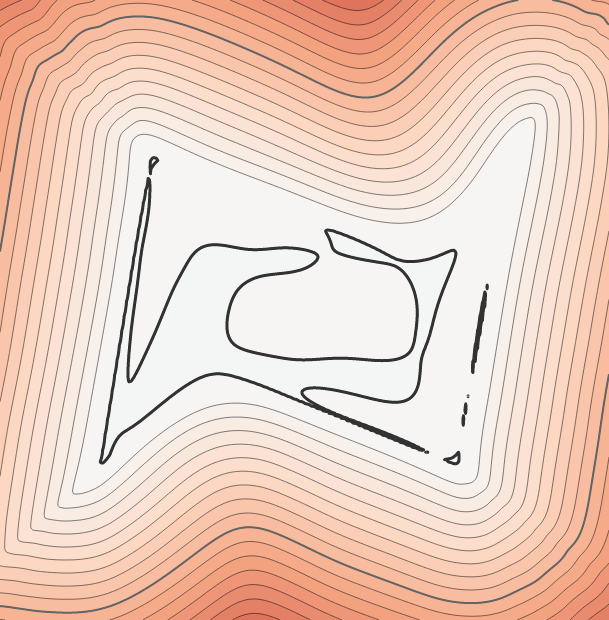} & \includegraphics[width=0.09\textwidth]{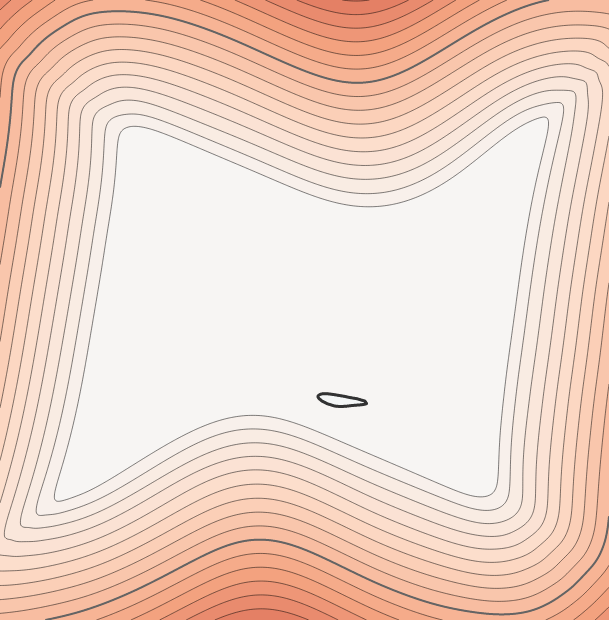} & \includegraphics[width=0.09\textwidth]{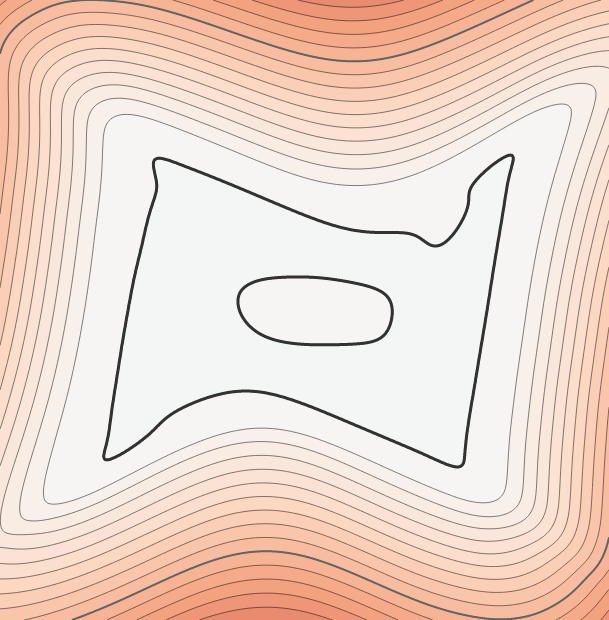} \\
    \includegraphics[width=0.09\textwidth]{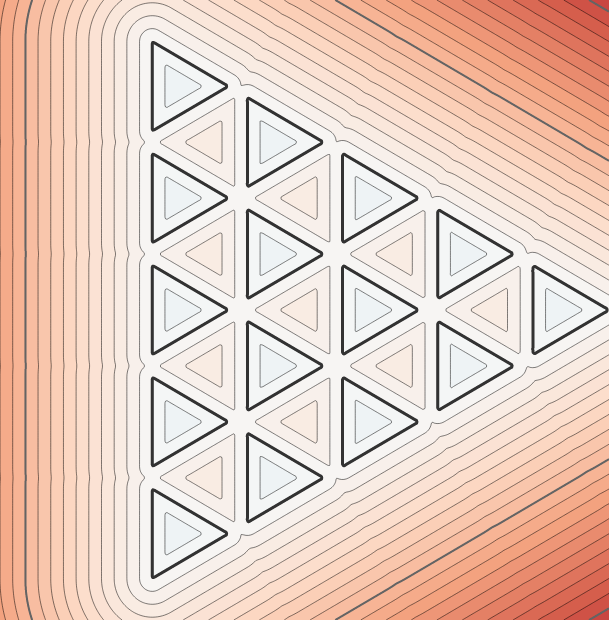} & \includegraphics[width=0.09\textwidth]{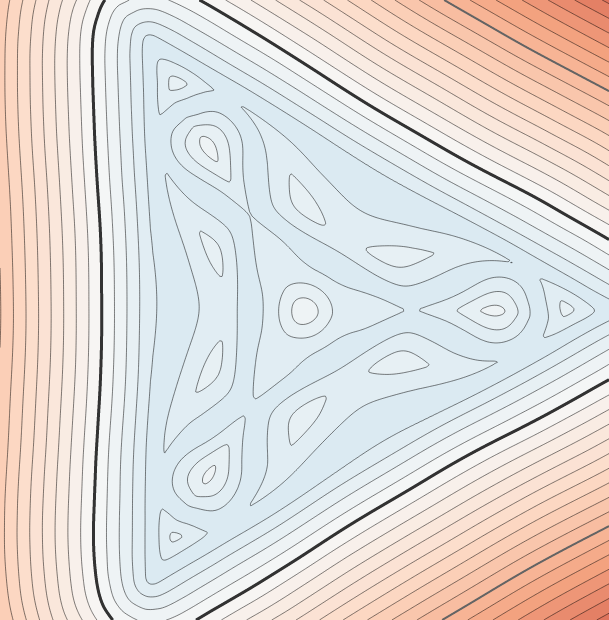} & \includegraphics[width=0.09\textwidth]{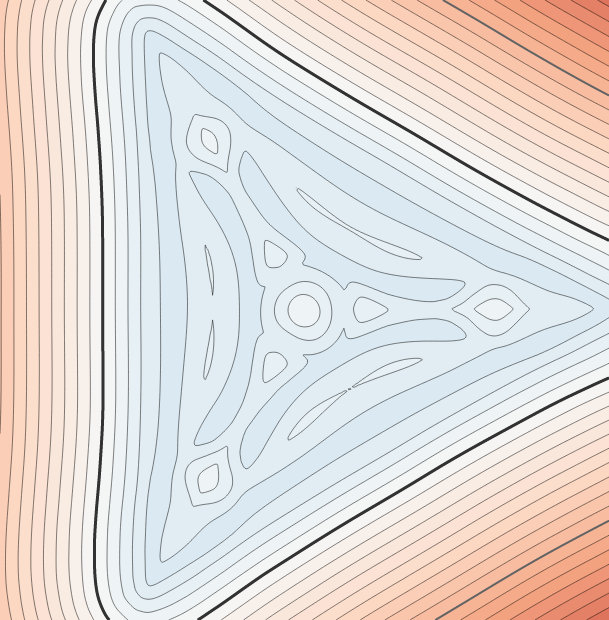} & \includegraphics[width=0.09\textwidth]{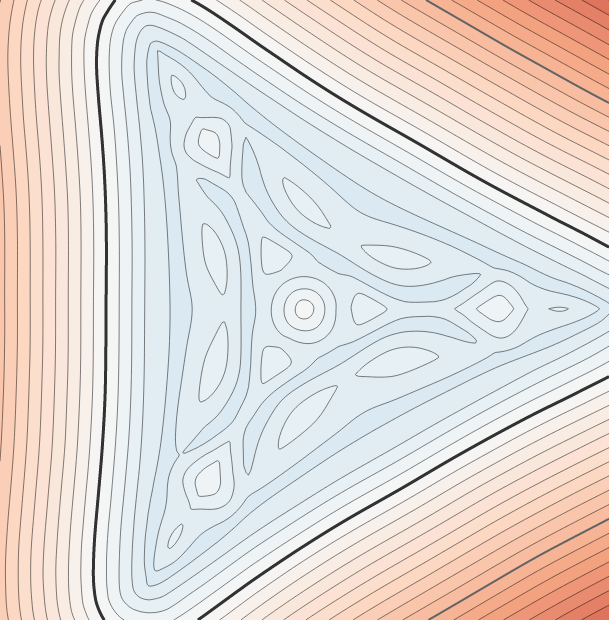} & \includegraphics[width=0.09\textwidth]{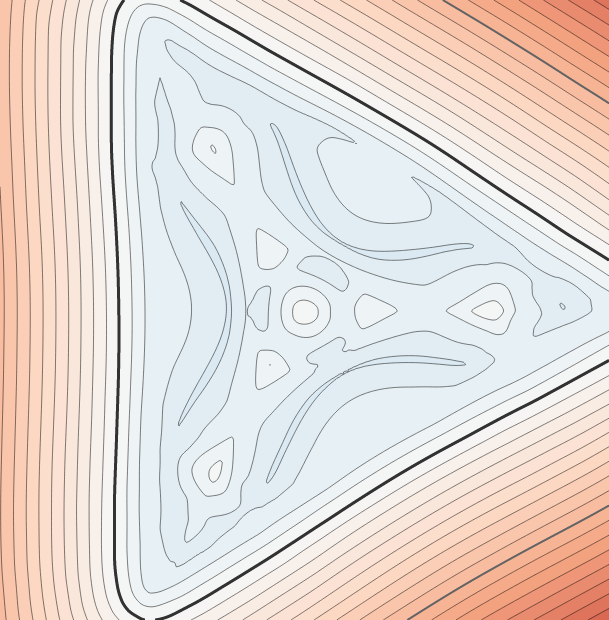} & \includegraphics[width=0.09\textwidth]{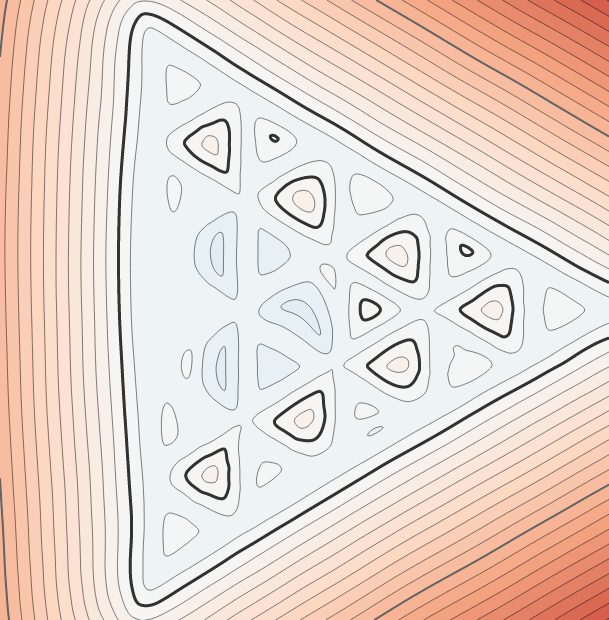} & \includegraphics[width=0.09\textwidth]{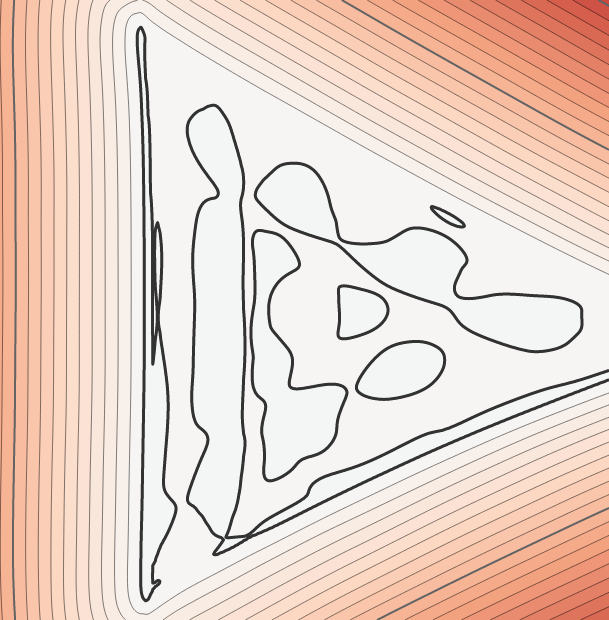} & \includegraphics[width=0.09\textwidth]{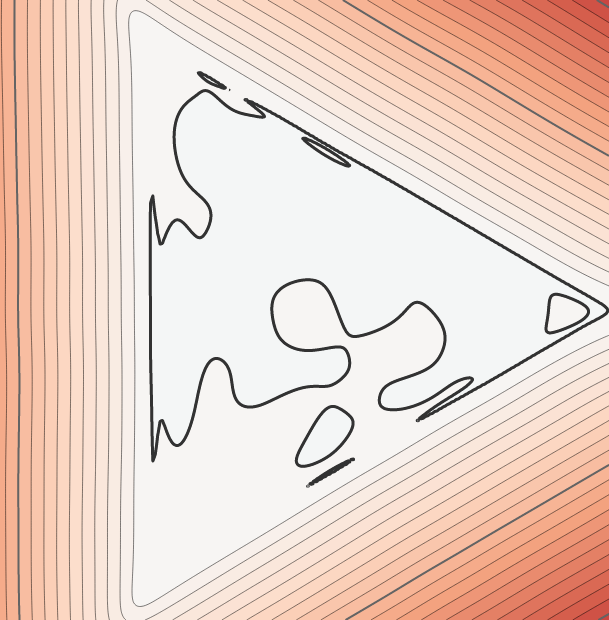} & \includegraphics[width=0.09\textwidth]{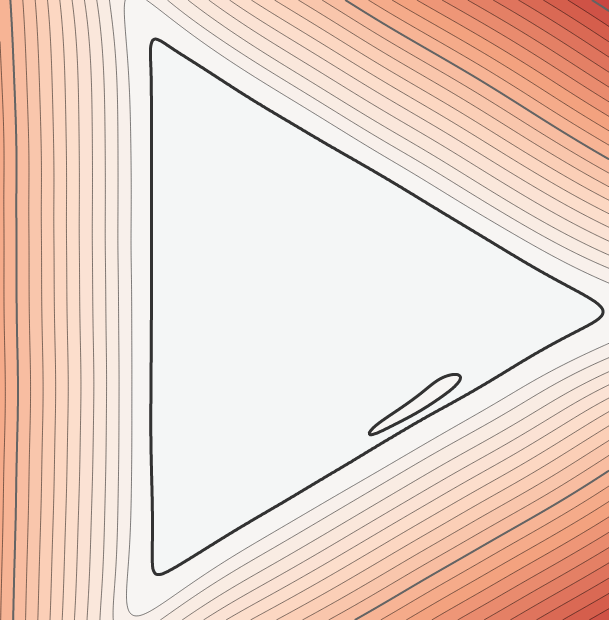} & \includegraphics[width=0.09\textwidth]{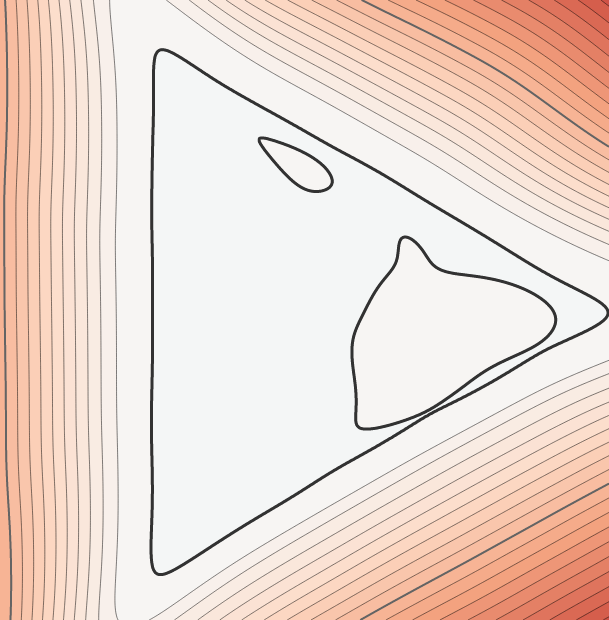} \\
    \includegraphics[width=0.09\textwidth]{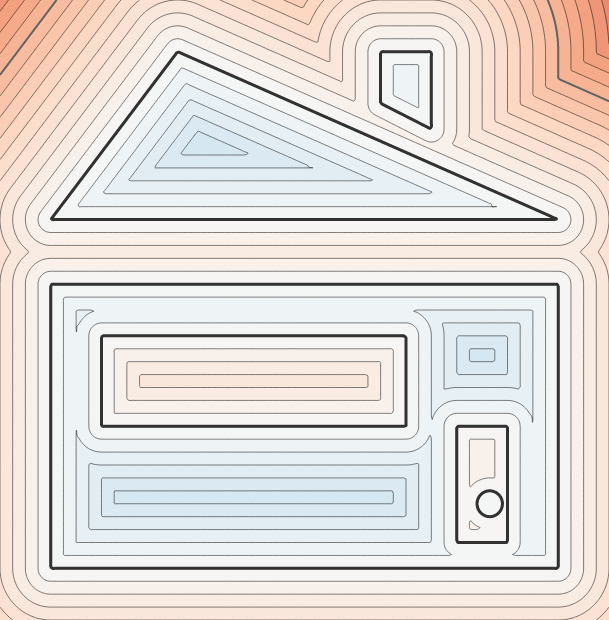} & \includegraphics[width=0.09\textwidth]{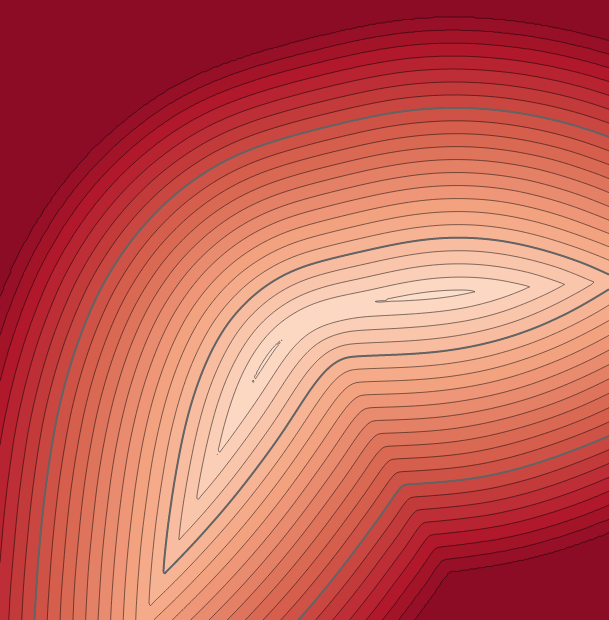} & \includegraphics[width=0.09\textwidth]{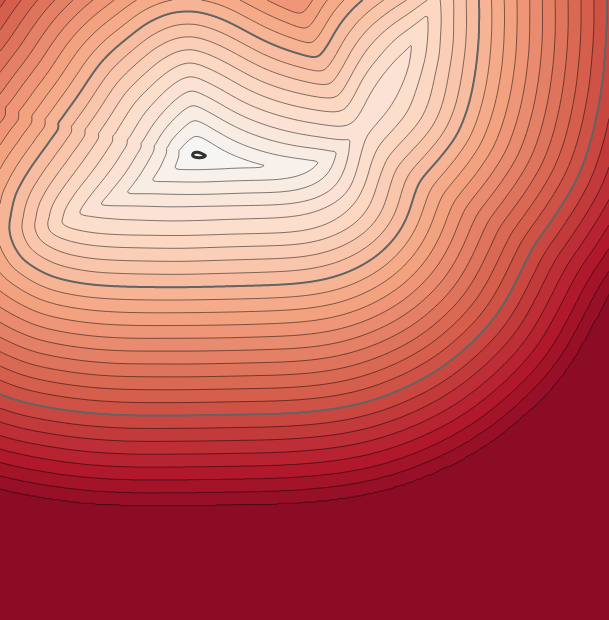} & \includegraphics[width=0.09\textwidth]{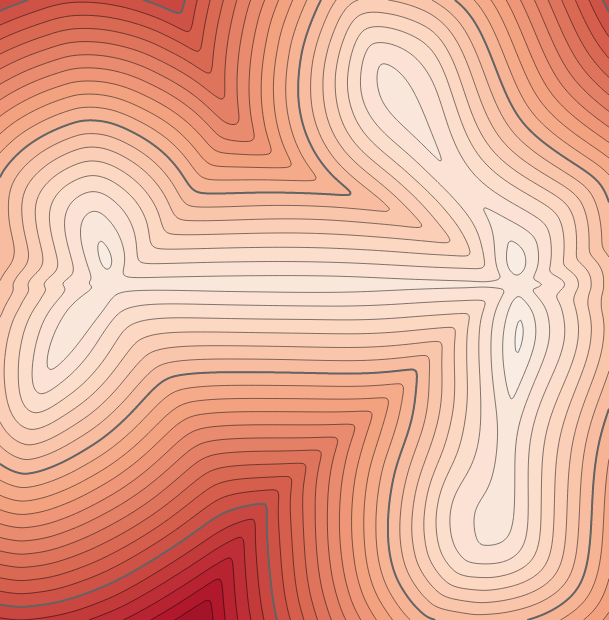} & \includegraphics[width=0.09\textwidth]{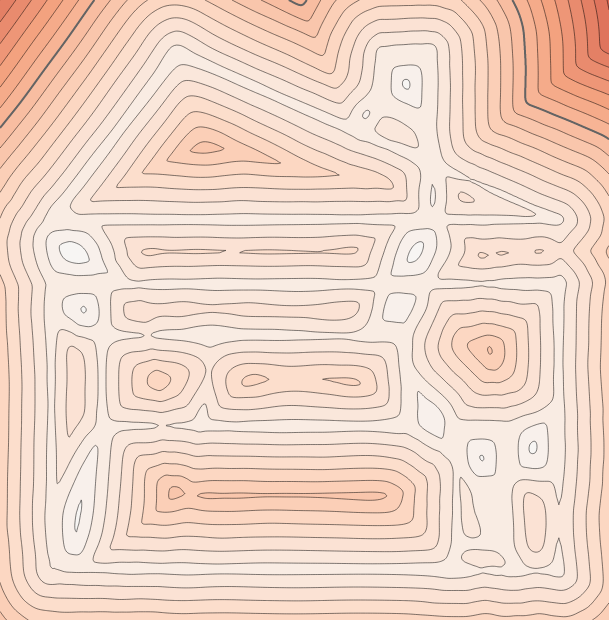} & \includegraphics[width=0.09\textwidth]{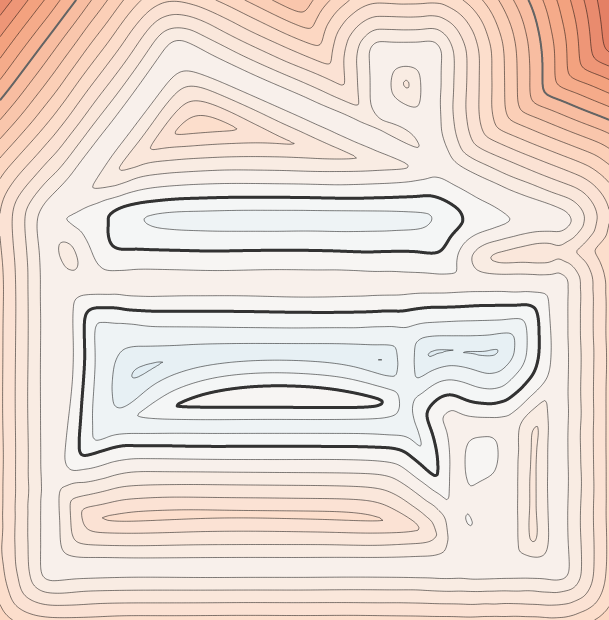} & \includegraphics[width=0.09\textwidth]{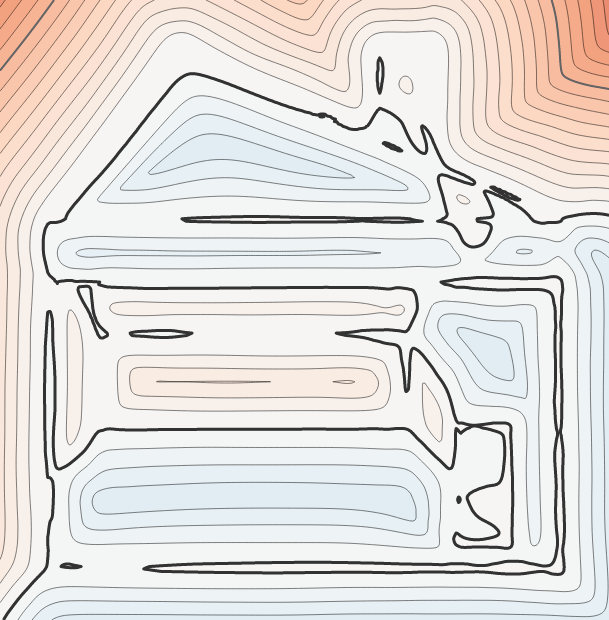} & \includegraphics[width=0.09\textwidth]{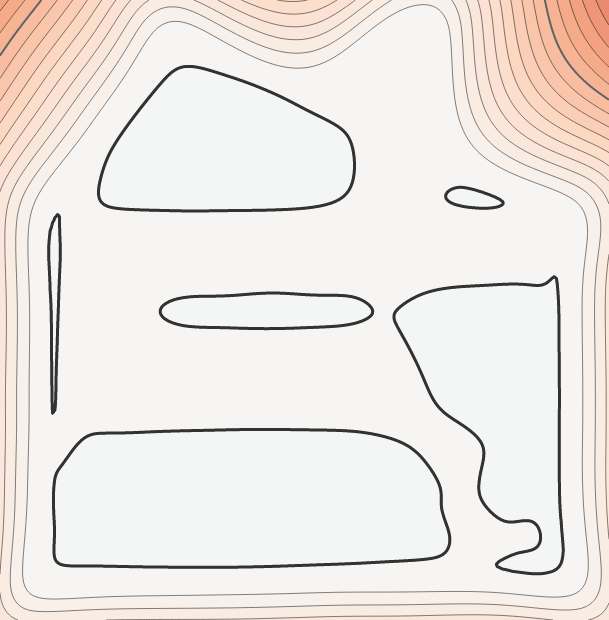} & \includegraphics[width=0.09\textwidth]{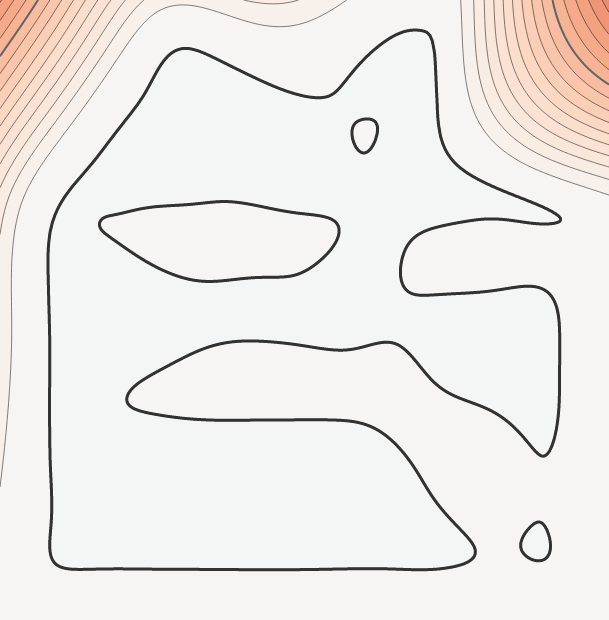} & \includegraphics[width=0.09\textwidth]{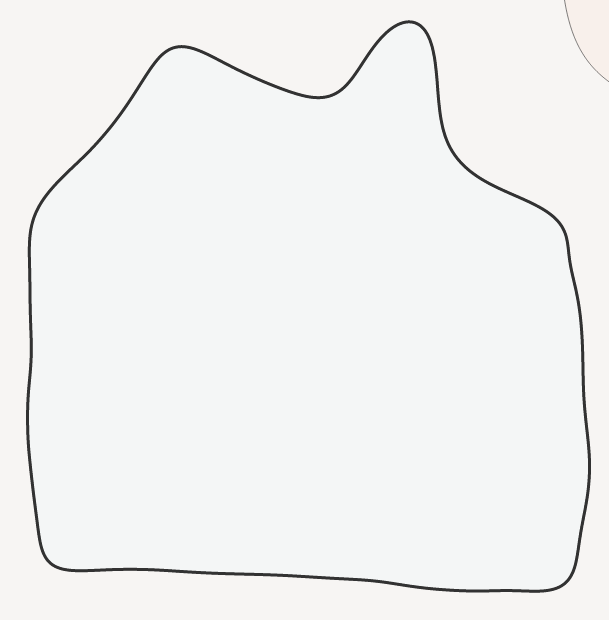} \\
        
    GT & $w_b = 0.1$ & $w_b = 0.2$ & $w_b = 0.3$ & $w_b = 0.5$ & $w_b = 1.0$ & $w_b = 2.0$ & $w_b = 5.0$ & $w_b = 10.0$ & $w_b = 20.0$ \\[1ex]
    \end{tabular}
    \caption{Comparison of PHASE results with different $w_b$ values on the rest of the 2D dataset. The color scale is the same as in \Cref{fig:phase-boundary-weight}.}
    \label{fig:phase-boundary-weight-more}
\end{figure*}